\documentclass[runningheads]{llncs}
\usepackage{amsmath}

\usepackage{eccv}

\usepackage{eccvabbrv}

\usepackage{graphicx}
\usepackage{booktabs}
\usepackage{textcomp}
\usepackage{algpseudocode}
\usepackage{algorithm}
\usepackage{paralist}

\usepackage[accsupp]{axessibility}  

\usepackage{hyperref}

\usepackage{orcidlink}
\usepackage{pifont}

\definecolor{darkgreen}{rgb}{0.0, 0.5, 0.0}
\definecolor{royalblue}{RGB}{50, 92, 168}
\definecolor{coral}{rgb}{1.0, 0.5, 0.31}
\definecolor{purple}{rgb}{0.5, 0., 0.5}

\def\ie{\emph{i.e.},\xspace}
\def\eg{\emph{e.g.},\xspace} 
\def\wrt{\emph{w.r.t.}\xspace}
\newcommand{\period}{Gfreq\xspace}

\newcommand{\objreg}{\texttt{\{object\} in \{region\}}\xspace}
\newcommand{\synreg}{\texttt{\{synonym\} in \{region\}}\xspace}

\newcommand{\methodname}
{Vendi Score Guidance\xspace}
\newcommand{\methodnameshort}{VSG\xspace}

\begin{document}

\title{Improving Geo-diversity of Generated Images with Contextualized Vendi Score Guidance}

\titlerunning{Contextualized Vendi Score Guidance}

\author{Reyhane Askari Hemmat\inst{1} \thanks{Equal Contribution.} \and
Melissa Hall\inst{1} $^*$\and
Alicia Sun\inst{1}\and
Candace Ross\inst{1}\and \\
Michal Drozdzal\inst{1}\and
Adriana Romero-Soriano \inst{1, 2, 3, 4}}

\authorrunning{R. Askari Hemmat, M. Hall et al.}

\institute{$^1$ FAIR at Meta, $^2$ Mila, $^3$ McGill University, $^4$ Canada CIFAR AI Chair}

\maketitle

\begin{abstract}
    With the growing popularity of text-to-image generative models, there has been increasing focus on understanding their risks and biases. 
    Recent work has found that state-of-the-art models struggle to depict everyday objects with the true diversity of the real world and have notable gaps between geographic regions. 
    In this work, we aim to increase the diversity of generated images of common objects such that per-region variations are representative of the real world.
    We introduce an inference-time intervention, \textit{contextualized \methodname (c-\methodnameshort)}, that guides the backwards steps of latent diffusion models to increase the diversity of a sample as compared to a ``memory bank'' of previously generated images while constraining the amount of variation within that of an exemplar set of real-world contextualizing images. 
    We evaluate c-\methodnameshort with two geographically representative datasets and find that it substantially increases the diversity of generated images, both for the worst-performing regions and on average, while simultaneously maintaining or improving image quality and consistency.
    Additionally, qualitative analyses reveal that diversity of generated images is significantly improved, including along the lines of reductive region portrayals present in the original model.
    We hope that this work is a step towards text-to-image generative models that reflect the true geographic diversity of the world.\footnote{\textcolor{blue}{\url{https://github.com/facebookresearch/Contextualized-Vendi-Score-Guidance}}}
  \keywords{geo-diversity \and image generation \and diffusion models}
\end{abstract}

\begin{figure}[!ht]
    \centering
    \begin{subfigure}[b]{0.45\textwidth}
        \includegraphics[width=0.99\textwidth]{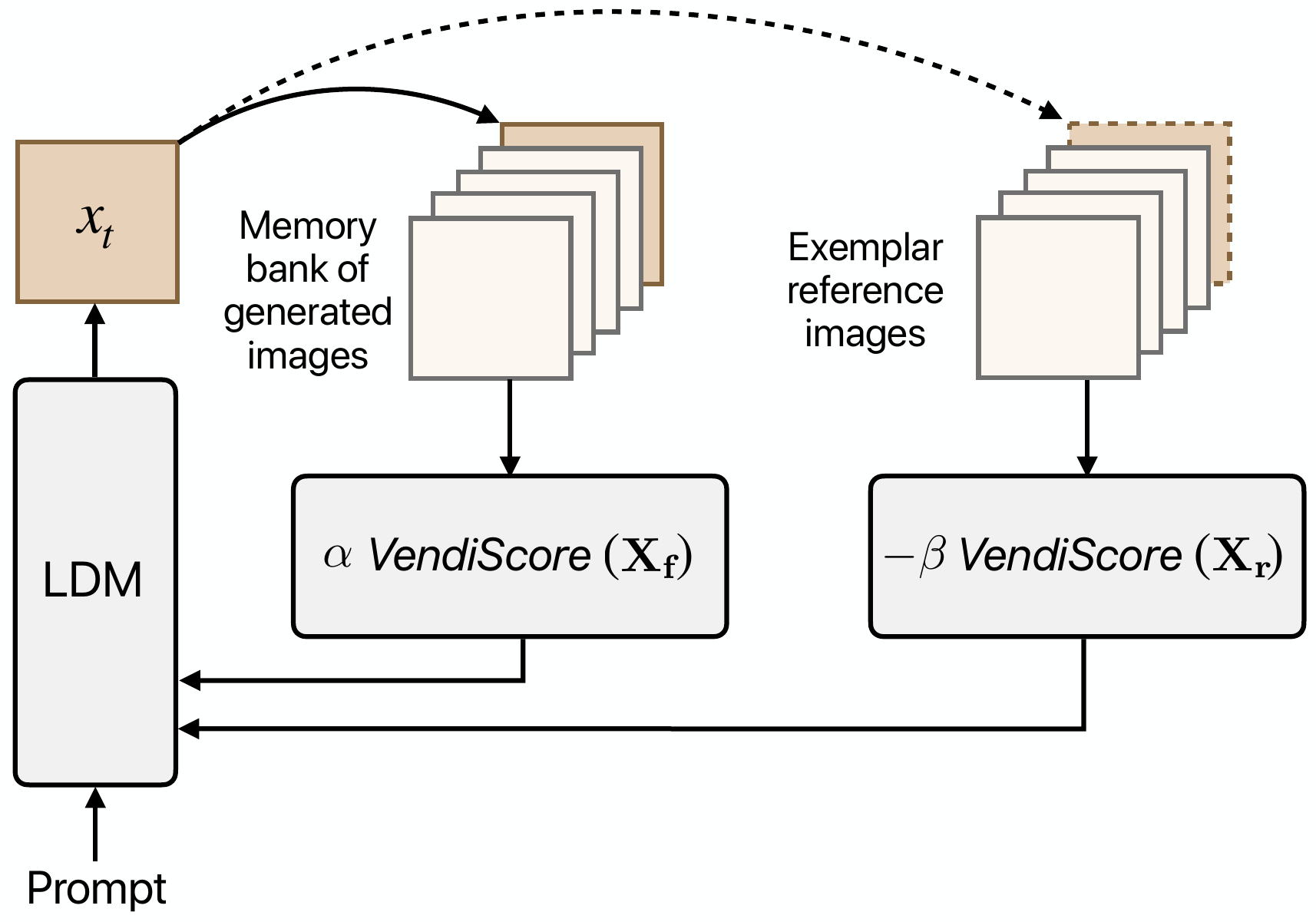}
        \caption{}
    \end{subfigure}
    \begin{subfigure}[b]{0.45\textwidth}
        \includegraphics[width=0.99\textwidth]{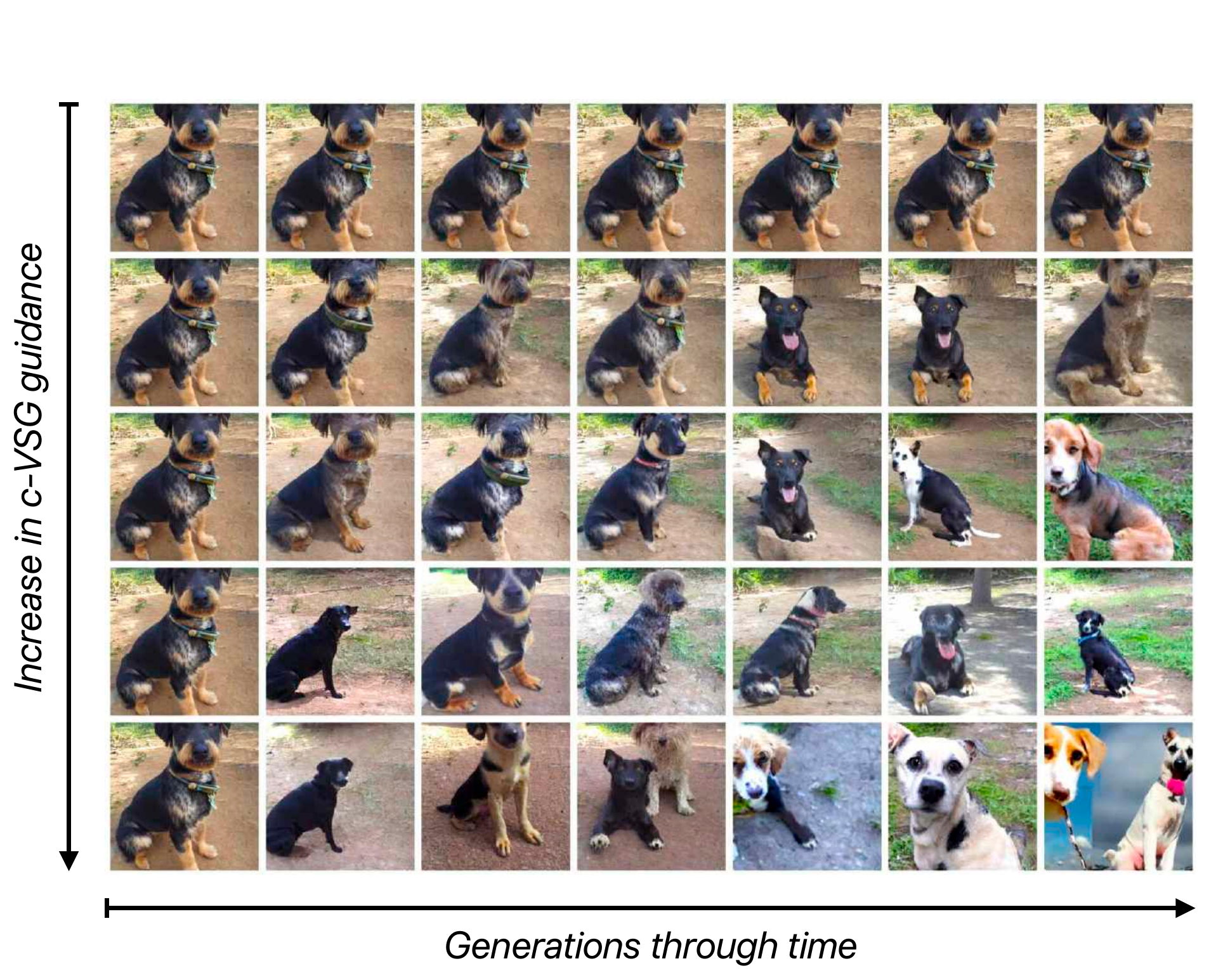}
        \caption{}
    \end{subfigure}
    
    \caption{
    (a) We present \textbf{Contextualized \methodname (c-\methodnameshort)}, an inference-time intervention to increase the diversity of images generated by latent diffusion models (LDMs). 
    c-\methodnameshort guides backwards steps of the diffusion process using the Vendi Score \cite{friedman2022vendi} to increase the diversity among a sample $x_t$ and a memory bank of previous generations (with weight $\alpha$) while constraining excessive variation using a small set of real, contextualizing exemplar images (with weight $\beta$). (b) Generations of \texttt{dog in Africa}, \textit{all with the same seed.} First row has zero c-VSG guidance scale and as a result all samples are the same. As we increase the c-VSG guidance scale, we observe increased diversity in generations.
    }
    \label{fig:enter-label}
\end{figure}

\section{Introduction}
\label{sec:intro}

The unprecedented results achieved by text-to-image systems~\cite{ramesh2022hierarchical,DBLP:journals/corr/abs-2112-10752,DBLP:journals/corr/abs-2112-10741,gafni2022make} and their wide-spread use as plug-and-play solutions have propelled a body of research to understand their potential risks and biases~\cite{Cho2022DallEval,luccioni2023stable,bansal2022how}. 
Recent works have highlighted through qualitative and quantitative evaluations the disparities in demographic traits of people represented in the generated images~\cite{naik2023social,wan2024male,hao2024harm}, triggering the design of mitigation strategies based on textual interventions~\cite{gal2022image}, attention-weight modification~\cite{gandikota2023erasing}, and semantic guidance~\cite{friedrich2023fair}.

Beyond human-centric representations, research has shown that stereotypical biases also occur for representations of objects and their surroundings across the globe~\cite{bianchi2022easily,basu2023inspecting,hall2023dig}. 
In particular, previous studies highlighted that progress in image quality or realism has come at the expense of representation diversity and text-image consistency ~\cite{hall2023dig}.
This trade-off affects some regions more than others, hindering the progress towards conditional image generative systems that truly work well for all geographic regions. 
Notably, images generated with text prompts that include regional information contain limited object diversity and heightened geographic stereotypes. 
As shown in~\cite{hall2023dig}, many generations depict \textit{limited diversity in object type}.
Yet, to the best of our knowledge, no mitigation strategies have directly targeted geo-diversity in text-to-image systems.\looseness-1

In this paper, we aim to mitigate disparities in representation diversity of common objects across worldwide regions. More precisely, our goal is to increase the objects' representation diversity such that per-region variation in type of object, 
as observed in the real world, is better reflected in the generations. In doing so, we aim to maintain (or improve) the quality of images as well as their text-image consistency.

We devise a novel approach called \methodname (VSG) that increases the representation diversity of the generations by leveraging the Vendi Score (VS) \cite{friedman2022vendi} – a dataset diversity evaluation metric. In particular, VSG extends the guidance toolbox of diffusion models by driving the generation process towards samples that are substantially different from each other, intuitively maximizing the effective number of samples among the obtained generations. This requires the proposed guidance strategy to operate \emph{auto-regressively \wrt the previous generations}. We implement this by keeping a \emph{memory bank} of past generations that ensures that the current sample differs from those in the memory.\looseness-1
do 
Yet, increasing diversity in an unconstrained way results in generations that lack either image realism or text-generation consistency. Therefore, we propose to contextualize VSG by providing a small set of randomly selected real images of objects, referred to as \emph{exemplar images}, and guiding the generation process towards diverse samples that are grounded on those exemplar images. We evaluate the proposed approach by measuring precision~\cite{Sajjadi2018_PR,kynkäänniemi2019improved}, recall~\cite{Sajjadi2018_PR,kynkäänniemi2019improved}, and CLIPScore~\cite{hessel2021clipscore} to assess image realism/quality, representation diversity, and text-image consistency, respectively. We report results on two geographically diverse datasets that contain images of the common objects collected around the world, GeoDE~\cite{ramaswamy2022geode} and DollarStreet~\cite{rojas2022dollar}. 

Our results show that contextualized \methodname improves worst-region and average F1 for both datasets, including a relative improvement of 40\% in worst-region F1 over the Latent Diffusion Model (LDM) baseline without any intervention when measured on GeoDE and 11\% over the next closest baseline, and reduces performance disparities across regions. 
Even without contextualization, \methodnameshort shows relative improvements of up to 10\% over the LDM baseline in F1. 
Upon visual inspection we find the object diversity in images to be significantly improved, with greater variation in object color, type, shape, and size. 
In addition, \methodnameshort improvements often coincide with improved regional representations beyond reductive portrayals present in the original model.
Finally, we find in ablations that our combined use of the Vendi Score to encourage image diversity relative to previous generations while constraining variation to that of the real world with contextualizing exemplar images meaningfully navigates the diversity \vs quality tradeoff.

The contributions of this work can be summarized as follows:
\begin{compactitem}
    \item  We introduce a new inference-time intervention, contextualized \methodname (c-\methodnameshort), to increase the diversity of images generated by LDMs using a memory bank of previously generated images and exemplar images.\looseness-1
    \item When evaluated with two geographically representative real world datasets, c-\methodnameshort shows significant improvement in diversity and quality over the vanilla LDM, while maintaining text-image consistency.\looseness-1 
    \item c-\methodnameshort shows improvements in average and worst-region F1 over state-of-the-art baselines, while exhibiting reduced disparities in quality, diversity and text-generation consistency across regions. 
    \item Our ablations demonstrate the efficacy of the memory and contextualizing components of the c-\methodnameshort criterion. 
\end{compactitem}
We hope this work contributes to image generations that better reflect the geographic diversity of the real world.

\section{Method}
\label{sec:methodology}

In this work, we focus on increasing the diversity of latent diffusion models (LDMs). We first provide preliminary details about the components of the LDM generation process in which we apply our interventions, as well as the Vendi Score metric. Then, we introduce our proposed methodology, \methodname.

\subsection{Preliminaries}
\subsubsection{Sampling in LDMs.} 
In this work, we focus on a class of LDMs called denoising diffusion implicit models (DDIMs)~\cite{song2020denoising}.
At inference or generation time, DDIMs use a reverse process defined as:
\begin{equation} \label{eq:DDIM}
    x_{t-1} = \sqrt{{\xi}_{t-1}} \hat{x}_{0, t} + \underbrace{\sqrt{1 - {\xi}_{t-1} - \sigma_t^2} \cdot \epsilon_\theta^{(t)}(x_t)}_{\text{direction pointing to \( x_t \)}} + \underbrace{\sigma_t \epsilon_t}_{\text{random noise}},
\end{equation}
where $x_t$ is the sample at time step $t$, $\epsilon_t \sim \mathcal{N}(\mathbf{0}, \mathbf{I}) $ and \( {\xi}_t \) and \( \sigma_t \) are time-dependent coefficients, and $\hat{x}_{0, t}$ is the DDIM approximation of the denoised sample given current $x_t$,
\begin{equation}\label{eq:DDIM_approx}
    \hat{x}_{0, t} = \text{DDIMApprox} (x_t) := \frac{x_t - \sqrt{1 - {\xi}_t} \epsilon_\theta^{(t)}(x_t)}{\sqrt{{\xi}_t}},
\end{equation}
and $ \epsilon_\theta^{(t)}$ is the pretrained denoising network parameterized by $\theta$ that approximates the score function conditioned on a class or prompt $y$,
\begin{equation}
    \epsilon_\theta^{(t)} \approx \nabla_x \log p(x_t |y).
\end{equation}
In this work, we modify this score function to guide the sampling process towards generating more diverse samples.

\textbf{Guidance in LDMs.}
LDMs often condition generated samples on labels or text prompts in the score function. 
If one has access to a pretrained denoising network that only approximates the \textit{unconditional} score function ($\nabla_x \log p(x)$), one can use classifier-guidance \cite{dhariwal2021diffusion} to generate a \textit{conditional} sample.
Classifier-guidance modifies the score function using a pretrained classifier that provides $\log p(y|x_t)$,
\begin{equation}\label{eq:conditiona_score}
 \epsilon_\theta^{(t)} \approx \nabla_x \log p_\gamma(x_t|y) = \nabla_x \log p(x_t) + \gamma \nabla_x \log p(y|x_t),
\end{equation}
where \( \gamma \) is a scaling factor controlling the strength of class-conditional generation. In this work, we adapt this concept of guidance by utilizing the Vendi Score \cite{friedman2022vendi}, rather than a classifier, as our guidance function.

\textbf{Vendi Score (VS).}
The VS is a metric for evaluating diversity in machine learning~\cite{friedman2022vendi}.
It applies a user-defined similarity function to a set of samples, without the need for a reference dataset as opposed to metrics such as recall or coverage~\cite{Sajjadi2018_PR,kynkäänniemi2019improved, naeem2020reliable}. VS measures the effective number of examples by computing the soft-rank of a similarity or kernel matrix. The similarity matrix is calculated using a similarity function that takes two samples and returns a positive value indicating how similar they are. Higher rank of the similarity matrix corresponds to higher Vendi Score and diversity.

Formally, given a set of samples $x_1, \ldots, x_n$, we can define a positive semi-definite similarity function $k$ with $k(x, x) = 1$ for all $x$. The similarity function is applied on every pair of samples in our dataset to create a similarity matrix, $\mathcal{K}$ where $\mathcal{K}_{i, j} = k(x_i, x_j)$. VS is then defined as the soft-rank of $\mathcal{K} / n$, \ie the exponential of the entropy of the eigenvalues of $\mathcal{K}/n$~\cite{friedman2022vendi}:

\begin{equation}
\label{eq:vendi-score}
    \text{VS} (x_1, \ldots, x_n) =  \exp\left( - \sum_{i=1}^n \lambda_i \log \lambda_i \right), 
  \end{equation}
  where $\lambda_1, \ldots, \lambda_n$ are the eigenvalues of $\mathcal{K}/n$.

A common choice for the similarity function is the cosine-similarity of every pair of samples in the dataset in a pre-defined feature space.\looseness-1 

\subsection{\methodname}
In this section, we introduce our proposed approach, called \methodname (\textbf{\methodnameshort}). Our goal is to increase the diversity of the generated samples by incorporating the Vendi Score as an auto-regressive guidance mechanism during diffusion model inference. 
In other words, every new generation will take into consideration all previously generated samples and will aim to generate a sample which is substantially different from the previous ones.

Let us define a \emph{memory bank} which contains all the previously generated samples as $\mathbf{X_{f}}: \{x^1,..., x^n \}$. Given the $\mathbf{X_{f}}$, and a newly generated sample $x$, we can compute $\text{VS}(x,\mathbf{X_{f}})$. Our goal is to modify the generation process of $x$ by steering it towards part of the manifold that increases VS, and thus, augment the sample diversity. We can achieve this via the following score function, 
\begin{equation}\label{eq:vendi_score_function}
    \epsilon_\theta^{(t)} \approx \nabla_x \log  {p}_{\gamma, \alpha} (x_t|y, \mathbf{X_{f}}) = \nabla_x \log {p_{\gamma}}(x_t| y) + \alpha \nabla_x  \text{{VS}}(\hat{x}_{0, t}, \mathbf{X_{f}}),    
\end{equation}
where $\alpha$ is a scaling factor that controls the strength of the Vendi Score guidance and $\hat{x}_{0, t}$ is the predicted denoised sample.

A naive way of finding $\hat{x}_{0, t}$ is to apply all the $T$ steps of the backward diffusion process and use the denoised image to compute the Vendi Score. 
However, this is computationally expensive. 
Thus, we use the efficient DDIM approximation of a denoised sample, given in Eq.~\ref{eq:DDIM_approx}. 

We note that the original Vendi Score formulation in Eq. \ref{eq:vendi-score} is not differentiable, since it requires finding the singular values of the similarity matrix, and singular value decomposition is not differentiable.
To overcome this limitation, we developed a simple algorithm to compute the Vendi Score in a differentiable way and present it in Appendix \ref{app:dvs_derivation}. 

\subsubsection{Contextualized \methodname (c-\methodnameshort).}
In experimentation, we found that unconstrained VSG can have limited efficacy, as generated images may drift from real world representations. Thus, in the second part of our method, we use Vendi Score a second time with a small set of randomly selected exemplar images, $\mathbf{X_r}$. This set of exemplar images is used to contextualize the Vendi Score computation and thus, the generation process. 
Similar to Eq.~\ref{eq:vendi_score_function}, we compute $\text{VS}({x}, \mathbf{X_r})$. However, because we want to \textit{reduce} excessive drift from the exemplar representation of images, we calculate the gradients with respect to the \textit{negative} of this score. Overall, we use the following score function which combines VSG with contextualized samples:
\begin{equation}\label{eq:c-VSG}
    \nabla_x \log  {p} (x_t|y, \mathbf{X_f}, \mathbf{X_r}) = \nabla_x \log  {p}_{\gamma, \alpha} (x_t|y, \mathbf{X_{f}}) - \beta \nabla_x  \text{{VS}}(\hat{x}_{0,t}, \mathbf{X_{r}}),    
\end{equation}
where $\beta$ is a scaling factor that controls the strength of VSG contextualization. This formulation allows the model to generate samples that remain close to the exemplar images without the need for additional training.
Intuitively, the modified score function (Eq.~\ref{eq:c-VSG}) steers the generation process towards parts of the space which would result in a sample that increases the rank of the memory bank while staying close to the exemplar images.

Algorithm \ref{alg:main} presents the c-\methodnameshort computation step by step. Note that we use $\period$ to control the rate at which c-\methodnameshort is applied in the diffusion process, allowing for a balance between efficiency and diversity. To generate image $x^n$, we apply the the proposed guidance in the backward diffusion process. In the intermediate steps and when the sample is not yet generated, we use DDIM's approximation of $\hat{x}_0$ and feed it to the VS computation. After the final diffusion step and once the image is completely denoised, $x^n$ is added to the memory bank of generated images and used for future image generations.\looseness-1

\begin{algorithm}[t!]
\caption{Contextualized \methodname (c-\methodnameshort)}
\label{alg:main}
\begin{algorithmic}
\Require Class label $y$, a single fake image $x_f$, a tensor of $M$ real images $\mathbf{X_r}$, pre-trained diffusion model $\epsilon_\theta$, $N$ number of generations to obtain, $T$ number of diffusion steps, $Gfreq$ guidance frequency, $\alpha$, $\beta$ and $\gamma$ guidance weights
\Ensure $\mathbf{X_f}$ a tensor with $N$ diverse model samples
\State $\mathbf{\mathbf{X_f}} \gets x_f$
\Comment{Tensor with diverse model samples}
\For{$n=1, \dots, N-1$}
    \State $x^n \sim \mathcal{N}(\mathbf{0}, \mathbf{I})$
    \Comment{Initialize sample with random noise}
    \For{$t=T, \dots, 1$} \Comment{Apply backward diffusion steps}
        \If{$t \ \% \ \period == 0$}
            \State $\hat{x}_{0, t}^n =$ DDIMApprox($x_t^n$) \Comment{Eq.~\ref{eq:DDIM_approx}}

            \State $\epsilon_\theta^{(t)} \gets \nabla_x \log {p_{\gamma}}(x_t| y) + \alpha \nabla_x  \text{{VS}}(\hat{x}_{0, t}^n, \mathbf{X_{f}}) - \beta \nabla_x  \text{{VS}}(\hat{x}_{0, t}^n, \mathbf{X_{r}})$
            \Comment{Eq.~\ref{eq:c-VSG}}
            \State $x_{t-1}^n \gets $ one step of DDIM 
            \Comment{Eq.~\ref{eq:DDIM}}
        \Else
        \State $\epsilon_\theta^{(t)} \gets \nabla_x \log {p_{\gamma}}(x_t| y)$
        \State $x_{t-1}^n \gets $ one step of DDIM
        \Comment{Eq.~\ref{eq:DDIM}}
        \EndIf
        \EndFor
        \State $\mathbf{\mathbf{X_f}} \gets \textrm{cat}(\mathbf{\mathbf{X_f}},x_0^n)$ \Comment{Add final denoised sample to the memory}
       
\EndFor

\end{algorithmic}
\end{algorithm}

\section{Experiments}
\label{sec:experiments}

In this section, we describe evaluation datasets and metrics, baselines, and set-up. We then present experimental results and ablations.\looseness-1

\subsection{Datasets and metrics}
\label{subsec:dataandmetrics}
In our experiments, we report worst-region and average F1, recall, precision, and 
CLIPScore. 
We follow~\cite{hall2023dig} and evaluate performance \wrt two geographically diverse datasets containing images of objects in multiple regions: GeoDE \cite{ramaswamy2022geode} and DollarStreet \cite{rojas2022dollar}. 
Both datasets contain images of objects in their everyday settings and are intended to be geographically representative. 
However, images in GeoDE were submitted by people living in the regions of interest and were collected with specific guidelines, \eg that the object of focus fills at least 25\% of the image, while images in DollarStreet were collected by photographers who travelled to various regions, with a special focus on ``disadvantaged and isolated areas'' \cite{rojas2022dollar}. 
Thus, DollarStreet likely has more income variation than GeoDE. 
Furthermore, DollarStreet contains large imbalances between regions. 
We filter GeoDE and DollarStreet. 
The remaining images are used as the pool for randomly selected exemplar real images. 
We subsample DollarStreet so that images contain only one concrete class, \eg removing classes corresponding to favorite objects and images with multiple classes, and use the original distribution exclusive of the images used for contextualizing. 
For CLIPScore, we compute the average of the 10$th$-percentile for each object class, as prior work suggests that the lower tail of the distribution tends to be most informative in capturing generation inconsistencies \cite{hall2023dig,he2022synthetic}.

\subsection{Baselines}
We split our discussion between baselines that do not utilize any information specific to the task in order to improve generation diversity, and those that use small amounts of information from the task. 
To see examples of generations for different methods, see Appendix \ref{app:samples}.

\noindent\textbf{Without any additional information:} 
\begin{compactitem}
    \item \texttt{LDM}: This is the baseline setup where the unaltered LDM is used with the prompt \objreg to condition the generation process.
    \item \texttt{Synonyms}: This strategy maps each object class to its corresponding ImageNet~\cite{deng2009imagenet} class and WordNet~\cite{miller-1994-wordnet} synset. Each synset contains lemmas representing a specific ``sense'' of the given class. We generate images stratified across all possible lemmas (including the original object word), for each class, \ie \synreg is used to condition the generation process. The list of synonyms can be found in Appendix \ref{sec:app_setup}.
\end{compactitem}
\noindent\textbf{With additional information:} 
\begin{compactitem}  
    \item \texttt{Paraphrasing}: We use the LLaMA-2-70B-chat \cite{touvron2023llama} large language model  to generate paraphrases of the original prompt template, \objreg. 
    We include the specifications and descriptions used in the collection of GeoDE and DollarStreet.
    The metaprompts and paraphrases, as well as our method of tuning prompts and model specifications, are included in Appendix \ref{sec:app_setup}. 
    \item \texttt{Semantic Guidance}: As observed in~\cite{hall2023dig} and highlighted in \cref{fig:diversity_issue_example}, generated images tend to contain diversity issues related to the magnification of region-specific representations of objects beyond those in the evaluation task. 
    To reduce this amplification of region information in the generated images, we employ Semantic Guidance~\cite{brack2023sega,friedrich2023fair} by adding negative guidance corresponding to the region term for each generation. 
    Note that this baseline requires knowing the biases embedded in the text-to-image system upfront.\looseness-1
    \item \texttt{Feedback Guidance (FG)}: Following~\cite{hemmat2023feedbackguided,sehwag2022generating}, during generation we use feedback from an external CLIP-based~\cite{radford2021learning} classifier to predict region labels of the generated images\footnote{In this baseline, we do not implement the additional real/fake discriminator as suggested in~\cite{sehwag2022generating}, as we focus only on methods that utilize existing classifiers.}. We experiment with two versions of feedback guidance to encourage more diverse generations: one that maximizes the \texttt{loss} of the classifier and one that maximizes the \texttt{entropy} of predicted class distributions.\looseness-1
    \item \texttt{Textual Inversion}: Textual inversion injects a learned token embedding using small subset of images from the evaluation task to better represent the reference images~\cite{gal2022image}. We apply textual inversion by learning an embedding for each object in the evaluation dataset using four images per object.
\end{compactitem} 

\subsection{Experimental set-up}
\label{subsec:set_up}

We perform all experiments using an open sourced version of an LDM trained on a large scale dataset~\cite{DBLP:journals/corr/abs-2112-10752}. 
For c-\methodnameshort, we use $\alpha=1$ and  $\beta=2$ unless otherwise specified, with two exemplar images as real world context and set $Gfreq$ to 5. 
We calculate the Vendi Score in the CLIP ViT-B32 \cite{radford2021learning} feature space. 
For each method, we perform a hyperparameter search across guidance scales, as well as the weighting of contextual information, where applicable. 
We perform one-region-out hyper-parameter selection, selecting the best hyper-parameters for a given region based on the F1 for all the other regions. For computational complexity and efficiency see Appendix \cref{app:comp_complexity}.\looseness-1

\subsection{Results}
\label{sec:results}

\begin{table}[t!]
 \caption{\textbf{Comparison to baselines on GeoDE.} Contextualized \methodname contributes to greater diversity improvements (recall) with increases, or little cost to, quality (precision) and consistency (CLIPScore), both on average and for the worst performing region (determined by F1). AF: Africa, WAS: West Asia, ``label'' refers to region label, ``desc'' to text description, and ``img'' to exemplar images.\looseness-1}
  \label{tab:geode_res}
\centering
\resizebox{\textwidth}{!}{%
\setlength{\tabcolsep}{2pt}
\begin{tabular}{lcccccccccc}
\toprule
\textbf{Method} & \textbf{Ref.} & \textbf{Worst-} & \multicolumn{2}{c}{\textbf{F1}} & \multicolumn{2}{c}{\textbf{Precision}} & \multicolumn{2}{c}{\textbf{Recall}} & \multicolumn{2}{c}{\textbf{CLIPScore}}\\
\cmidrule(lr){4-5}
\cmidrule(lr){6-7}
\cmidrule(lr){8-9}
\cmidrule(lr){10-11}
& \textbf{Info.} & \textbf{Reg.} & \text{Avg.} & \text{Worst-Reg.} & \text{Avg.} & \text{Worst-Reg.} & \text{Avg.} & \text{Worst-Reg.} & \text{Avg.} & \text{Worst-Reg.}\\
\midrule
    LDM   &\ding{55} & AF & 0.364 & 0.322 & {0.413} & 0.273 & 0.337 & 0.395 & 0.242 &0.218 \\
    Synonyms   &\ding{55} & AF & 0.357 & 0.306 & 0.350 & 0.298 & 0.366 & 0.315 & 0.215 & 0.203 \\
    \textbf{\methodnameshort (ours)}  &\ \ding{55} & AF& {0.399} & {0.356} & 0.349 & {0.307} & {0.470} & 0.424 & 0.180 & 0.191\\
    \midrule
    Paraphrasing & desc  & WAS & 0.384 & 0.354 & 0.338 & 0.309 & 0.449 & 0.415 & 0.231 & 0.228 \\
    Semantic Guidance  & label &WAS & 0.420 & 0.401 & \textbf{0.459} & \textbf{0.519} & 0.391 & 0.326 & 0.245 & \textbf{0.253} \\
    FG CLIP (Loss)  & label & WAS & 0.409& 0.378 & 0.387 & 0.383 & 0.436 & 0.373 & 0.228 & 0.223 \\
    FG CLIP (Ent.)  & label & AF & 0.380 & 0.337 & 0.344 &0.329  & 0.429 & 0.345 & 0.224 & 0.227\\
    Textual Inversion  & img & AF & 0.369 & 0.363 & 0.409 & 0.444 & 0.338 & 0.308 & 0.234 & 0.232\\
    \textbf{c-\methodnameshort (Ours)}  & img & AF &  \textbf{0.455} & \textbf{0.444} & 0.424 & 0.417 & \textbf{0.493} & \textbf{0.476} & \textbf{0.254} & 0\textbf{.253}\\
\bottomrule
\end{tabular}}%
\end{table}

\begin{table}[t!]
   \caption{\textbf{Comparison to baselines on DollarStreet.} Contextualized \methodname contributes to greater diversity improvements (recall) with little cost to quality (precision) and consistency (CLIPScore), both on average and for the worst performing region (determined by F1). c-\methodnameshort outperforms all the methods by at least 8\% relative F1 improvement. AF: Africa, AS: Asia, ``label'' refers to region label, ``desc'' to text description, and ``img'' to exemplar images.\looseness-1}
   \label{tab:ds_res}
\centering
\resizebox{\textwidth}{!}{%
\setlength{\tabcolsep}{2pt}
\begin{tabular}{lcccccccccc}
\toprule
\textbf{Method} & \textbf{Ref.} &  \textbf{Worst-} & \multicolumn{2}{c}{\textbf{F1}} & \multicolumn{2}{c}{\textbf{Precision}} & \multicolumn{2}{c}{\textbf{Recall}} & \multicolumn{2}{c}{\textbf{CLIPScore}}\\
\cmidrule(lr){4-5}
\cmidrule(lr){6-7}
\cmidrule(lr){8-9}
\cmidrule(lr){10-11}
& \textbf{Info.} & \textbf{Reg.} &  \text{Avg.} & \text{Worst Reg.} & \text{Avg.} & \text{Worst Reg.} & \text{Avg.} & \text{Worst Reg.} & \text{Avg.} & \text{Worst Reg.}\\
\midrule
    LDM   &\ding{55} &AS &0.448 &0.442 &0.428 & 0.434& 0.472 &0.450 &0.231  &0.235  \\
    Synonyms   &\ding{55} & AS & 0.464 & 0.457 & 0.451 & 0.448 & 0.479 & 0.467 & 0.216 & 0.220\\
    \textbf{\methodnameshort (ours)}  &\ \ding{55}& AS & 0.457 &0.444 & 0.413& 0.388&0.516 & 0.518 & 0.191 & 0.198\\
    \midrule
    Paraphrasing & desc & AF & 0.454 & 0.445 & 0.445 &  0.454 & 0.465 & 0.437 & 0.226 & 0.215 \\
    Semantic Guidance  & label &AS & 0.470 & 0.458&0.447 & 0.449&0.498 & 0.467 & 0.230 & 0.233\\
    FG CLIP (Loss)  & label & AS & 0.437 & 0.394& 0.401& 0.321& 0.488& 0.510 & 0.223 & 0.206 \\
    FG CLIP (Entropy)  & label & AS & 0.465 &0.462 & 0.412 &0.404 & 0\textbf{.535} & \textbf{0.540} & 0.222 & 0.219\\
    Textual Inversion  &\ img & AS &0.425 &0.398 & 0.478& \textbf{0.491} &0.386 &0.335 & 0.217&0.219 \\
    \textbf{c-\methodnameshort (Ours)}  &\ img & AS &  \textbf{0.497} &\textbf{ 0.483} & \textbf{0.486} & {0.486}&0.511 &0.479  & \textbf{0.234} & \textbf{0.238} \\
\bottomrule
\end{tabular} }%
\end{table}

\begin{figure}[!t]
    \begin{subfigure}[b]{0.99\textwidth}
        \begin{subfigure}[b]{\textwidth}
            \centering
            \includegraphics[width=0.49\textwidth]{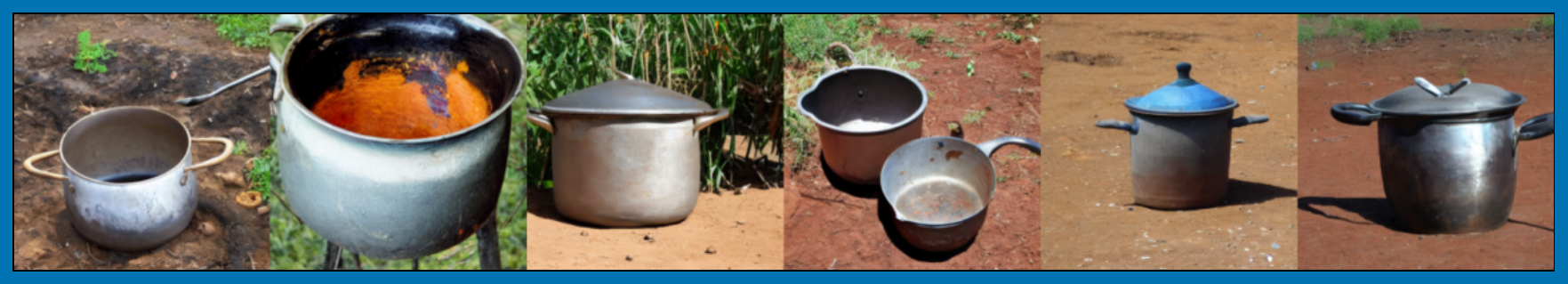}  
            \includegraphics[width=0.49\textwidth]{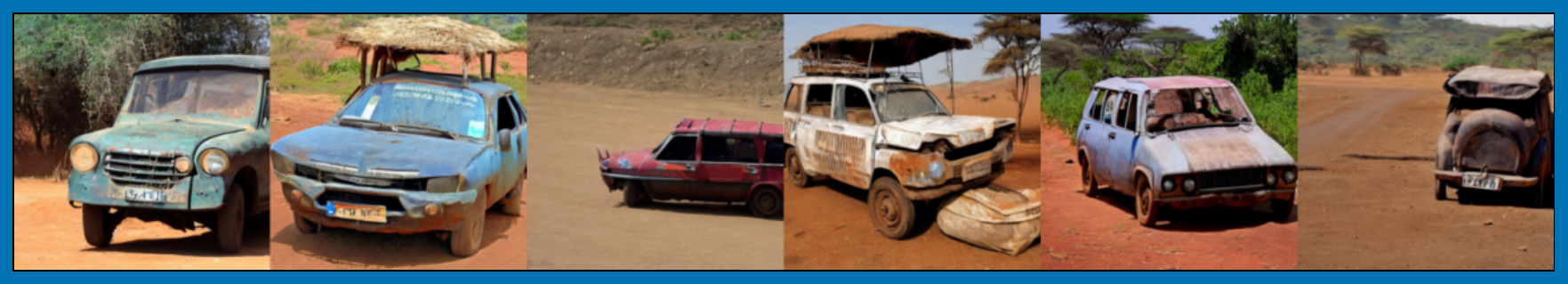}  \includegraphics[width=0.49\textwidth]{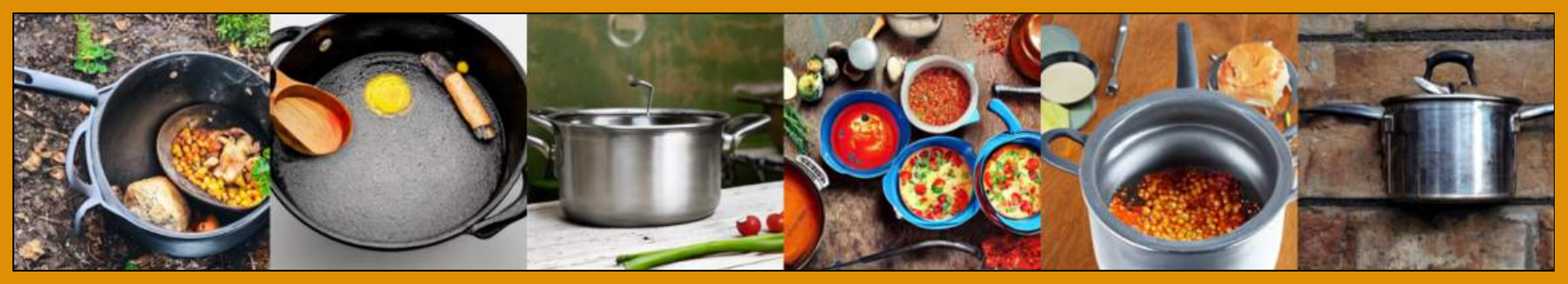}  
            \includegraphics[width=0.49\textwidth]{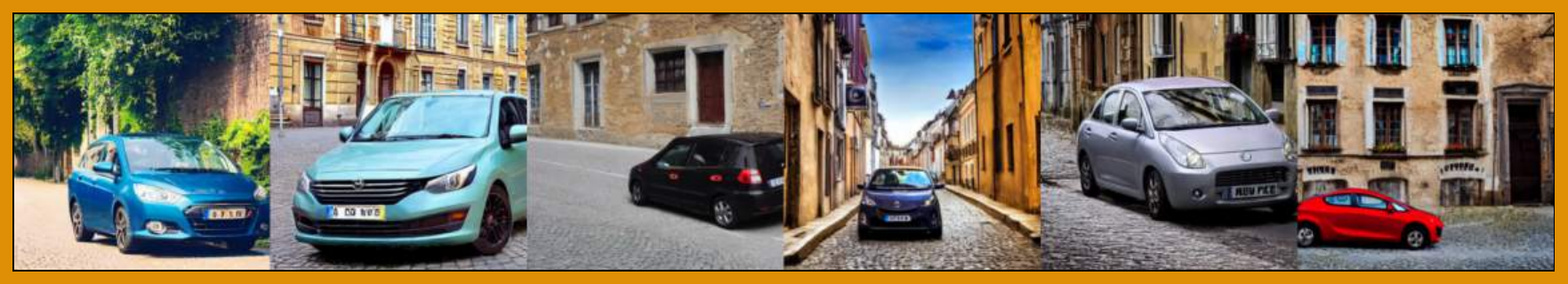}  \includegraphics[width=0.49\textwidth]{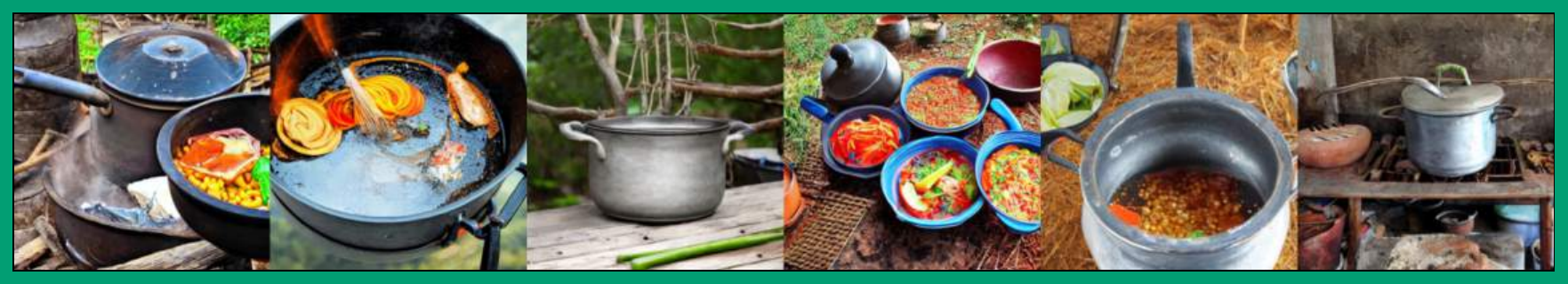}  \includegraphics[width=0.49\textwidth]{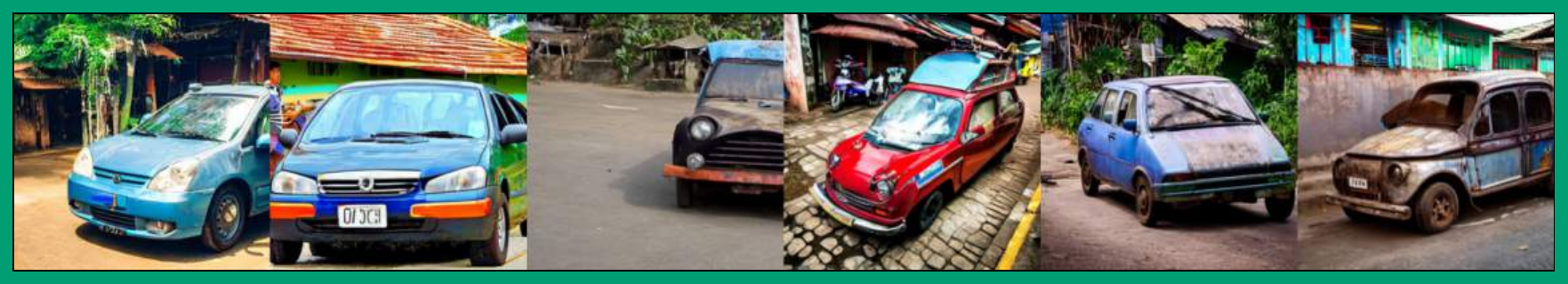}  

            \caption{No \methodnameshort, \ie \objreg baseline}
        \end{subfigure} 
        \begin{subfigure}[b]{\textwidth}
            \centering
            \includegraphics[width=0.49\textwidth]{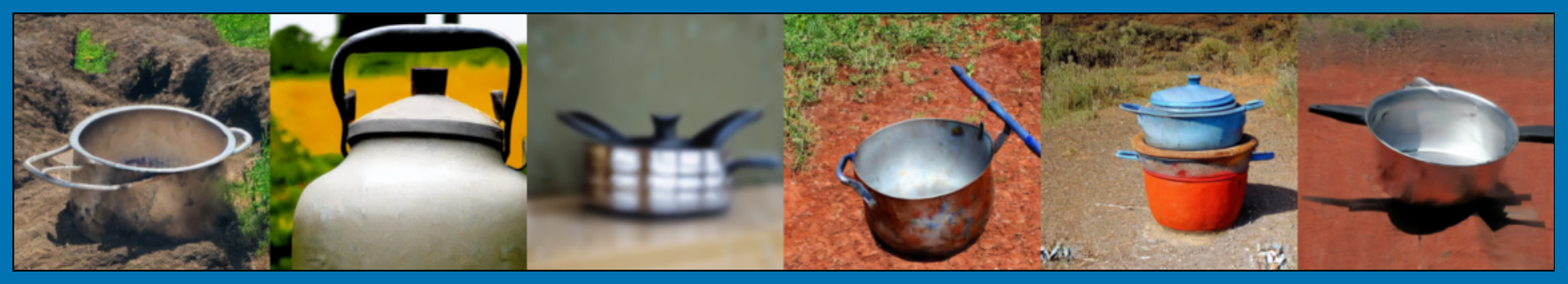}  
            \includegraphics[width=0.49\textwidth]{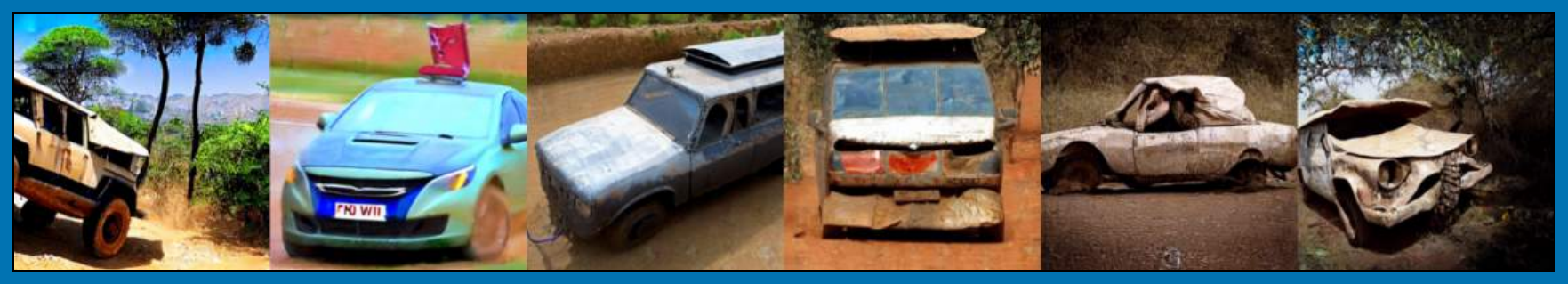}  
            \includegraphics[width=0.49\textwidth]{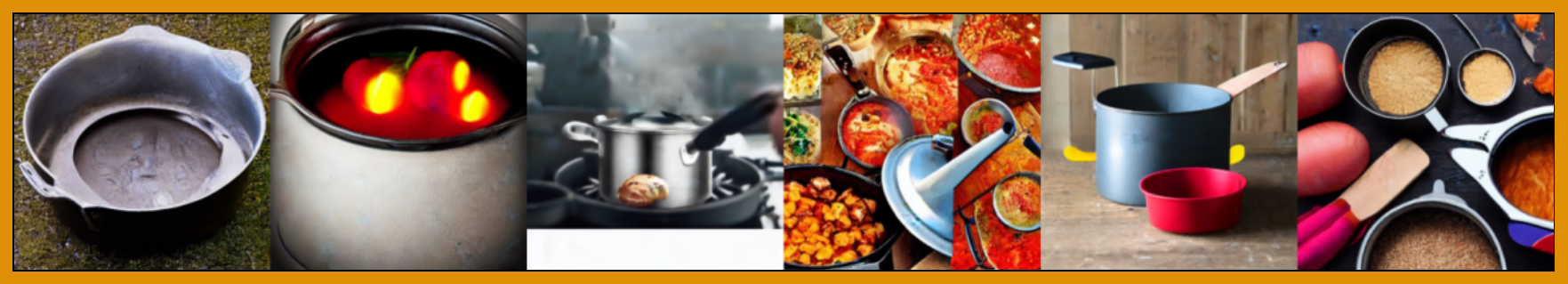}  
            \includegraphics[width=0.49\textwidth]{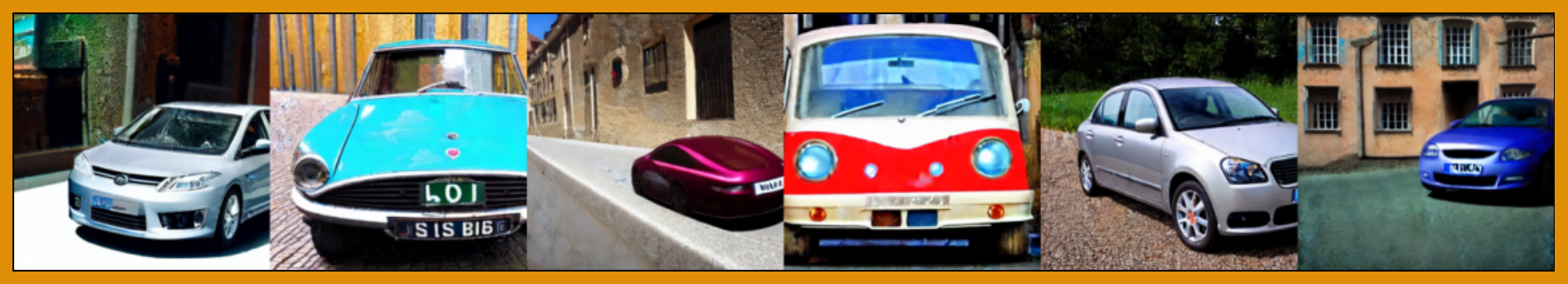}  
            \includegraphics[width=0.49\textwidth]{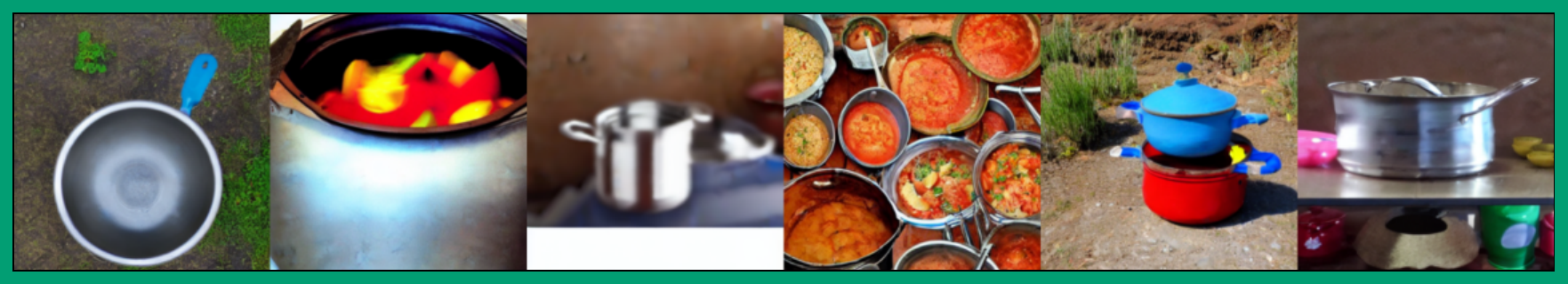}  
            \includegraphics[width=0.49\textwidth]{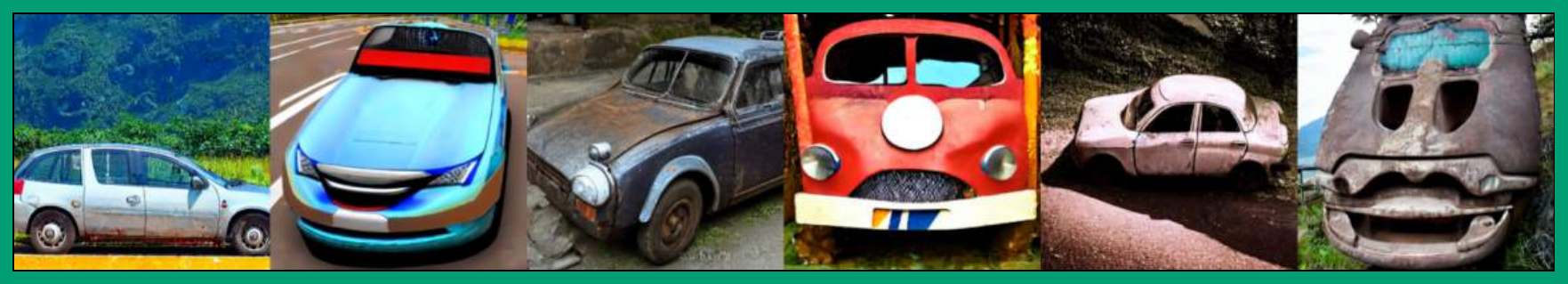}  
            \caption{ \methodname}
            \end{subfigure} 
            \begin{subfigure}[b]{\textwidth}
            \centering\includegraphics[width=0.49\textwidth]{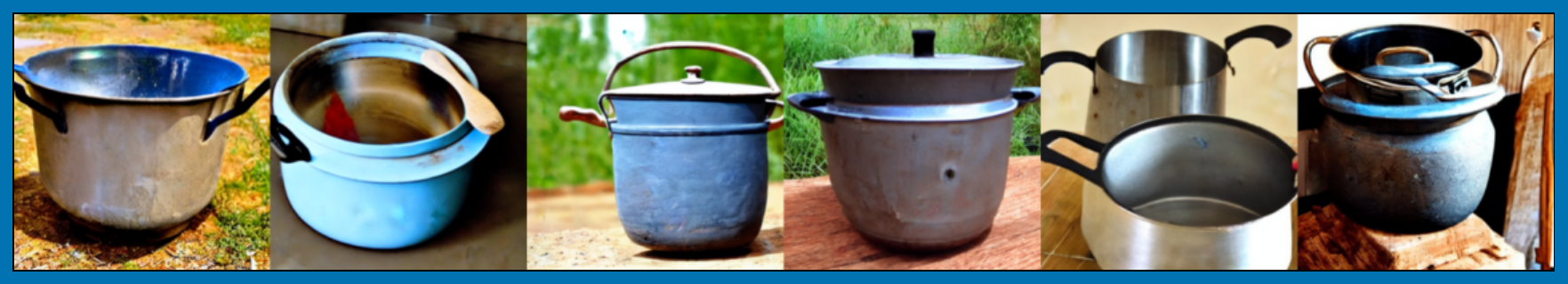}
            \includegraphics[width=0.49\textwidth]{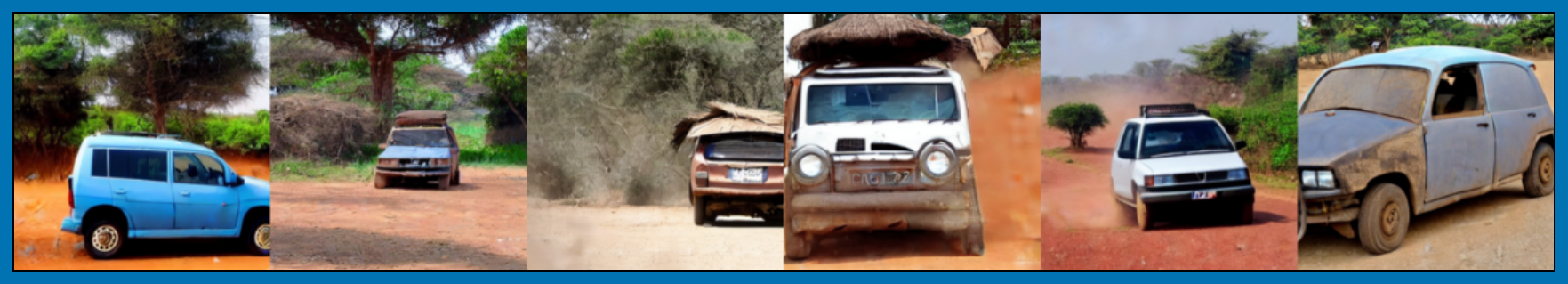}
            \includegraphics[width=0.49\textwidth]{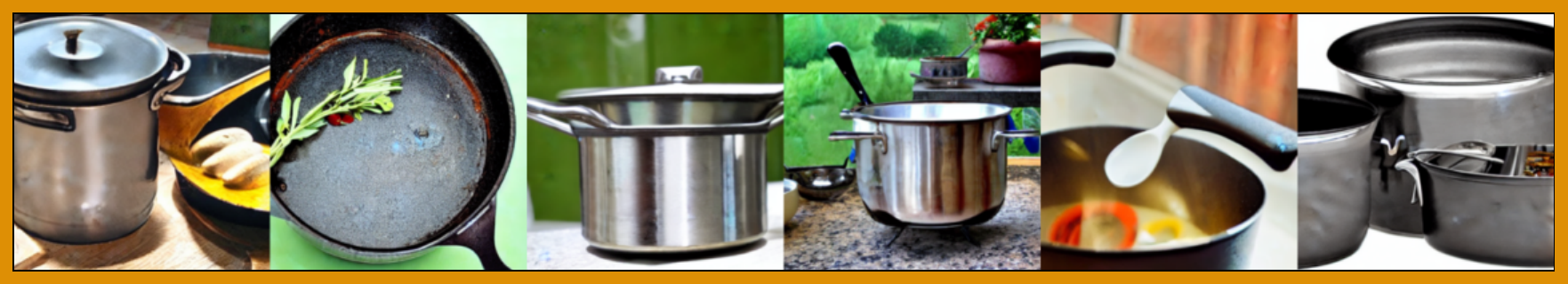}
            \includegraphics[width=0.49\textwidth]{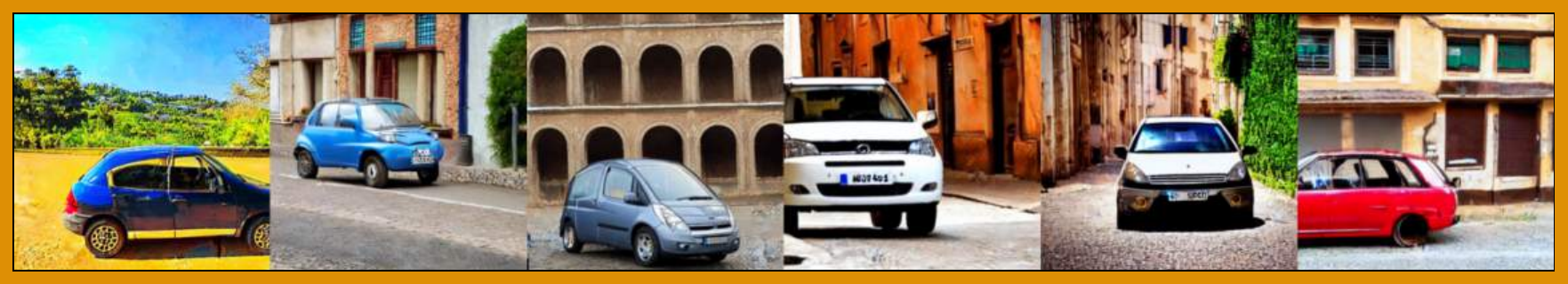}
            \includegraphics[width=0.49\textwidth]{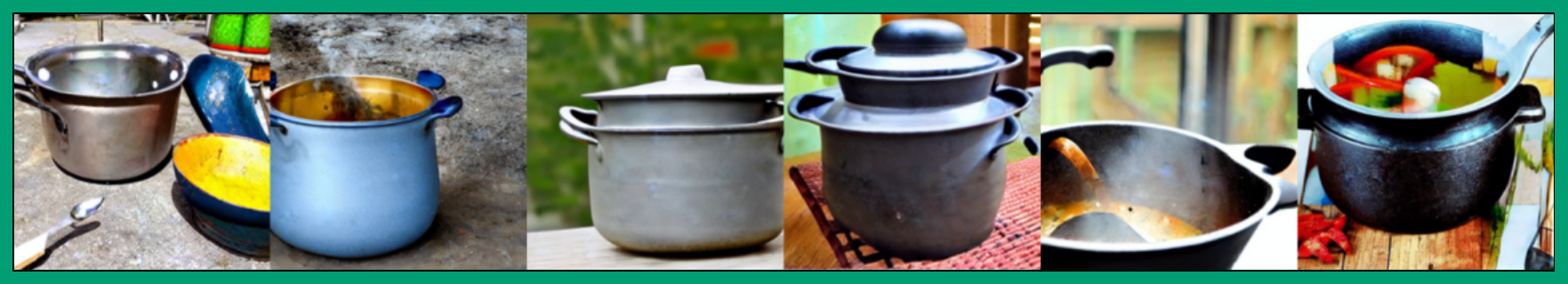}
            \includegraphics[width=0.49\textwidth]{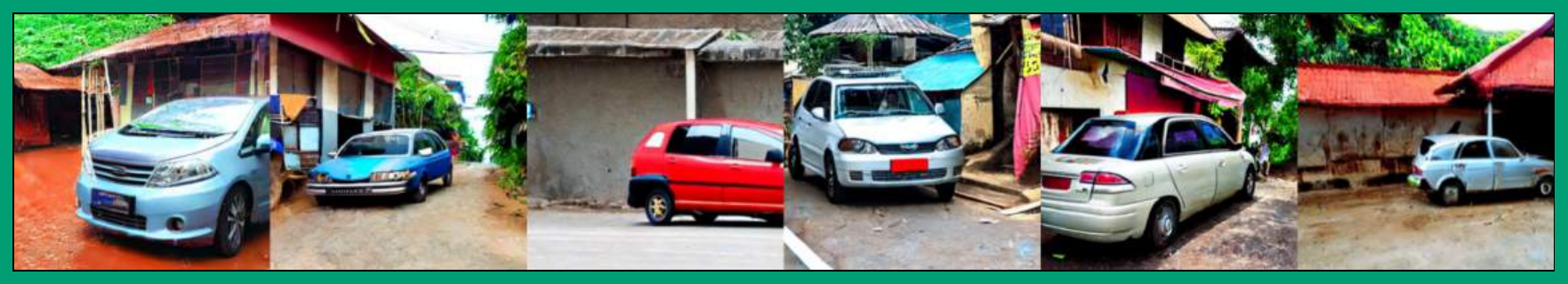}
            \caption{ Contextualized \methodname}
        \end{subfigure} 
    \end{subfigure}
    \caption{
    Generated images of cooking pots (\textbf{Left}) and cars (\textbf{Right}).
    The same six seeds are shared among the examples, and the box colors indicate images pertaining to \textcolor{blue}{Africa}, \textcolor{orange}{Europe}, and \textcolor{darkgreen}{Southeast Asia}.
    \methodname increases the diversity of generated images, including object type, positioning, and quality. 
    Contextualization with exemplar images increases similarity to real world diversity. 
    (More examples are shown in Appendix Figures \ref{fig:res_visuals_abalations} and \ref{fig:res_visuals_abalations_2}.)
    }
    \label{fig:res_visuals}
\end{figure}

\subsubsection{Overall results.} On GeoDE (reported in \cref{tab:geode_res}), \methodnameshort without contextualization achieves a \textbf{relative} improvement in average F1 of 9.6\% over generations obtained with the LDM (\methodnameshort achieves F1 of 0.399 and LDM achieves F1 of 0.364), an improvement over baselines that do not leverage any additional information.
Similarly, \methodnameshort yields relative improvement in worst-region F1 by 10.6\%. 
Interestingly, using Synonyms tends to decrease performance, perhaps due to its enforcement of a strict, yet limited, distribution of object names.

When considering methods that include additional information, we observe that leveraging textual descriptions of the evaluation dataset (Paraphrasing) is useful to improve average and worst-region F1. 
However, leveraging region information (Semantic Guidance, FG CLIP) yields better results. 
Notably, when the additional region information corresponds to \emph{known biases} embedded in text-to-image systems, as is the case of Semantic Guidance, we observe substantial relative performance improvements of 15.4\% and 24.5\% in average and worst-region F1 over the LDM baseline. 

Leveraging exemplar images to contextualize \methodnameshort results in further performance gains. In particular, c-\methodnameshort shows a 25\% and 37.9\% relative improvement in average and worst F1 compared to the LDM baseline. 
Comparing c-\methodnameshort with the closest competitor, Semantic Guidance, we observe a relative improvement of 8.3\% and 10.7\% in average and worst-region F1.

\cref{tab:ds_res} presents results on DollarStreet. In this case, uncontextualized \methodnameshort only moderately improves over the LDM baseline in average and worst-region F1. However, when including exemplar images per c-\methodnameshort, the patterns of improvement are consistent with GeoDE. c-\methodnameshort achieves 5.7\% and 4.5\% improvements over the closest baselines in average and worst-region F1, respectively.\looseness-1
 
In the reminder of this section, we analyze the findings described above in the context of improvements to diversity, quality, and consistency. 

\textbf{\methodnameshort improves diversity.}
The improvements in F1 relate to consistent boosting of both worst-region and average recall: for GeoDE (\cref{tab:geode_res}, recall), \methodnameshort both with and without contextualization show improvements for all baselines, and by at least 7.3\% (and up to 46.2\%) over the LDM  baseline. 
Similarly, \methodnameshort exhibits among the highest recall values in DollarStreet (\cref{tab:ds_res}, recall). 
FG CLIP (Ent.) achieves the highest recall values, but those come with significant drops in precision. 
Notably, for DollarStreet, \methodnameshort has better recall when exemplar images are \emph{not} used. Again, this comes at a cost to precision. 

Diversity improvements are also reflected in qualitative examples. \cref{fig:res_visuals} reveals that \methodnameshort leads to greater variation in object color, type, and shape, such as more colorful cars for Africa, sports-cars for Europe, and lengthened hoods for Southeast Asia. 
With \methodnameshort, image backgrounds also show greater variety in textures and diversity in indoor \vs outdoor settings. In addition, \methodnameshort enables more \textit{inter-region diversity}. 
For example, in~\cref{fig:res_visuals}, the same seed can look much more similar across regions in the baseline than with c-\methodnameshort.

\textbf{Contextualized \methodnameshort improves quality.}
In GeoDE, c-\methodnameshort also allows for improvement in both average and worst-region precision over the LDM baseline, the latter by up to 52.7\% (\cref{tab:geode_res}, precision). 
Similar precision gains are observed in DollarStreet (\cref{tab:ds_res}, precision): leveraging exemplar images results in relative improvements of 12\% in worst-region over the LDM baseline. 
Comparing c-\methodnameshort with \methodnameshort shows relative gains of 17.7\% and 25.3\% in average and worst-region precision, respectively. 
Interestingly, Semantic Guidance allows for higher precision than c-\methodnameshort on GeoDE, whose images were submitted by people living in the regions of interest. 
The lower precision achieved by Semantic Guidance on DollarStreet might be related to the datasets' focus on  disadvantaged and isolated areas. 
Thus, Semantic Guidance requires an understanding of whether applying region information helps or hurts the diversity of the generations. \looseness-1

\textbf{Contextualized \methodnameshort improves consistency.} In both GeoDE and DollarStreet, we observe that \methodnameshort without contextualization slightly reduces consistency (\cref{tab:geode_res} and~\ref{tab:ds_res}, CLIPScore). 
This highlights the importance of contextualization when using \methodnameshort: leveraging exemplar images in c-\methodnameshort results in the highest CLIPscores across the board, on average and for the worst-region.

\textbf{Contextualized \methodnameshort reduces disparities across regions.} When comparing c-\methodnameshort to \methodnameshort, we observe that disparities across regions appear substantially reduced on both datasets. On GeoDE, contextualization reduces the performance gap between average and worst-region in F1, recall, precision, and CLIPScore. Similar trends may be observed for DollarStreet, where c-\methodnameshort exhibits remarkably lower disparities across all metrics when compared to the LDM baseline. It is worth noting that Semantic Guidance also has low disparities.

\textbf{Contextualized \methodnameshort shows early improvements in region representation.}
While this work does not focus particularly on mitigation of reductive regional representations, we find in initial qualitative inspections that representations of regions can change positively after applying contextualized \methodnameshort. 
We find that for Africa, objects like cars and cooking pots are less dilapidated with \methodnameshort, as compared to the original generations.
In addition, some objects tend to be larger and more central in the image, reducing the prevalence of unrepresentative background fixtures, as shown in Appendix \ref{app:geode}. 
The improvement of consistency with contextualization similarly reduces the dominance of non-representative regional background information.
Remaining images without objects tend to have less repetition in reductive backgrounds associated with regional tropes. 
For example, Appendix \ref{app:geode} shows that generated images of bags in Europe can have more variety in their backgrounds when using c-\methodnameshort. 

\subsection{Ablations}
\label{sec:ablations}

We next discuss ablations on the variations of \methodnameshort criteria and  the strength, quantity, and type of exemplar images used in c-\methodnameshort. 

\begin{table}[b]
   \caption{Ablation study on variations of \methodnameshort criteria.}
   \label{tab:geode_res_ablate_criterion}
\centering
\resizebox{\textwidth}{!}{%
\setlength{\tabcolsep}{2pt}
\begin{tabular}{lccccccccccc}
\toprule
\textbf{Dataset}  &\textbf{$\alpha$} & \textbf{$\beta$} & \multicolumn{2}{c}{\textbf{F1}} & \multicolumn{2}{c}{\textbf{Precision}} & \multicolumn{2}{c}{\textbf{Recall}} & \multicolumn{2}{c}{\textbf{CLIPScore}}\\
\cmidrule(lr){4-5}
\cmidrule(lr){6-7}
\cmidrule(lr){8-9}
\cmidrule(lr){10-11}
& & & \text{Avg.} & \text{Worst-Reg.} & \text{Avg.} & \text{Worst-Reg.} & \text{Avg.} & \text{Worst-Reg.} & \text{Avg.} & \text{Worst-Reg.}\\
\midrule
    GeoDE  &0 & 0 & 0.364 & 0.322 & 0.413 & 0.273 & 0.337 & 0.395 & 0.242 & 0.218\\
    GeoDE &tuned &0 & 0.399 & 0.356 & 0.349 & 0.307 & 0.470 & 0.424 &  0.180 & 0.191\\
    GeoDE &0 &tuned & 0.446 & 0.427 & \textbf{0.431} & 0.409 & 0.464 & 0.447 & \textbf{0.261} & \textbf{0.254}\\
    GeoDE & tuned & tuned  & \textbf{0.455} & \textbf{0.444} & 0.424 & \textbf{0.417} & \textbf{0.493} & \textbf{0.476} & 0.254 & 0.253\\
    \midrule
    DollarStreet &0 & 0 & 0.448 &0.442 &0.428 & 0.434& 0.472 &0.450 & 0.231  &0.235 \\
    DollarStreet &tuned & 0 & 0.457 &0.444 & 0.413& 0.388& \textbf{0.516} & \textbf{0.518} &  0.191 & 0.198\\
    DollarStreet &0 & tuned & 0.492 & 0.480 & \textbf{0.486} & 0.483 &0.500 &0.476 & \textbf{0.240} & \textbf{0.242}\\
    DollarStreet  &tuned & tuned & \textbf{0.497} &\textbf{ 0.483} & \textbf{0.486} & \textbf{0.486}&0.511 &0.479 & {0.234} & {0.238}  \\
\bottomrule
\end{tabular}}%
\end{table}

\begin{table}[t]
\caption{Ablation study on exemplar images strength in contextualized \methodnameshort.}
   \label{tab:geode_res_ablate_strength}
\centering
\resizebox{\textwidth}{!}{%
\setlength{\tabcolsep}{2pt}
\begin{tabular}{lccccccccccc}
\toprule
\textbf{Dataset}  &\textbf{$\alpha$} & \textbf{$\beta$} & \multicolumn{2}{c}{\textbf{F1}} & \multicolumn{2}{c}{\textbf{Precision}} & \multicolumn{2}{c}{\textbf{Recall}} & \multicolumn{2}{c}{\textbf{CLIPScore}}\\
\cmidrule(lr){4-5}
\cmidrule(lr){6-7}
\cmidrule(lr){8-9}
\cmidrule(lr){10-11}
& & & \text{Avg.} & \text{Worst-Reg.} & \text{Avg.} & \text{Worst-Reg.} & \text{Avg.} & \text{Worst-Reg.} & \text{Avg.} & \text{Worst-Reg.}\\
\midrule
    GeoDE &tuned & $\alpha$ & {0.451} & 0.434 & 0.384 & 0.380&\textbf{ 0.547} & \textbf{0.506} & 0.216  & 0.221 \\
    GeoDE &tuned & 2$\alpha$ & \textbf{0.455} & \textbf{0.444} & {0.424} & 0.417 & {0.493} & {0.476} & 0.254 & 0.253\\
    GeoDE &tuned & 4$\alpha$ & 0.448 & 0.437 & \textbf{0.428}& \textbf{0.435} & 0.472 & 0.438 & \textbf{0.261} & \textbf{0.261}\\
\bottomrule
\end{tabular}}%
\end{table}

\textbf{Variations of c-\methodnameshort Criteria.}
We compare the use of the full contextualized Vendi Score Guidance criterion from our approach ($\alpha > 0$, $\beta > 0$, c-\methodnameshort) to Vendi Score guidance without contextualization ($\alpha > 0$, $\beta=0$, \methodnameshort), a criterion using the exemplar contextualizing images exclusively ($\alpha=0$, $\beta > 0$), and the LDM without intervention ($\alpha=0$, $\beta = 0$).
Results are shown in \cref{tab:geode_res_ablate_criterion} and visual examples in Appendix \ref{app:ablations}.

When evaluating with GeoDE, we find that Vendi Score guidance without contextualization can lead to images with improved diversity but unrealistic object shapes and colors, as well as extreme variation in styles, such as black-and-white or film-style photos.
While using exclusively contextualizing images helps in almost all aspects as opposed to only the Vendi Score guidance on the bank of generated images, the methods combined tend to have the best diversity. 
For DollarStreet, there is a slight trade-off in precision and recall between the inclusion of contextualizing images versus previously generated images in the criterion:
while the combination of the two has moderately better measures over only contextualizing images, the use of only previously generated images shows the highest recall.\looseness-1

\textbf{Strength of exemplar images.} \cref{tab:geode_res_ablate_strength} shows ablations of the strength of exemplar images used in the contextualized \methodnameshort criterion. 
Unsurprisingly, lower weight $\beta$ of real exemplar images yields greater recall while a higher weight leads to a larger precision, emphasizing the quality \vs diversity trade-off.
For example, increasing $\beta$ can correlate with changes in camera angle of cars to more closely match those in the reference images, as shown in Appendix \ref{app:ablations}.
In addition, consistency is improved with additional weight to exemplar images.

\textbf{Quantity and type of exemplar images.}
In \cref{tab:geode_ablate_num_real}, we first study how the number of images selected as exemplar images in  c-\methodnameshort affects image generation diversity. 
Generally, using fewer exemplar images tends to help with diversity while increasing this number helps with quality, although these trends appear minor in some cases. 
In Appendix \ref{app:ablations}, we visually inspect images and find that trends in quality and consistency for a single seed can also reverse as the number of images increases, first improving, then worsening (or vice versa). 
In addition, we study whether exemplar images randomly selected across all possible regions or the same number images selected specifically for a given region yields better diversity. 
We find that region-stratified exemplar images yield results within 1-2\% of those when using randomly selected images for all measures except for worst-region recall, for which there is a 5\% improvement.
\looseness-1

\begin{table}[t]
\caption{Ablation study on the number of exemplar images per object used in the contextualized Vendi Score computation, or $M$ in Algorithm \ref{alg:main}. Results are reported based on GeoDE dataset.}
  \label{tab:geode_ablate_num_real}
\centering
\resizebox{\textwidth}{!}{%
\setlength{\tabcolsep}{2pt}
\begin{tabular}{lccccccccc}
\toprule
{\textbf{Method}} & {\textbf{ \# imgs}} & \multicolumn{2}{c}{\textbf{F1}} & \multicolumn{2}{c}{\textbf{Precision}} & \multicolumn{2}{c}{\textbf{Recall}} & \multicolumn{2}{c}{\textbf{CLIPScore}}\\
\cmidrule(lr){3-4}
\cmidrule(lr){5-6}
\cmidrule(lr){7-8}
\cmidrule(lr){9-10}
&  & \text{Avg.} & \text{Worst Reg.} & \text{Avg.} & \text{Worst Reg.} & \text{Avg.} & \text{Worst Reg.} & \text{Avg.} & \text{Worst Reg.}\\
\midrule
    c-\methodnameshort  &\ 2 per obj. & 0.455 & \textbf{0.444} &0.424 & 0.417& \textbf{0.493} & \textbf{0.476} & 0.254 & 0.253 \\
    c-\methodnameshort  &\ 4 per obj. & 0.448 & 0.437 &0.428 & \textbf{0.435}&0.472 & 0.438 & 0.261 & \textbf{0.257} \\
    c-\methodnameshort  &\ 8 per obj.& 0.455 & 0.442 & \textbf{0.445} & 0.436&0.466 &0.449 & \textbf{0.262} & \textbf{0.257} \\
    c-\methodnameshort &\ 20 per obj. & \textbf{0.460} & 0.433 &0.440 & 0.434&0.485 &0.433 & 0.251 & 0.246 \\
    \midrule
    c-\methodnameshort  &\ 4 per obj. & 0.448 & 0.437 &0.428 & \textbf{0.435}&0.472 & 0.438& 0.261 & \textbf{0.257} \\
    c-\methodnameshort &\ 4 per obj.-reg. & \textbf{0.454} & \textbf{0.445} & \textbf{0.436} &0.431 & \textbf{0.475} &\textbf{0.459} & 0.259 & 0.253\\
\bottomrule
\end{tabular}}%
\end{table}

\section{Related Work}
\label{sec:related_work}

Prior work often focuses on mitigating \textit{general} diversity issues in text-to-image models.
For example, \cite{yu2020inclusive,rangwani2022improving} focus on preventing mode collapse while \cite{shin2023fill} generates a balanced distribution, focusing on the long-tail, by using synthetic images.
In addition, there is growing research focused on issues related to diversity through a lens of reducing \textit{person-related bias}.
One line of work reduces people-related biases, \eg related to gender, through concept forgetting or erasure \cite{heng2024selective,zhang2023forget,orgad2023editing}, rather than diversification. 
The idea is that, by modifying a model's understanding of a concept, the harmful or biased content is not generated. 
\cite{clemmer2024precisedebias} augments the original prompt with real-world distributions of gender and ethnicity using a finetuned language model, and
\cite{zhang2023iti,kim2023stereotyping} learn token embeddings for a diverse set of sensitive attributes and concatenate them to the original prompt. 
Building on work in NLP \cite{bolukbasi2016man}, other approaches remove the components of sensitive attributes like race and gender from the text embeddings altogether \cite{tanjim2024discovering,chuang2023debiasing} or modify the cross attention layer \cite{orgad2023editing,gandikota2023erasing} to remove implicit assumptions about the world. 
However, there is little work investigating improvements to \textit{geographic diversity}. 
To the best of our knowledge, prior work primarily discusses the (minimal) improvement in diversity when including more granular geographic information in text prompts \cite{basu2023inspecting,hall2023dig}.

Our proposed method is most relevant to works that use classifier guidance during inference time. 
\cite{brack2023mitigating,schramowski2023safe} use classifier guidance to guide generations away from inappropriate content. 
In the context of augmenting a dataset with synthetic samples, ~\cite{hemmat2023feedbackguided} has extended the idea of classifier guidance to feedback guidance where loss and entropy are used as the guidance mechanism. 
While the Vendi Score has been used at training time to increase the diversity of generative adversarial networks (GANs) across modes in the training dataset \cite{berns2023mode}, to the best of our knowledge, we are the first to apply it to diffusion models and as guidance at inference time.\looseness-1

\section{Conclusion}
\label{sec:conclusion}

In this work, we introduced an inference time intervention that extends the guidance toolbox of diffusion models by driving the generation process towards samples that are substantially different from each other while still representative of the real world. 
Through extensive experiments, we showed that our approach produces images with higher intra- and inter-region diversity, while exhibiting increased image quality and text-image consistency; overall resulting in reduced disparities in generation diversity, quality and consistency.\looseness-1

\noindent\textbf{Limitations.} 
In this work, we focus on improving worst-region F1 while also analyzing worst and average region precision, recall, and consistency. 
Automatic metrics are susceptible to several challenges, including region representations in the reference dataset, reliance on pre-existing feature extractors, and the composition of geographic groupings. 
The metrics are aggregates and do not account for individual preferences. A human evaluation study would be necessary for capturing subjective and personal perspectives about the effect of  c-\methodnameshort .
Finally, our approach is only a single effort at improving the diversity of text-to-image generative models, and there remains future work for their further improvement.\looseness-1 

\noindent\textbf{Societal Impacts.} 
This work involves representations of geographic regions.
While our qualitative analyses show that c-\methodnameshort tends to help mitigate reductive representations of regions, our method may not always remove harmful representations in the underlying generative models.
Furthermore, increasing the diversity of text-to-image models may yield unexpected generations.
Thorough study of the full range of potential outcomes is necessary before deployment of these methods in real world systems.\looseness-1

\section*{Acknowledgments}
We would like to thank Oscar Mañas and Pietro Astolfi for helpful discussion and pointers regarding in-context learning and paraphrasing approaches. We also thank Mohammad Pezeshki for useful discussions, feedback and support. In addition, we thank the participants of the qualitative study of c-\methodnameshort, discussed in Appendix \ref{human_study}.

\newpage

%
%
\bibliographystyle{splncs04}
\bibliography{main}
\clearpage
\section{Appendix}
\label{sec:appendix}

\subsection{Set-up}
\label{sec:app_setup}

\begin{figure}[!ht]
    \centering
    \begin{subfigure}[b]{0.23\textwidth}
        \includegraphics[width=0.9\textwidth]{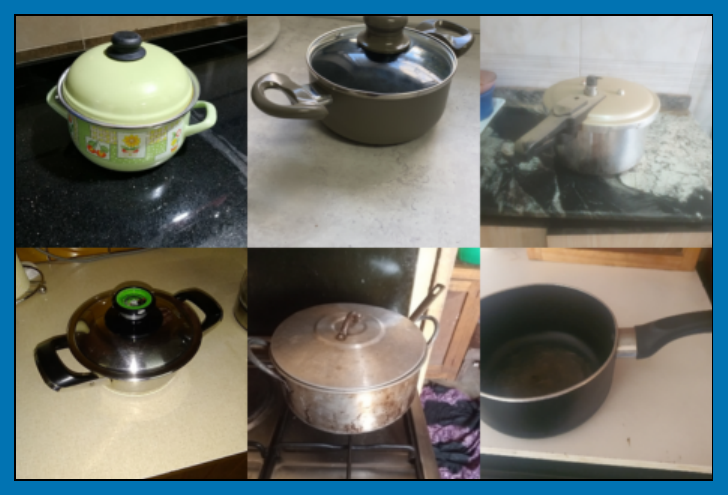}
        \includegraphics[width=0.9\textwidth]{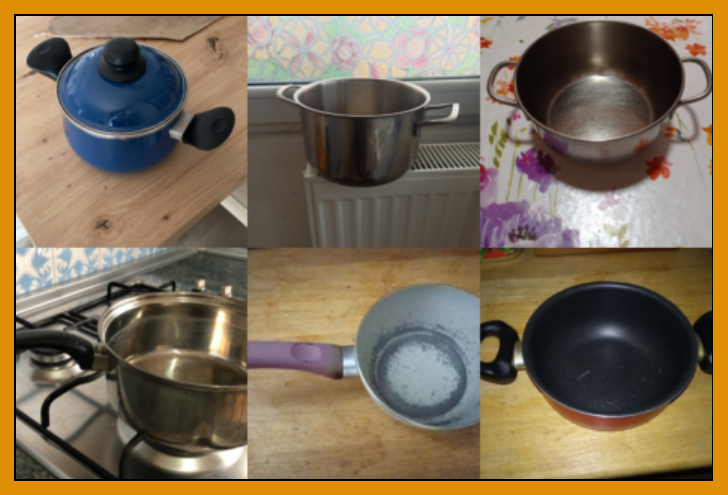}
        \includegraphics[width=0.9\textwidth]{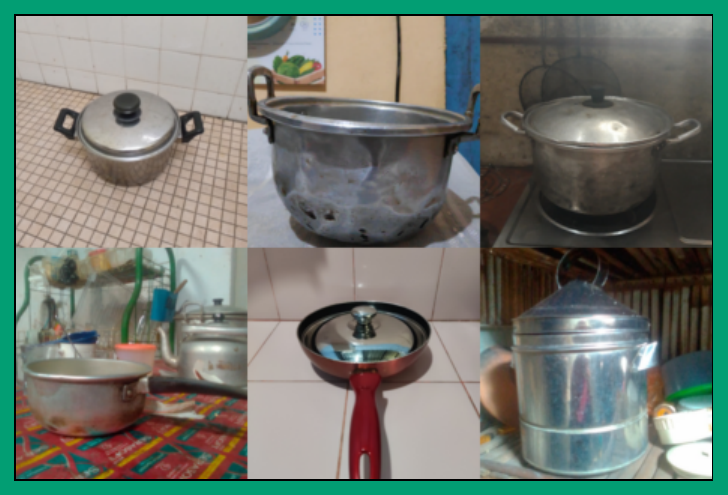}
        \caption{Real: cooking pot}
     \end{subfigure}
    \begin{subfigure}[b]{0.23\textwidth}
        \includegraphics[width=0.9\textwidth]{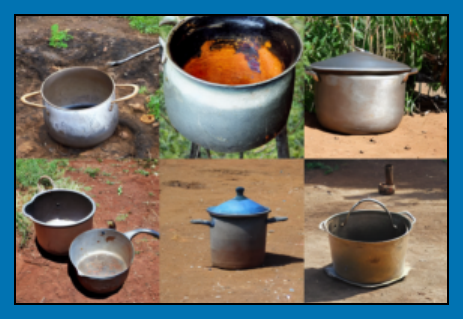}
        \includegraphics[width=0.9\textwidth]{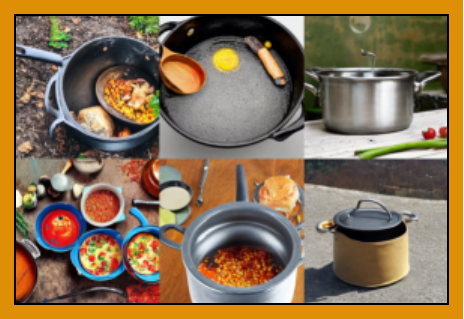}
        \includegraphics[width=0.9\textwidth]{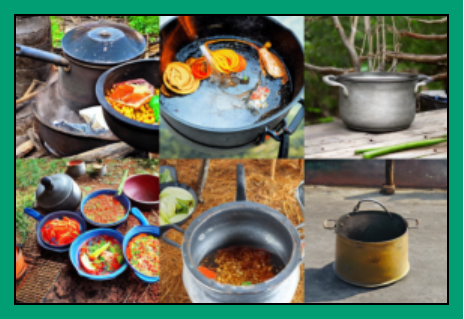}
        \caption{Gen: cooking pot}
     \end{subfigure}
     \begin{subfigure}[b]{0.23\textwidth}
        \includegraphics[width=0.9\textwidth]{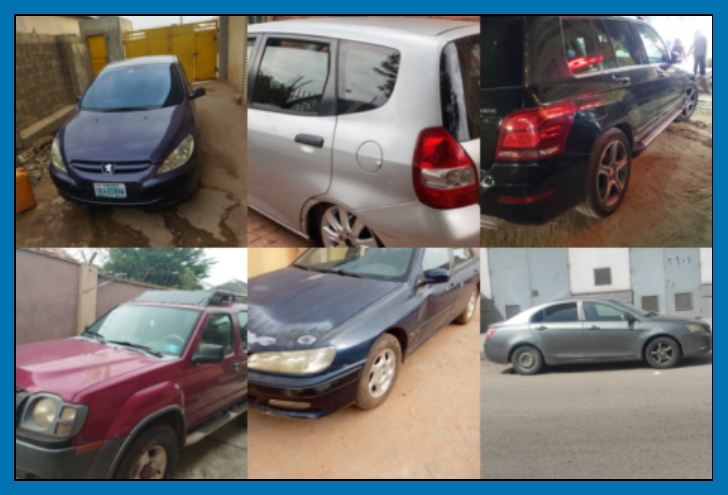}
        \includegraphics[width=0.9\textwidth]{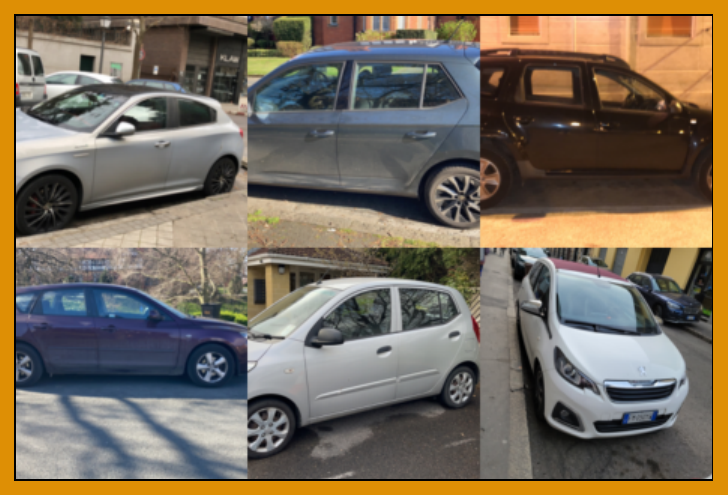}
        \includegraphics[width=0.9\textwidth]{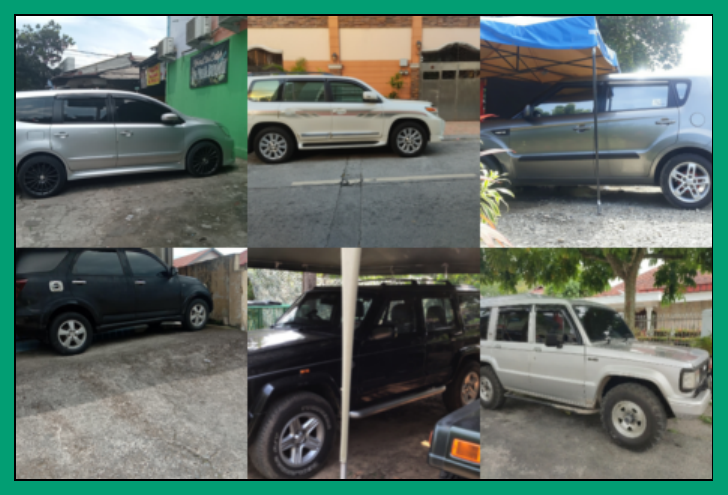}
        \caption{Real: car}
     \end{subfigure}
    \begin{subfigure}[b]{0.23\textwidth}
        \includegraphics[width=0.9\textwidth]{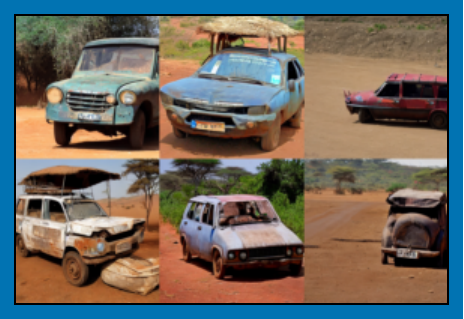}
        \includegraphics[width=0.9\textwidth]{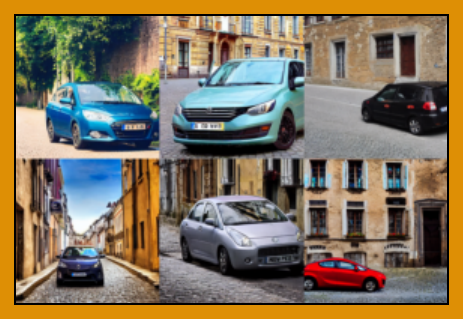}
        \includegraphics[width=0.9\textwidth]{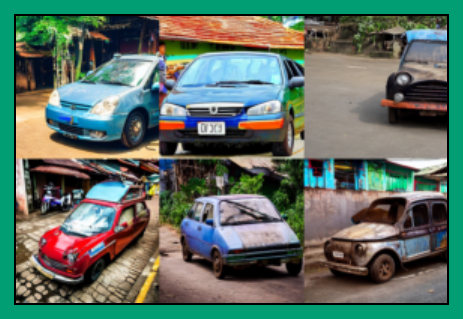}
        \caption{Gen: car}
     \end{subfigure}
         \begin{subfigure}[b]{0.23\textwidth}
        \includegraphics[width=0.9\textwidth]{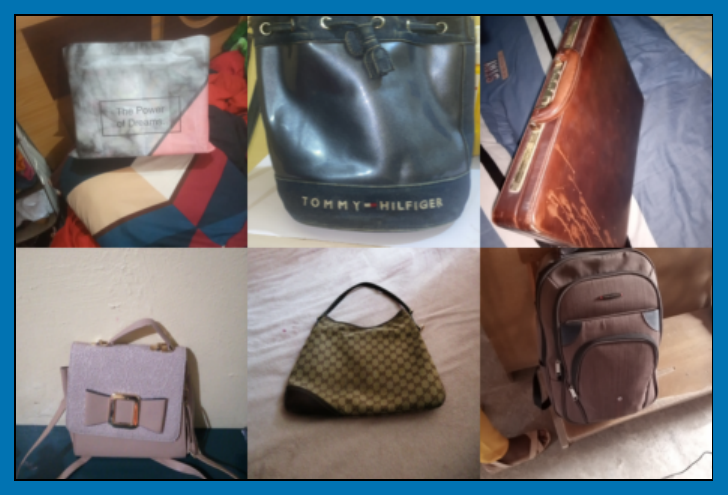}
        \includegraphics[width=0.9\textwidth]{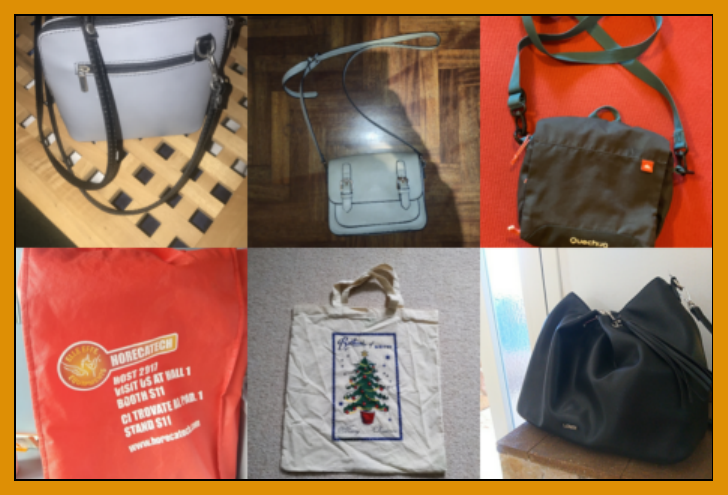}
        \includegraphics[width=0.9\textwidth]{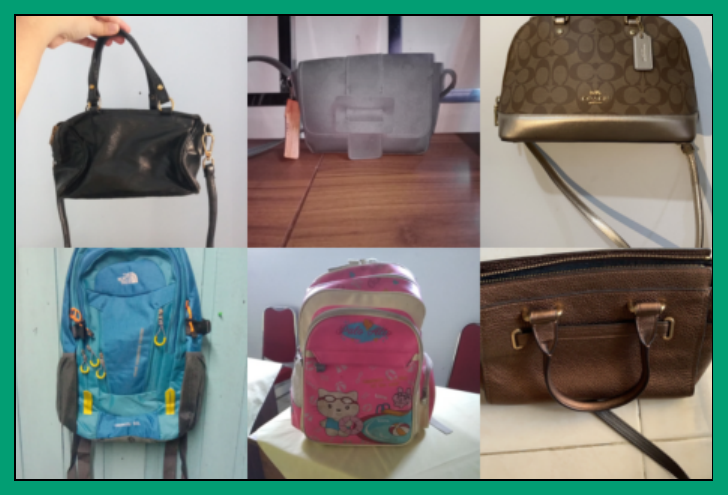}
        \caption{Real: bag}
     \end{subfigure}
    \begin{subfigure}[b]{0.23\textwidth}
        \includegraphics[width=0.9\textwidth]{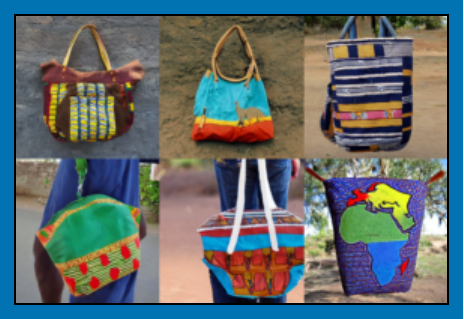}
        \includegraphics[width=0.9\textwidth]{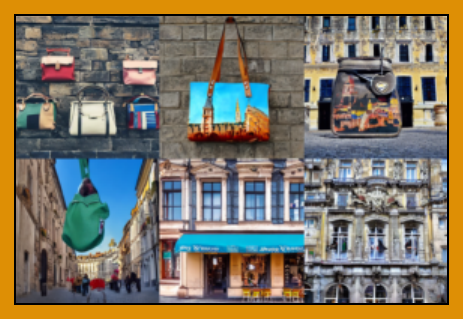}
        \includegraphics[width=0.9\textwidth]{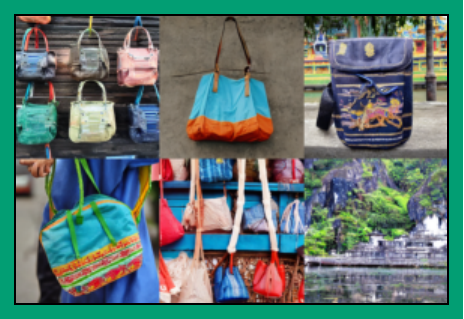}
        \caption{Gen: bag}
     \end{subfigure}
     \begin{subfigure}[b]{0.23\textwidth}
        \includegraphics[width=0.9\textwidth]{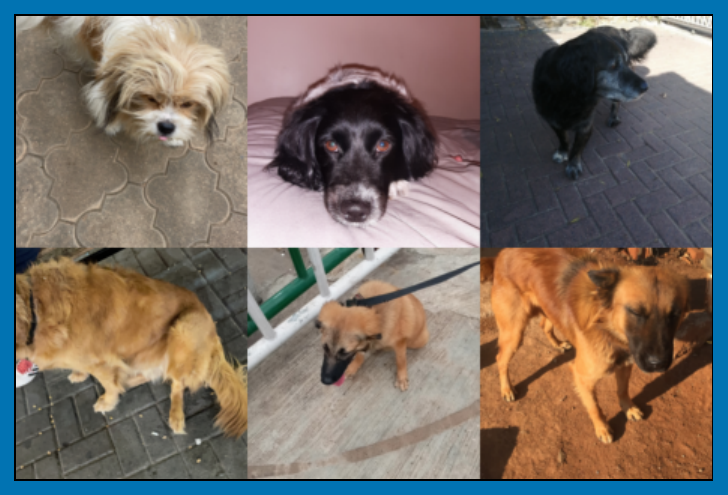}
        \includegraphics[width=0.9\textwidth]{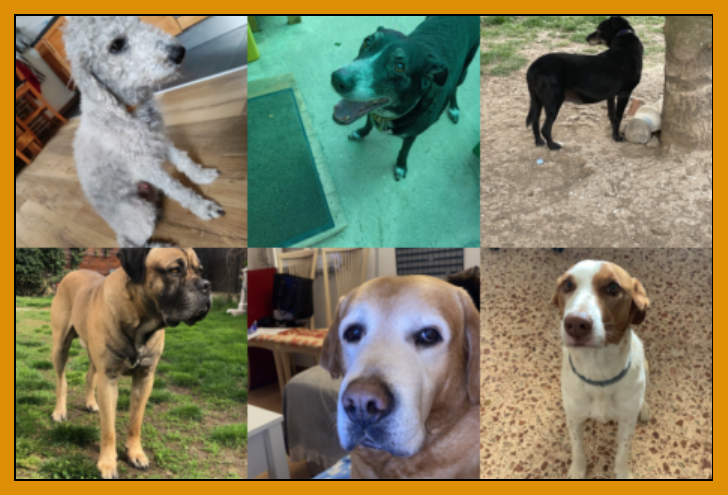}
        \includegraphics[width=0.9\textwidth]{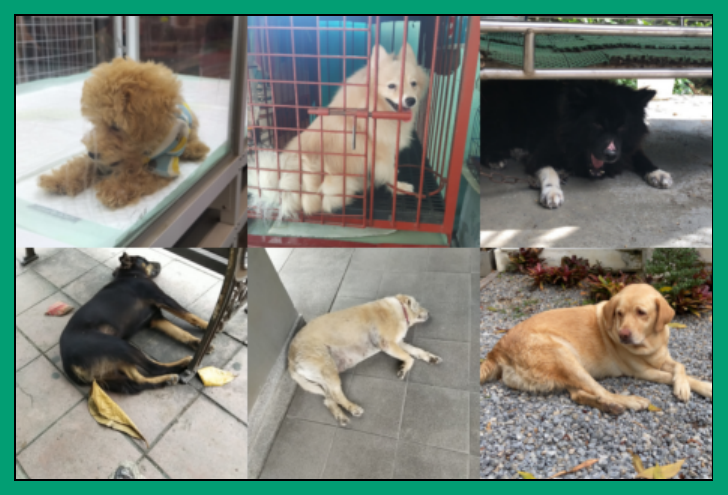}
        \caption{Real: dog}
     \end{subfigure}
    \begin{subfigure}[b]{0.23\textwidth}
        \includegraphics[width=0.9\textwidth]{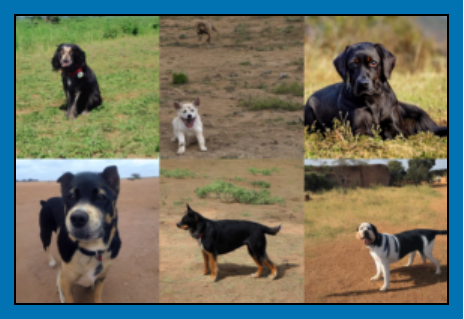}
        \includegraphics[width=0.9\textwidth]{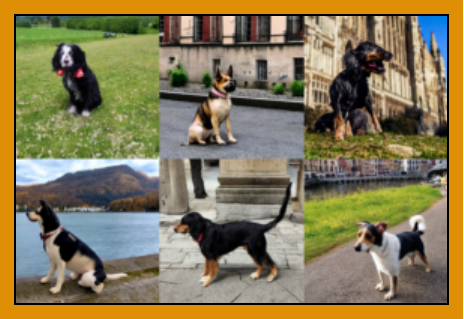}
        \includegraphics[width=0.9\textwidth]{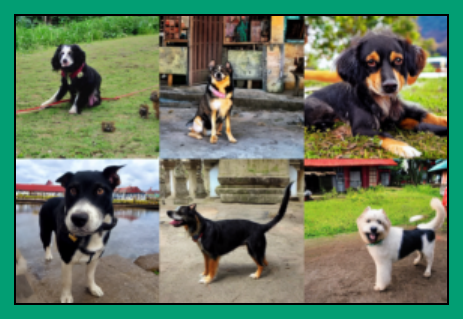}
        \caption{Gen: dog}
     \end{subfigure}
    \caption{Examples of images from the real world reference dataset, GeoDE\cite{ramaswamy2022geode}, and images generated with the original LDM using the prompt \texttt{object in region}, for example \texttt{cooking pot in Africa}. 
    Generated images of objects lack diversity compared to real world and introduce some region level dependencies in object depiction not seen in GeoDE, such as dilapidated cars for Africa.
    The colors indicate images pertaining to \textcolor{blue}{Africa}, \textcolor{orange}{Europe}, and \textcolor{darkgreen}{Southeast Asia}.}
    \label{fig:diversity_issue_example}
\end{figure}

Figure \ref{fig:diversity_issue_example} shows example images from the real, GeoDE dataset and generated images from the original LDM using the prompt \objreg.

\subsubsection{Synonym Baseline}

In Table \ref{tab:synonyms}, we show the WordNet lemmas associated with each original GeoDE class. 
For the Synonym baseline, we stratify the generations evenly across the different synonyms, including the original class name.

\begin{table}[]
\resizebox{\textwidth}{!}{%
\begin{tabular}{ll}
\textbf{Original Class} & \textbf{Synonym List}                                                          \\
bag           & `bag', `grip', `handbag', `pocketbook', `purse', `suitcase', `traveling bag', `travelling bag'                           \\
bicycle                 & `bicycle', `bike', `cycle', `wheel'                                            \\
candle                  & `candle', `taper', `wax light'                                                 \\
car                     & `car', `auto', `automobile', `machine', `motorcar'                             \\
chair                   & `chair'                                                                        \\
cleaning equipment      & `cleaning equipment', `cleaning device', `cleaning implement'                  \\
cooking pot             & `cooking pot', `cooking utensil', `cookware'                                   \\
dog                     & `dog', `canis familiaris', `domestic dog'                                      \\
dustbin       & `dustbin', `ash bin', `ash-bin', `ashbin', `ashcan', `garbage can', `trash barrel', `trash bin', `trash can', `wastebin' \\
front door              & `front door', `door'                                                           \\
hairbrush comb          & `hairbrush comb', `comb', `hairbrush'                                          \\
hand soap               & `hand soap', `soap dispenser'                                                  \\
hat                     & `hat', `chapeau', `lid'                                                        \\
jug                     & `jug'                                                                          \\
light fixture           & `light fixture', `source of illumination'                                      \\
light switch            & `light switch', `electric switch', `electrical switch', `switch'               \\
lighter                 & `lighter', `igniter', `ignitor', `light'                                       \\
medicine                & `medicine', `multivitamin', `multivitamin pill', `pill bottle', `vitamin pill' \\
plate of food & `plate of food', `aliment', `alimentation', `nourishment', `nutriment', `nutrition', `sustenance', `victuals'            \\
spices                  & `spices', `spice'                                                              \\
storefront              & `storefront', `display window', `shop window', `shopwindow', `show window'     \\
stove                   & `stove', `cooking stove', `kitchen range', `kitchen stove', `range'            \\
toothbrush              & `toothbrush'                                                                   \\
toothpaste              & `toothpaste toothpowder'                                                       \\
toy                     & `toy'                                                                          \\
tree                    & `tree'                                                                         \\
waste container         & `waste container', `dumpster'                                                 
\end{tabular}
}%
\caption{Synonym list for GeoDE}
\label{tab:synonyms}
\end{table}

\subsubsection{LLM Paraphrasing Baseline}

We include additional details about the LLM paraphrasing baseline. 

To determine how best to prompt the model, we started with three themes of prompts.
The first included only a verbal description of what we're trying to optimize, \ie ``increases the diversity of objects depicted in the image.''
The second included additional guardrails, \ie ``Generated images using your proposed template should contain objects that are realistic and representative of different geographic regions, with no amplification of geographic stereotypes. The images should always contain the desired object.''
For the third, we included specific descriptive criteria found in documentation of the reference datasets used for evaluation. 

We experimented with various phrasings similar to those in \cite{mañas2024improving} to ensure that the template paraphrasing includes both ``\{object\}'' and ``\{region\}'' tags and is compatible with our object names and regions. 
For the LLM, we used LLaMA-2-70B-chat \cite{touvron2023llama}, with a temperature of 1.0.

In Table \ref{tab:paraphrase} we include the best metaprompt for the three types of prompts described above and the model responses. 
We generated test images for each of the new prompt templates and found that the third theme, specific descriptive criteria, yields the highest precision and recall. 
We used that theme for the prompts reported in the main text of the paper.

\begin{table}[]
\caption{}
\resizebox{\textwidth}{!}{%
\begin{tabular}{l|l|ll}
\multicolumn{1}{c}{\textbf{Prompt Type}} &
  \multicolumn{1}{c}{\textbf{Metaprompt}} &
  \multicolumn{1}{c}{\textbf{LLM Response}} &
   \\ \midrule \midrule
\begin{tabular}[c]{@{}l@{}}Verbal description\\ of metric we're \\ trying to optimize.\end{tabular} &
  \begin{tabular}[c]{@{}l@{}}You are an expert prompt optimizer for text-to-image models. \\ Text-to-image models take a text prompt as input and generate images \\ depicting the prompt as output. You translate prompts written by \\ humans into better prompts for the text-to-image models. \\ Your answers should be concise and effective. \\ \\ Your task is to optimize this prompt template written by a human: \\ "{object} in {region}". This prompt template is used to generate many \\ images of objects such as bags, cars, and cooking pots in \\ regions such as Africa and Europe. \\ \\ Generate one paraphrase of the initial prompt template which contains \\ the keywords "{object}" and "{region}" but increases the diversity \\ of the objects depicted in the image. Only one object should be shown \\ per image.\end{tabular} &
  \begin{tabular}[c]{@{}l@{}}Sure, here's a paraphrased prompt template that could potentially increase the \\ diversity of objects depicted in the image:\\ \\ "A lone {object} stands proudly against the backdrop of {region}'s unique \\ landscape."\\ \\ This revised prompt still incorporates the keywords "{object}" and "{region}", \\ but it adds a new element of the object being the central focus of the image, set \\ against a distinctive landscape that highlights the region's characteristics. This \\ could encourage the model to generate a wider range of objects, such as a solitary\\  tree in the African savannah, a windmill in the Dutch countryside, or a fishing \\ boat  on the coast of Maine.\end{tabular} &
   \\ \midrule
\begin{tabular}[c]{@{}l@{}}Verbal description\\ of metric we're \\ trying to optimize \\ + guardrails.\end{tabular} &
  \begin{tabular}[c]{@{}l@{}}You are an expert prompt optimizer for text-to-image models. \\ Text-to-image models take a text prompt as input and generate images\\ depicting the prompt as output. You translate prompts written by \\ humans into better prompts for the text-to-image models. \\ Your answers should be concise and effective. \\ \\ Your task is to optimize this prompt template written by a human: \\ "{object} in {region}". This prompt template is used to generate many \\ images of objects such as bags, cars, and cooking pots in \\ regions such as Africa and Europe. \\ \\ Generate one paraphrase of the initial prompt template which contains \\ the keywords "{object}" and "{region}" but increases the diversity \\ of the objects depicted in the image. Only one object should be shown\\  per image. Generated images using your proposed template should \\ contain objects that are realistic and representative of different \\ geographic regions, with no amplification of geographic \\ stereotypes. The images should always contain the desired object.\end{tabular} &
  \begin{tabular}[c]{@{}l@{}}Sure, I can help you with that! Here's a paraphrased prompt template that could \\ potentially increase the diversity of objects depicted in the image while still \\ containing the keywords "{object}" and "{region}":\\ \\ "A lone {object} stands proudly against the backdrop of {region}, symbolizing \\ the rich cultural heritage of its people."\\ \\ This revised prompt template incorporates the same keywords, but with a\\  slightly different context. By using the phrase "a lone {object}", we're more \\ likely to see a diverse range of objects in the generated images, as the model \\ is encouraged to depict a single, prominent object rather than a collection of\\  objects. Additionally, the phrase "symbolizing the rich cultural heritage of its \\ people" adds a sense of significance and meaning to the object, which could \\ help to avoid reinforcing geographic stereotypes.\\ \\ For example, using this paraphrased prompt template, we might generate \\ images like:\\ \\ * A traditional African wooden mask standing on a rocky outcrop overlooking \\ a vast savannah in Kenya.\\ * A colorful Indian saree draped over a bamboo shoulder in a\end{tabular} &
   \\ \midrule
\begin{tabular}[c]{@{}l@{}}Verbal description \\ matching \\ GeoDE\end{tabular} &
  \begin{tabular}[c]{@{}l@{}}You are an expert prompt optimizer for text-to-image models. \\ Text-to-image models take a text prompt as input and generate images \\ depicting the prompt as output. You translate prompts written by \\ humans into better prompts for the text-to-image models. \\ Your answers should be concise and effective. \\ \\ Your task is to optimize this prompt template written by a human: \\ "{object} in {region}". This prompt template is used to generate many \\ images of objects such as bags, cars, and cooking pots in \\ regions such as Africa and Europe. \\ \\ Generate one paraphrase of the initial prompt template which contains \\ the keywords "{object}" and "{region}" but is optimized for the \\ following criteria:\\ 1. A single object per image.\\ 2. A well-lit environment and the object is clearly visible in the photos.\\ 3. The object occupies at least 25\% of the image.\\ 4. Objects captured are foregrounded and not occluded.\\ 5. Objects should not be blurred, e.g., motion blur.\\ 6. No effects or filters added (cropping is acceptable).\end{tabular} &
  \begin{tabular}[c]{@{}l@{}}Sure! Here's an optimized prompt template that meets the given criteria:\\ \\ "A well-lit {object} stands alone in a spacious {region}, occupying at\\ least 25\% of the frame. The object is stationary and not occluded, and \\ its details are clearly visible with no motion blur or added filters. \\ The image can be cropped to emphasize the object's presence and isolation."\\ \\ Paraphrased prompt examples using the optimized template:\\ \\ * A well-lit cooking pot stands alone on the barren ground of Africa, \\ occupying at least 25\% of the frame.\\ * A black modern car is on a highway in Europe, not occluded, and \\ its details are clearly visible with no motion blur or added filters.\\ This paraphrased prompt template puts emphasis on the object's isolation,\\  clarity, and lack of motion blur, which helps in generating images that \\ match the criteria. Using "stands alone" and "not occluded" ensures \\ that the object is the main focus of the image and is not obstructed by \\ other elements. "A spacious {region}" suggests\end{tabular} &
   \\ \midrule
\begin{tabular}[c]{@{}l@{}}Verbal description \\ matching \\ DollarStreet\end{tabular} &
  \begin{tabular}[c]{@{}l@{}}You are an expert prompt optimizer for text-to-image models. \\ Text-to-image models take a text prompt as input and generate images \\ depicting the prompt as output. You translate prompts written by \\ humans into better prompts for the text-to-image models. \\ Your answers should be concise and effective. \\ \\ Your task is to optimize this prompt template written by a human: \\ "{object} in {region}". This prompt template is used to generate many \\ images of objects such as bags, cars, and cooking pots in \\ regions such as Africa and Europe. \\ \\ Generate one paraphrase of the initial prompt template which contains \\ the keywords "{object}" and "{region}" but is optimized for the \\ following criteria:\\ 1. A single object per image.\\ 2. An accurate portrayal of everyday life. \\ 3. A wide representation of images, including low- and middle-income \\ households and disadvantaged and isolated areas. \end{tabular} &
  \begin{tabular}[c]{@{}l@{}}Sure! Here's a paraphrase of the initial prompt template that \\ meets your criteria: \\ \\ "Everyday object in a real-world {region} setting"\\ \\ This revised prompt still incorporates the essential keywords \\ of the original template but with a twist that meets your criteria. \\ By emphasizing "everyday objects" and "real-world settings," the \\ prompt encourages the model to generate images depicting objects \\ in their actual context and use, providing a more authentic representation \\ of everyday life. It also implies that the generated images should be\\ diverse, including low- and middle-income households and \\ disadvantaged and isolated areas, to provide a wider representation \\ of real-world settings.\\ Additionally, this paraphrased prompt encourages the model \\ to focus on a single object per image, ensuring that the generated \\ images are clear and focused, with a single dominant object or feature. \\ By using this optimized prompt, you can generate images that \\ accurately depict real-world scenarios while showcasing diverse \\ perspectives and objects, making it a great starting point for your \\ text-to-image model.\end{tabular} &
  
\end{tabular}
}%
\label{tab:paraphrase}
\end{table}

\subsection{Results and implementation details}
We discuss additional results for GeoDE and DollarStreet, as well as ablations in this section. Furthermore, we discuss how to implement the VS in a differentiable way to implement VSG.

\subsubsection{Implementation of Differentiable Vendi Score.}\label{app:dvs_derivation}
We note that original Vendi Score formulation in Eq. \ref{eq:vendi-score} is not differentiable since it requires finding the singular values of the similarity matrix and singular value decomposition is not differentiable.
We simply start by applying the singular value decomposition on the similarity matrix, $\mathcal{K}$,
\begin{equation}
 U, \Sigma', V^T = \text{SVD}(\mathcal{K}),
\end{equation}
where $\Sigma'$ are the singular values and $U$ and $V$ are unitary matrices containing the left and right singular vectors of the matrix $\mathcal{K}$, respectively. We then discard $\Sigma'$ (because back-propagation through it is not possible) and find the singular values by the following differentiable operation:
\begin{equation}
    \Sigma = U^T \mathcal{K} V.
\end{equation}
While $\Sigma'$ and $\Sigma$ are equal, only the latter is differentiable with respect to $x$. Note that $\Sigma$ is the square root of the eigenvalues of $\mathcal{K}$.

\subsubsection{Additional Results}\label{app:geode}

\begin{figure}
    \centering
    \begin{subfigure}[b]{0.49\textwidth}
        \includegraphics[width=0.99\textwidth]{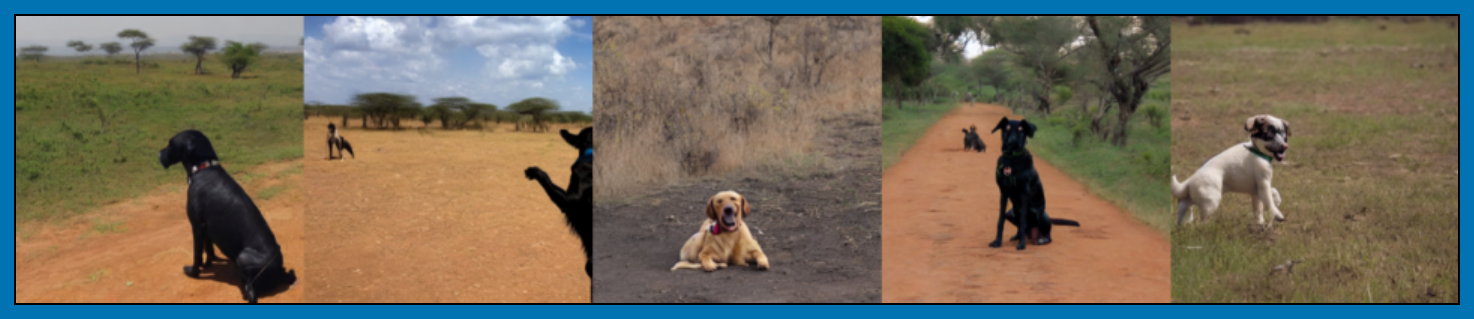}
         \includegraphics[width=0.99\textwidth]{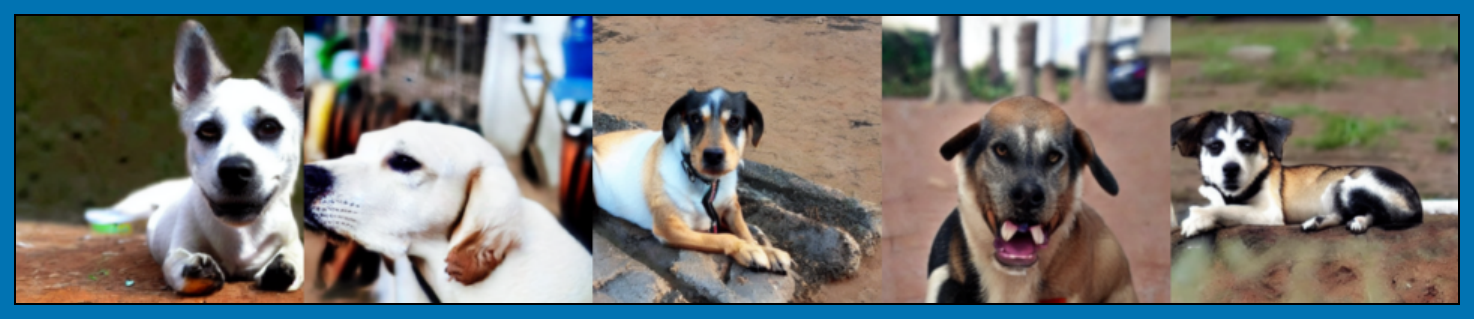}
         \caption{Africa}
    \end{subfigure}
        \begin{subfigure}[b]{0.49\textwidth}
        \includegraphics[width=0.99\textwidth]{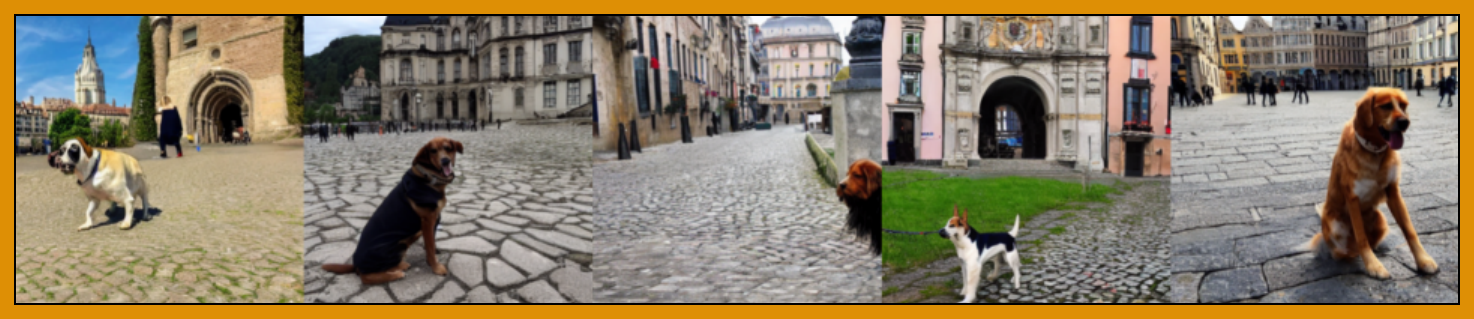}
         \includegraphics[width=0.99\textwidth]{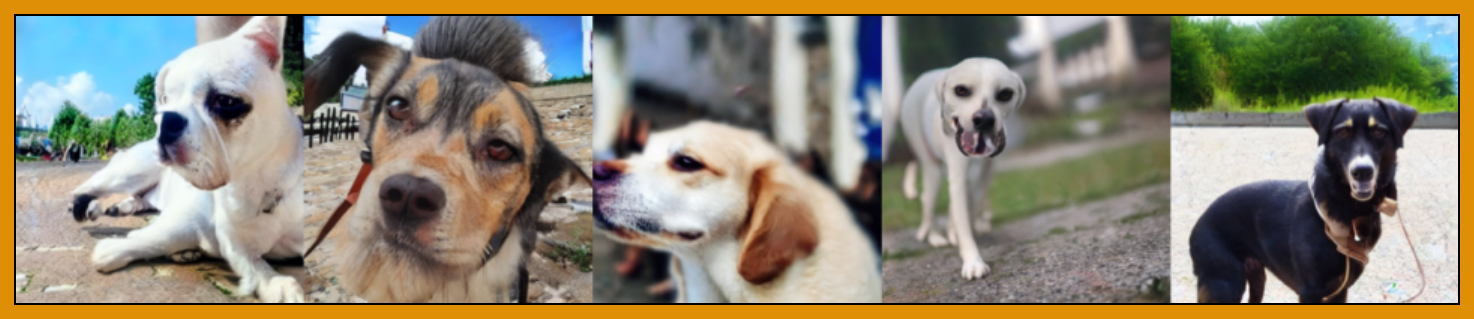}
         \caption{Europe}
    \end{subfigure}
    \caption{Examples of image seeds before \textbf{(TOP)} and after \textbf{(BOTTOM)} applying c-\methodnameshort to the LDM. 
    The prevalence of unrepresentative background info in a given image is reduced as the dogs become larger and more centrally focused.}
    \label{fig:consistency}
\end{figure}

In Section \ref{sec:results} we discuss how implementing c-\methodnameshort can reduce the prevalence of unrepresentative background features in some cases, as objects become larger and more central in the image. 
Figure \ref{fig:consistency} shows examples of this pattern for images of dogs, where many images without c-\methodnameshort have much smaller animals and more dominant backgrounds. 
These backgrounds are not representative of those in reference datasets like GeoDE (shown in Figure \ref{fig:diversity_issue_example}.G). 
For example, the real images for both Africa and Europe feature dogs on simple surfaces with little vegetation or architectural features, while the generated images with the original LDM have significant grass/dirt features for Africa and city features for Europe. 
While these background trends can remain even after c-\methodnameshort is applied, they are less prevalent in the images since the dogs are often larger and more centrally focused.

\begin{figure}
    \centering
    \begin{subfigure}[b]{0.75\textwidth}
        \includegraphics[width=0.99\textwidth]{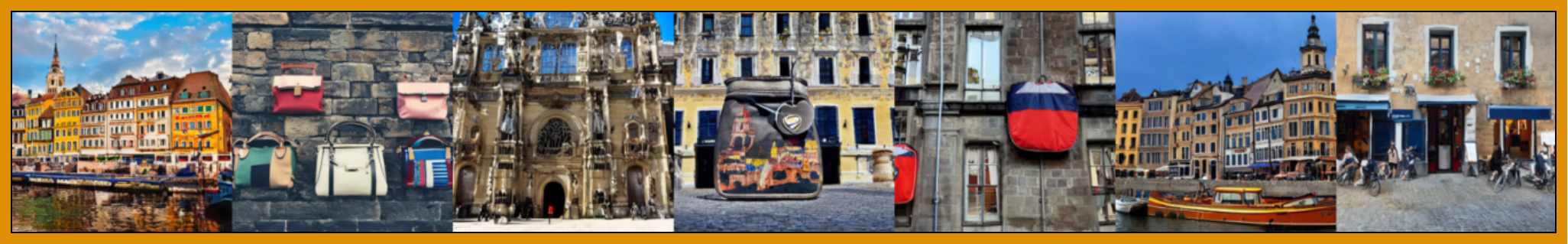}
         \caption{No \methodnameshort, \ie \objreg baseline}
    \end{subfigure}
    \begin{subfigure}[b]{0.75\textwidth}
        \includegraphics[width=0.99\textwidth]{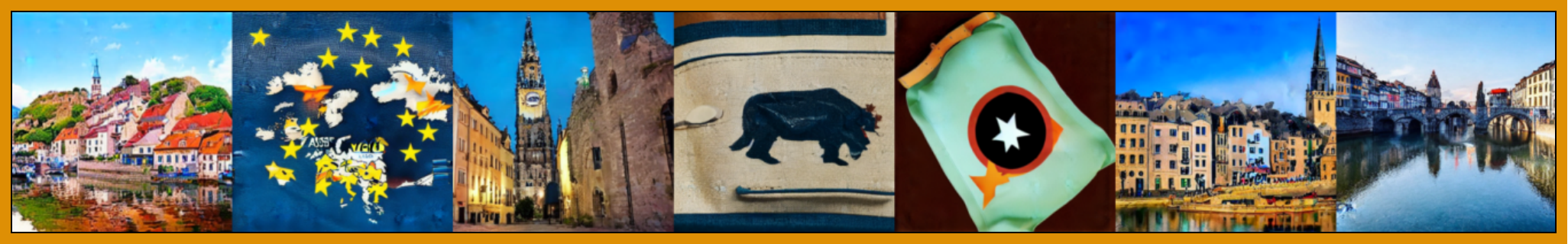}
        \caption{\methodname}
    \end{subfigure}
    \begin{subfigure}[b]{0.75\textwidth}
         \includegraphics[width=0.99\textwidth]{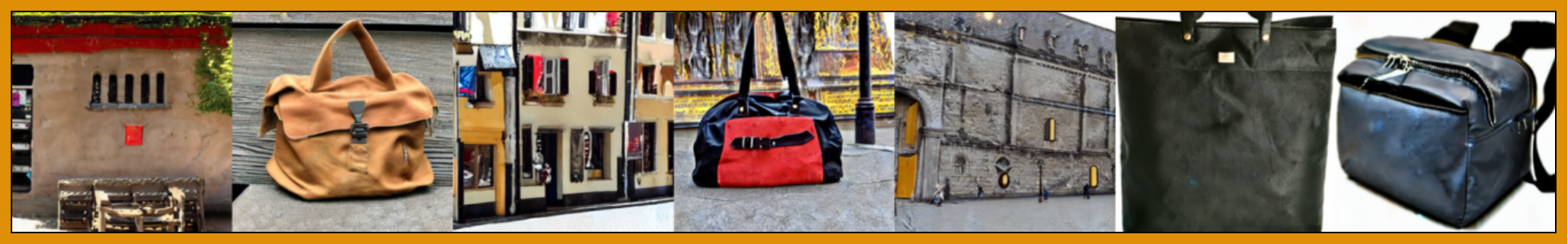}
         \caption{Contextualized \methodname}
    \end{subfigure}
    \caption{Examples of increased variation in background with c-\methodnameshort, even for low consistency images.}
    \label{fig:consistency2}
\end{figure}

In Section \ref{sec:results}, we also discuss how c-\methodnameshort can lead to less repetition of reductive backgrounds associated with regional tropes for images with poor consistency. 
In Figure \ref{fig:consistency2}, we show examples of random generations corresponding to the prompt ``\textit{bag in Europe}'' with different levels of guidance. 
We see that while there still remain some consistency issues, they are improved with contextualized-\methodnameshort. 
Furthermore, there is more variety in the background representation with c-\methodnameshort, including for images without bags: highly aesthetic European cityscapes (which are not present in the reference dataset) can be replaced with more neutral backgrounds. 

\subsubsection{Other forms of related diversity}
\begin{figure}[h]
    \centering
\includegraphics[width=0.50\textwidth]{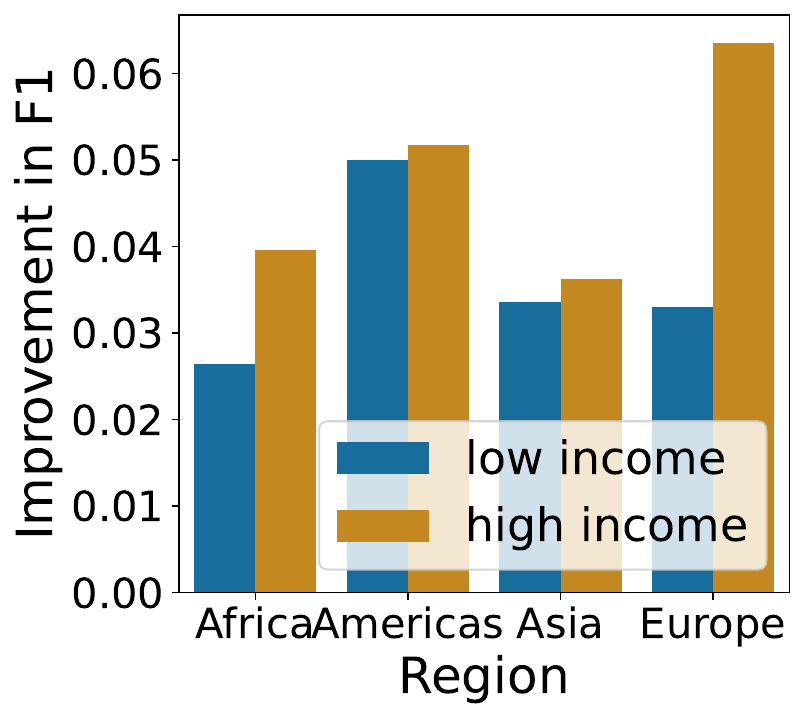} 
    \caption{C-VSG improves socioeconomic F1.}
            \label{fig:income}
\end{figure}

We perform preliminary studies into how socioeconomic and cultural contexts can also influence object representation by measuring the impact of c-VSG along \textit{income} buckets using DollarStreet.
We find that both higher and lower income buckets across all regions show improvement in F1 over the baseline LDM, with results depicted in Figure \ref{fig:income}.

\subsubsection{Ablations}
\label{app:ablations}

\begin{figure}[!t]
    \begin{subfigure}[b]{0.99\textwidth}
        \begin{subfigure}[b]{\textwidth}
            \centering
            \includegraphics[width=0.49\textwidth]{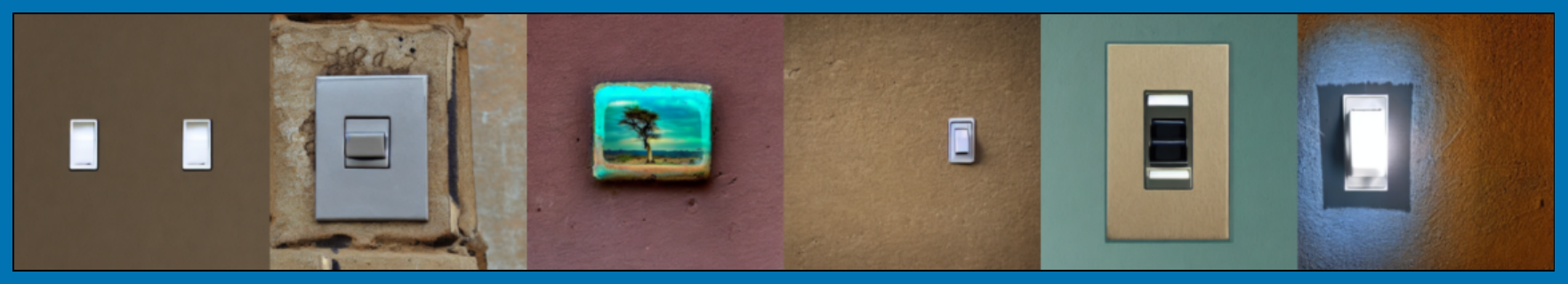}  
            \includegraphics[width=0.49\textwidth]{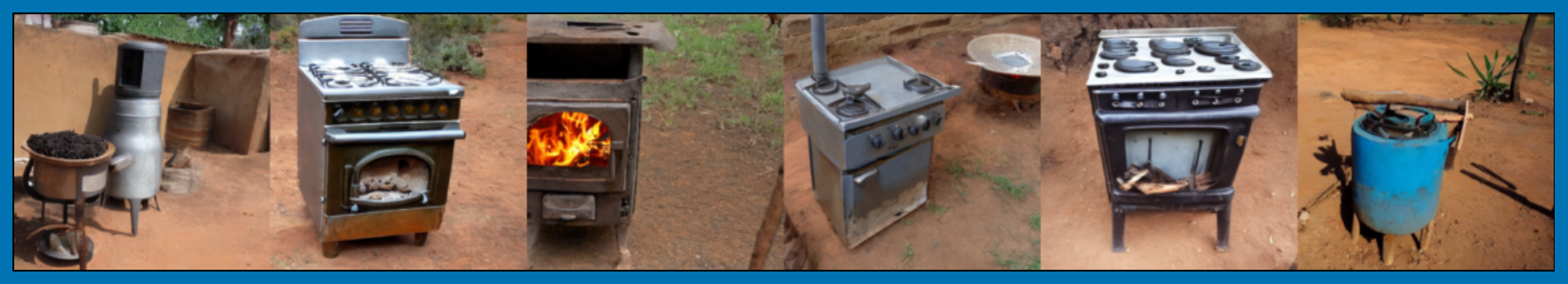}  \includegraphics[width=0.49\textwidth]{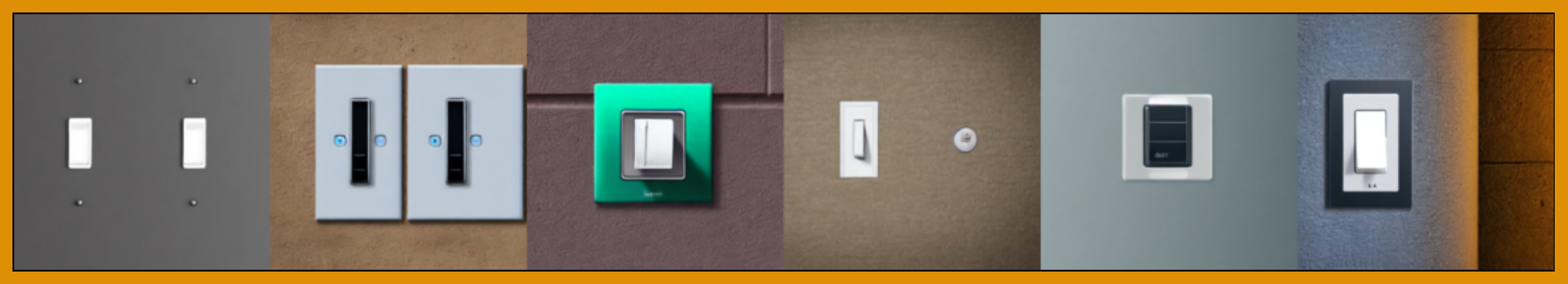}  
            \includegraphics[width=0.49\textwidth]{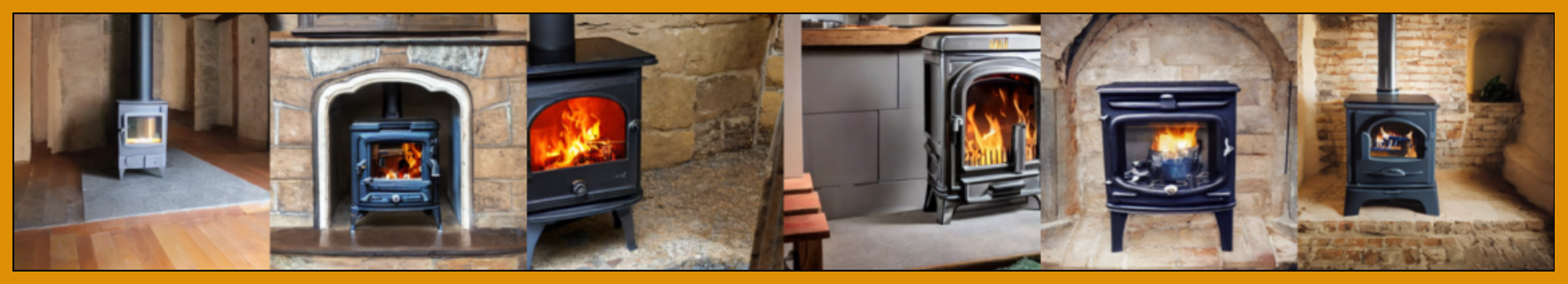}  \includegraphics[width=0.49\textwidth]{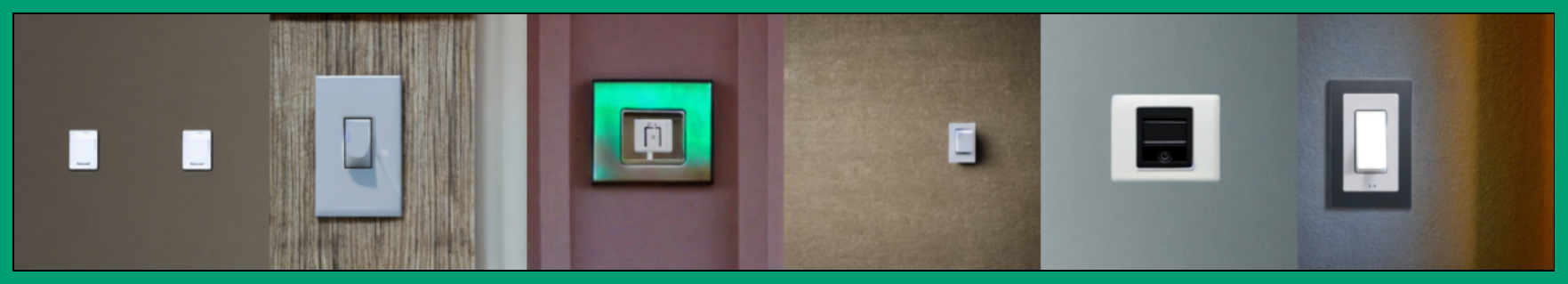}  \includegraphics[width=0.49\textwidth]{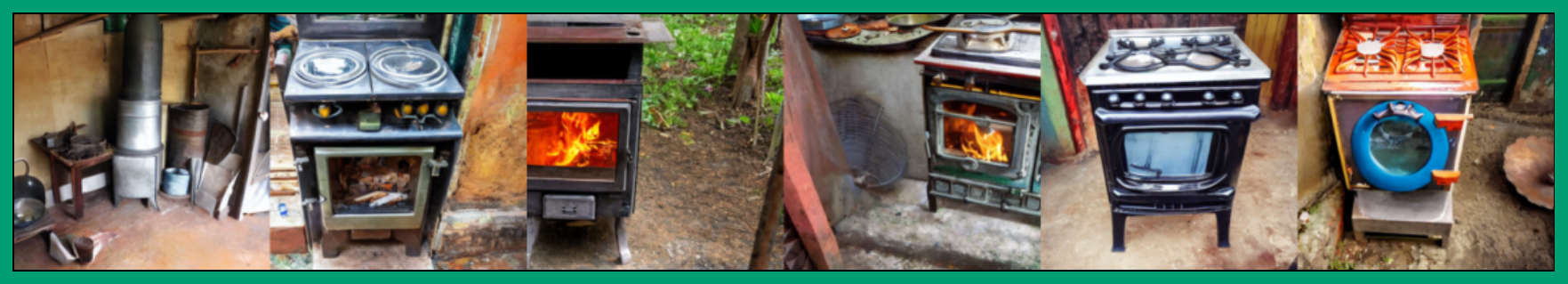}  

            \caption{LDM baseline, No \methodnameshort, \ie \objreg baseline}
        \end{subfigure} 
        \begin{subfigure}[b]{\textwidth}
            \centering
            \includegraphics[width=0.49\textwidth]{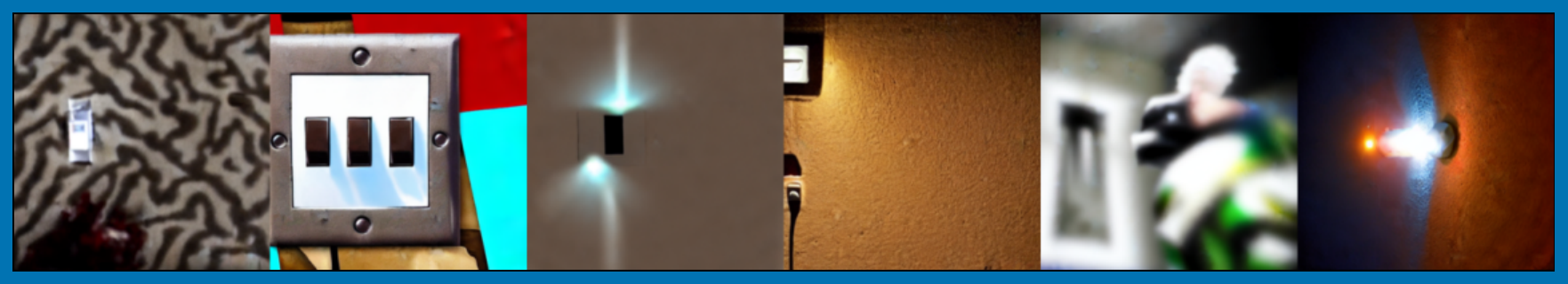}  
            \includegraphics[width=0.49\textwidth]{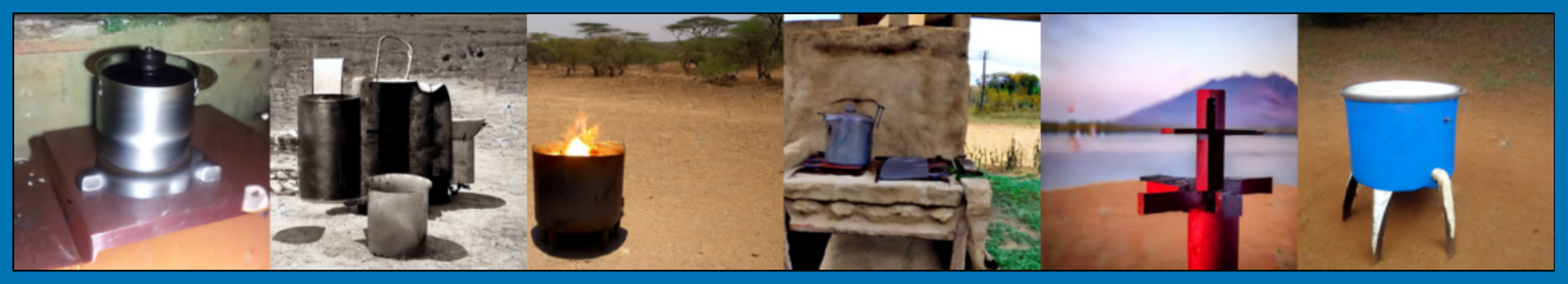}  
            \includegraphics[width=0.49\textwidth]{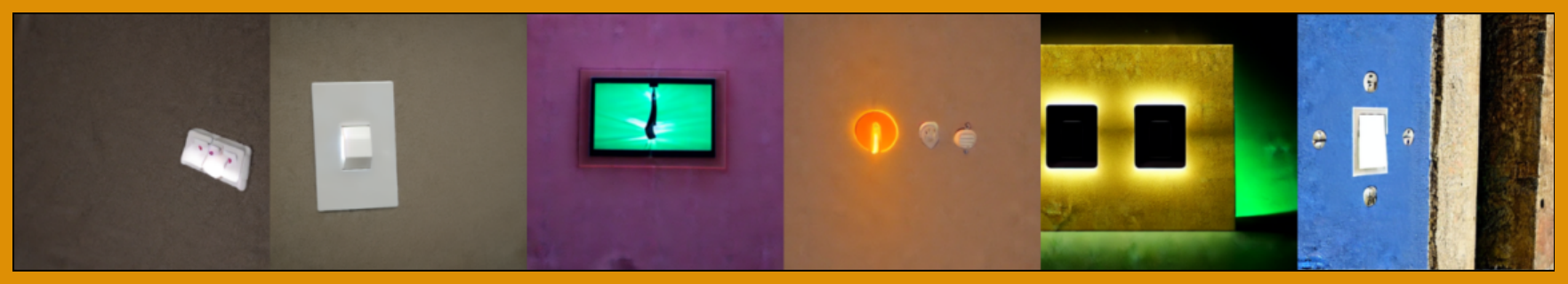}  
            \includegraphics[width=0.49\textwidth]{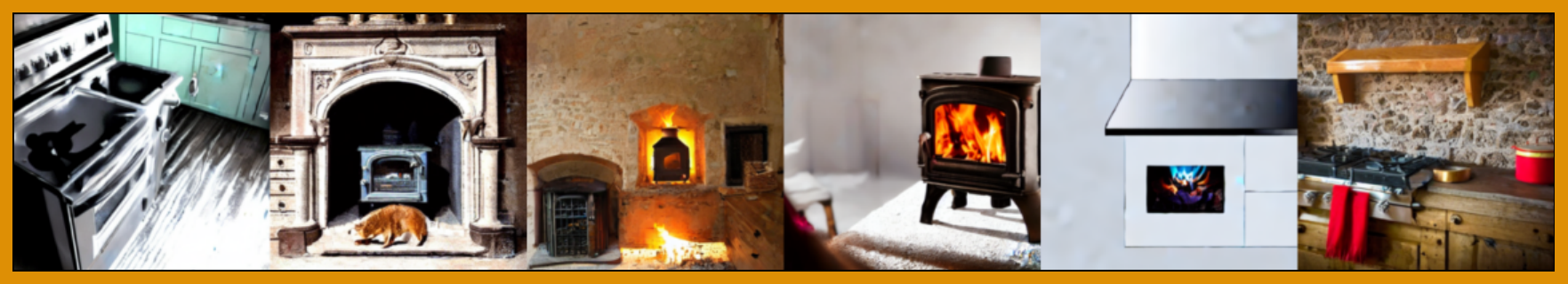}  
            \includegraphics[width=0.49\textwidth]{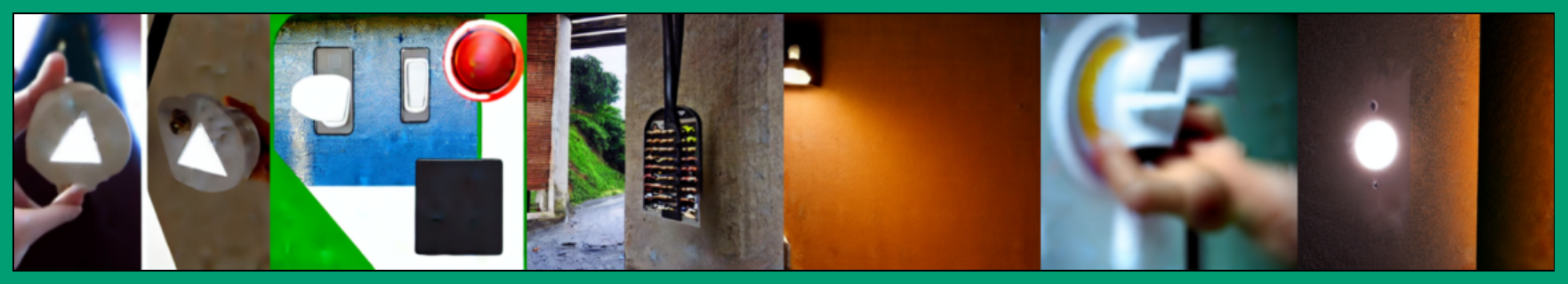}  
            \includegraphics[width=0.49\textwidth]{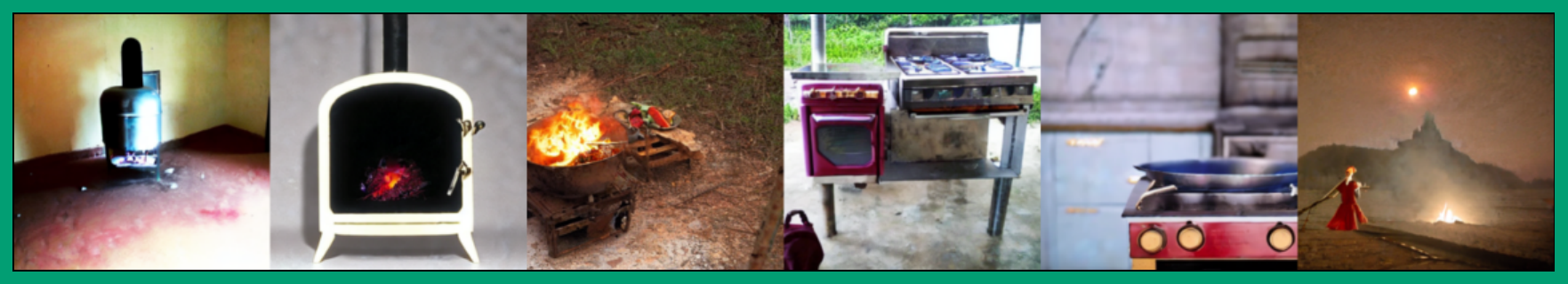}  
            \caption{ \methodname}
        \end{subfigure} 
        \begin{subfigure}[b]{\textwidth}
            \centering
            \includegraphics[width=0.49\textwidth]{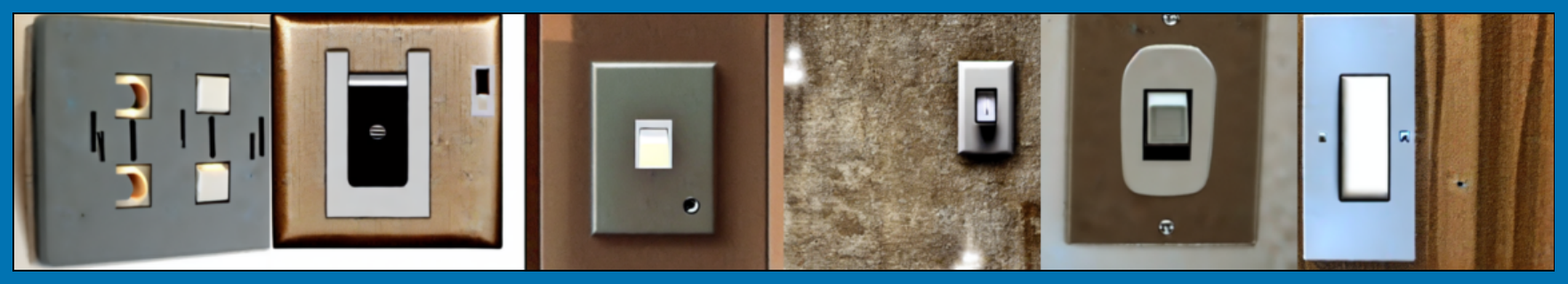}  
            \includegraphics[width=0.49\textwidth]{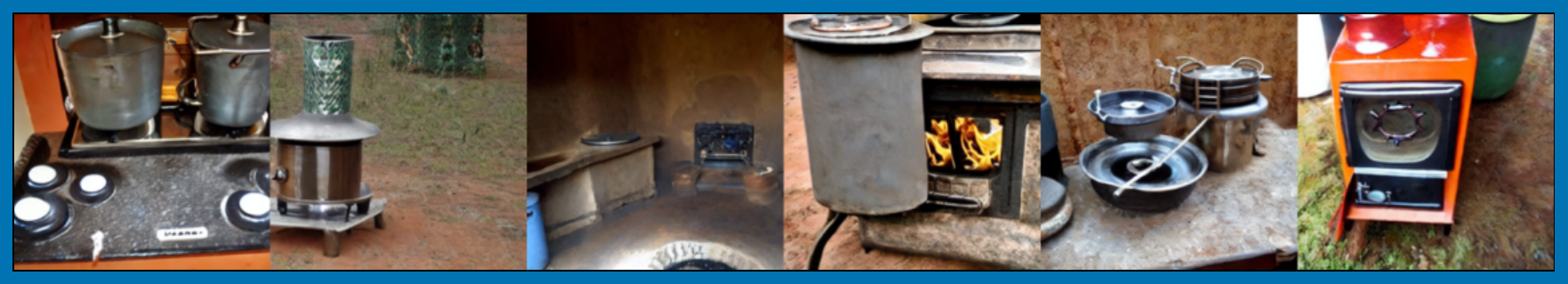}  
            \includegraphics[width=0.49\textwidth]{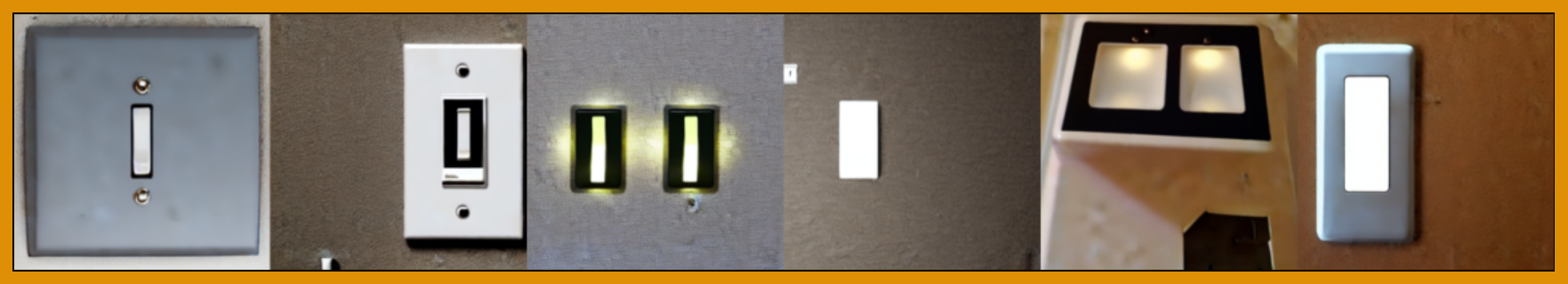}  
            \includegraphics[width=0.49\textwidth]{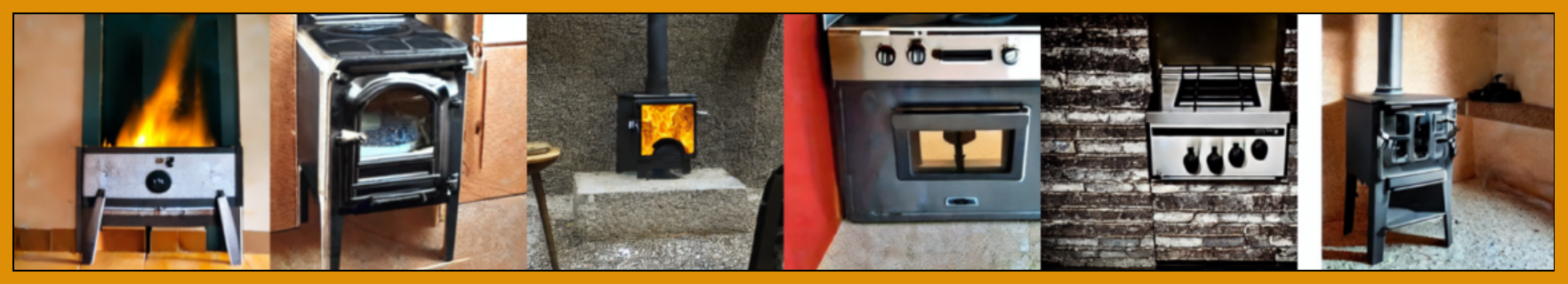}  
            \includegraphics[width=0.49\textwidth]{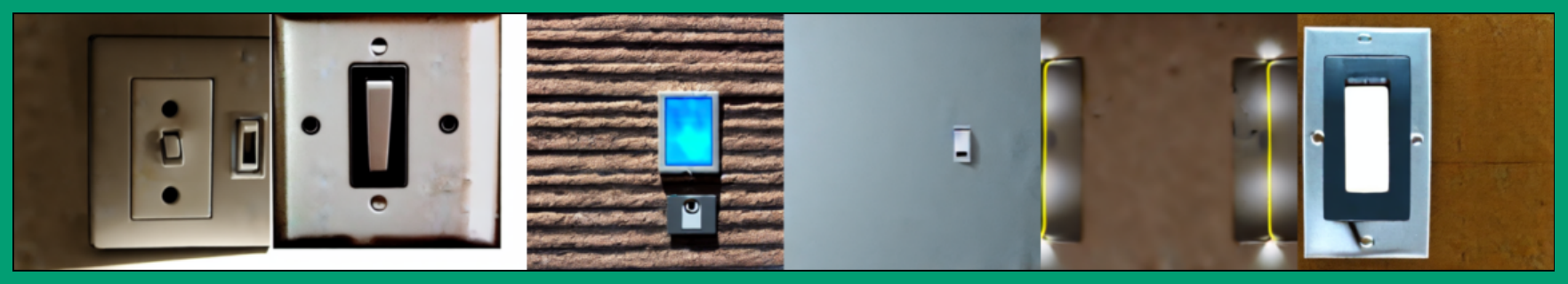}  
            \includegraphics[width=0.49\textwidth]{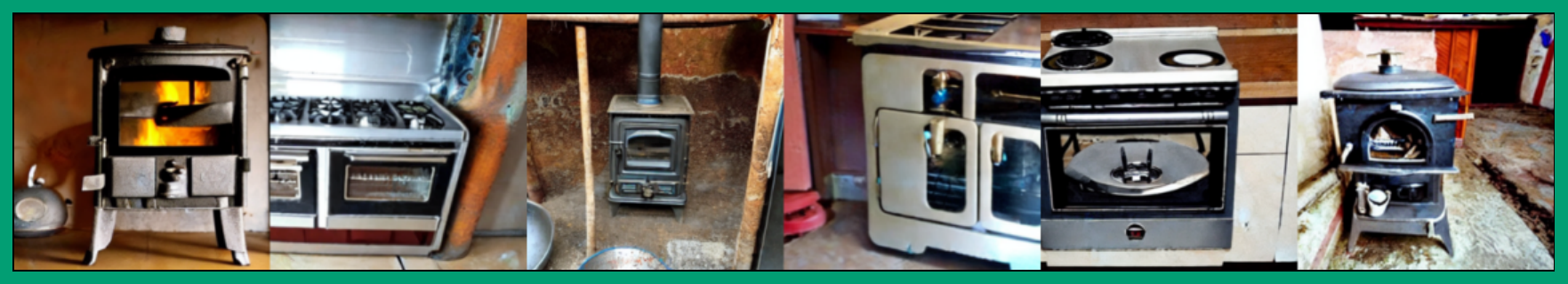}  
            \caption{ Only contextualizing images}
        \end{subfigure} 
        \begin{subfigure}[b]{\textwidth}
            \centering\includegraphics[width=0.49\textwidth]{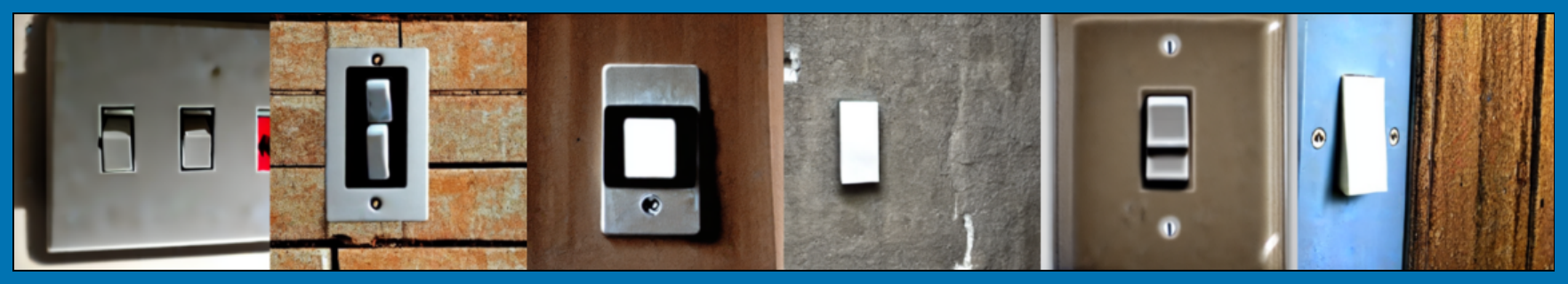}
            \includegraphics[width=0.49\textwidth]{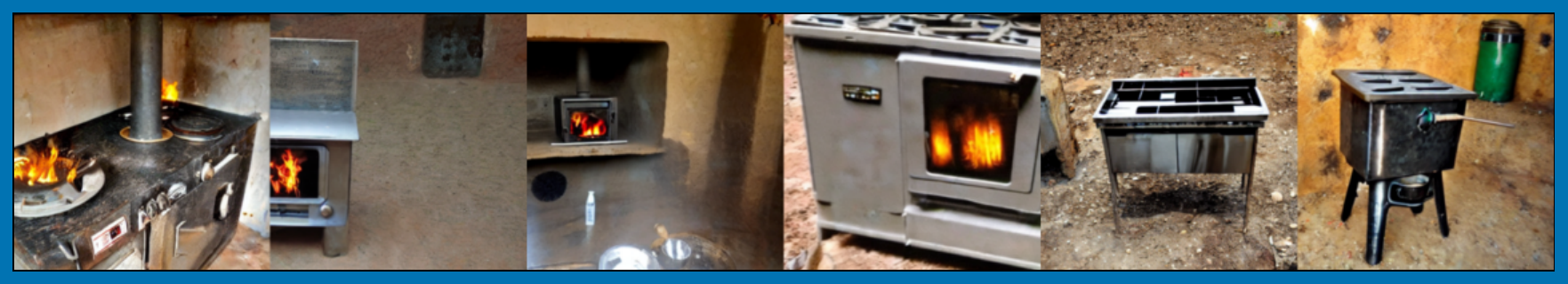}
            \includegraphics[width=0.49\textwidth]{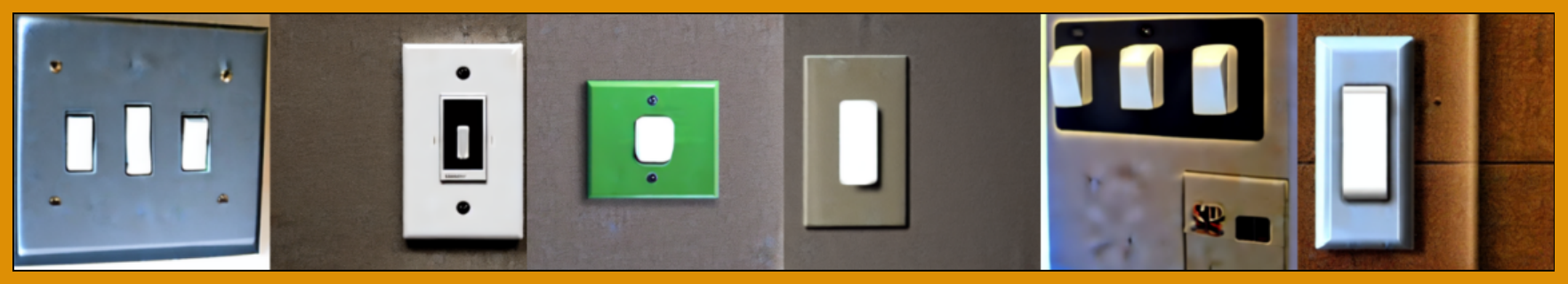}
            \includegraphics[width=0.49\textwidth]{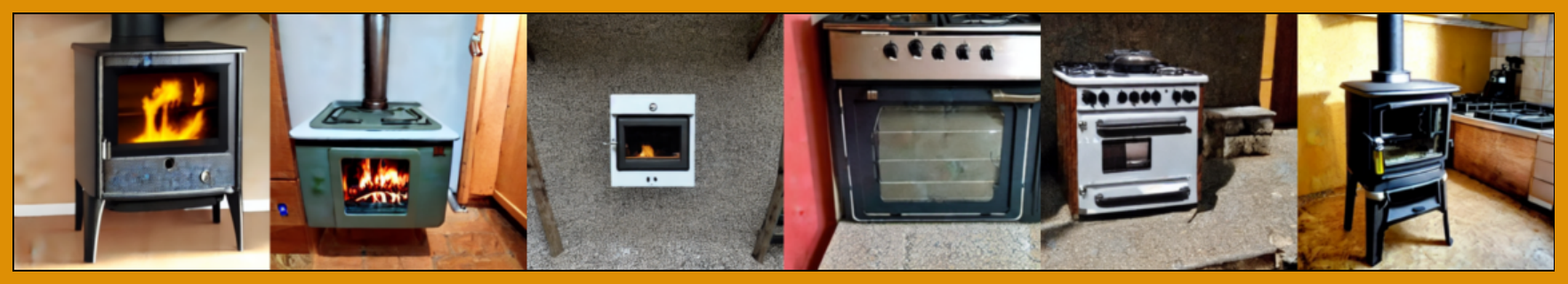}
            \includegraphics[width=0.49\textwidth]{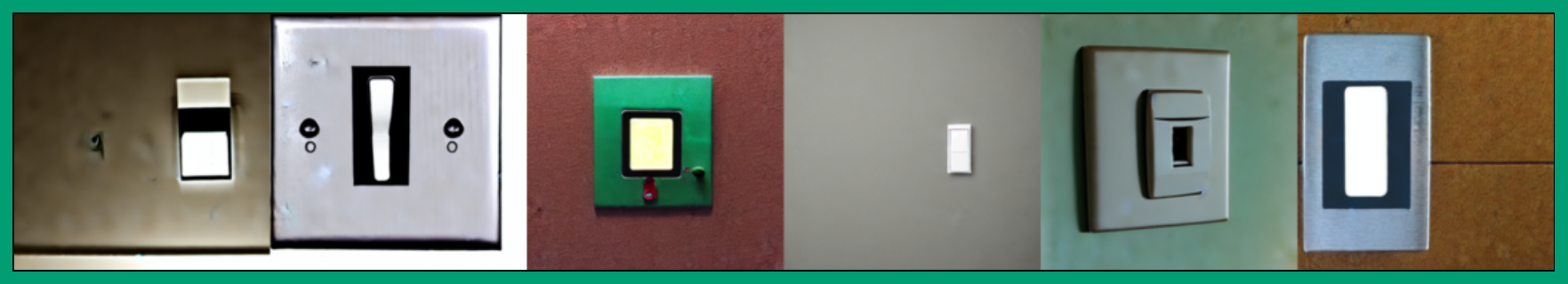}
            \includegraphics[width=0.49\textwidth]{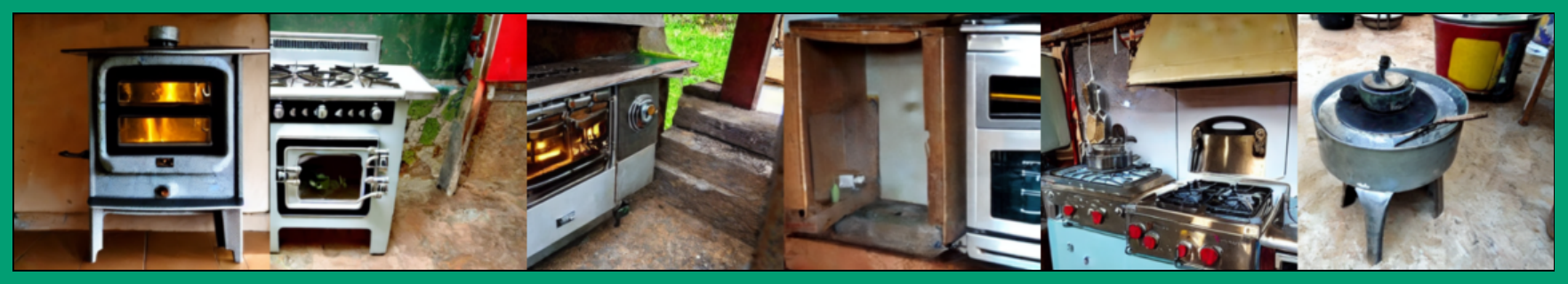}
            \caption{ Contextualized \methodname}
        \end{subfigure} 
    \end{subfigure}
    \caption{
    Generated images of light switches (\textbf{Left}) and stoves (\textbf{Right}).
    The same seed is used for a given column, and the box colors indicate images pertaining to \textcolor{blue}{Africa}, \textcolor{orange}{Europe}, and \textcolor{darkgreen}{Southeast Asia}.
    }
    \label{fig:res_visuals_abalations}
\end{figure}

In Section \ref{sec:ablations}, we study variations of the c-\methodnameshort criteria. 
Figure \ref{fig:res_visuals_abalations}.B shows examples of how Vendi Score Guidance without
contextualization can lead to images with greater diversity but unrealistic object shapes and colors.
This includes some extreme variation in styles, such as black-and-white or film-style photos. 
Figure \ref{fig:res_visuals_abalations}.C shows that exclusively contextualizing images helps
in almost all aspects as opposed to only the Vendi Score guidance on the bank of
generated images.
However, the highest F1 measures occur for contextualized \methodname (shown in Figure \ref{fig:res_visuals_abalations}.D).

\begin{figure}[!t]
    \begin{subfigure}[b]{0.99\textwidth}
        \begin{subfigure}[b]{\textwidth}
            \centering
            \includegraphics[width=0.49\textwidth]{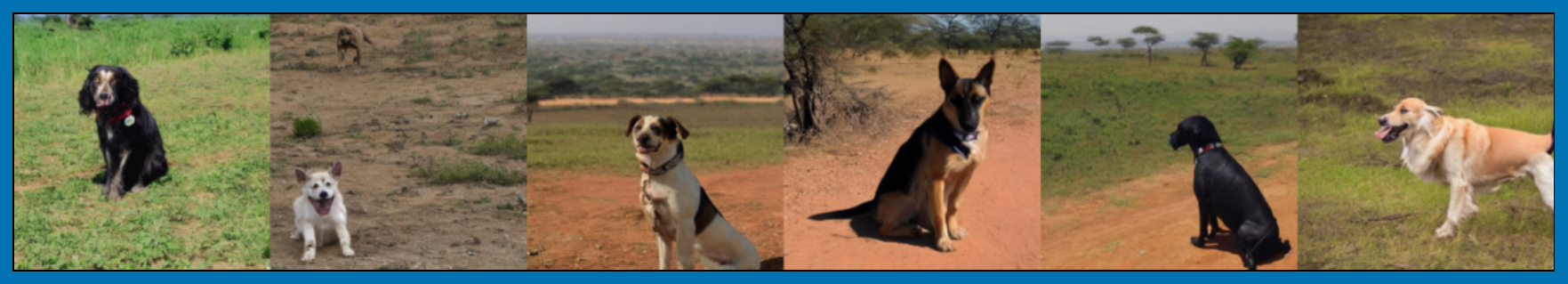}  
            \includegraphics[width=0.49\textwidth]{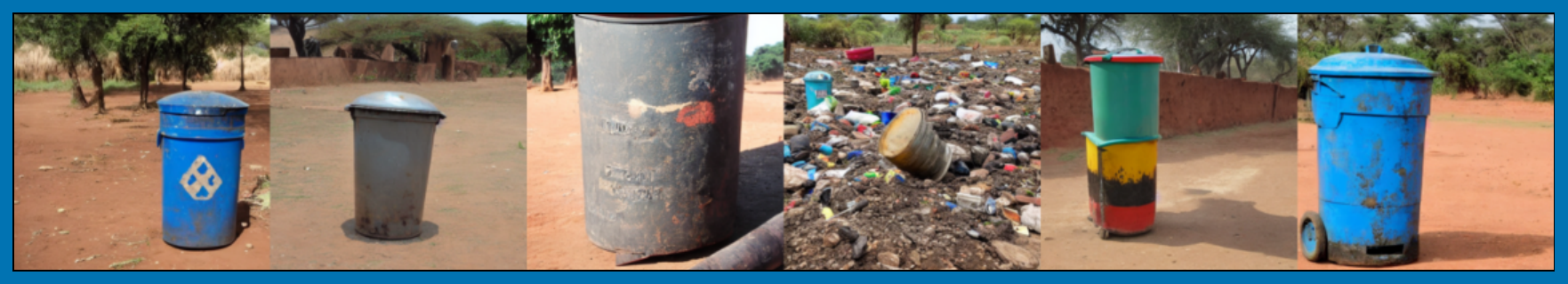}  \includegraphics[width=0.49\textwidth]{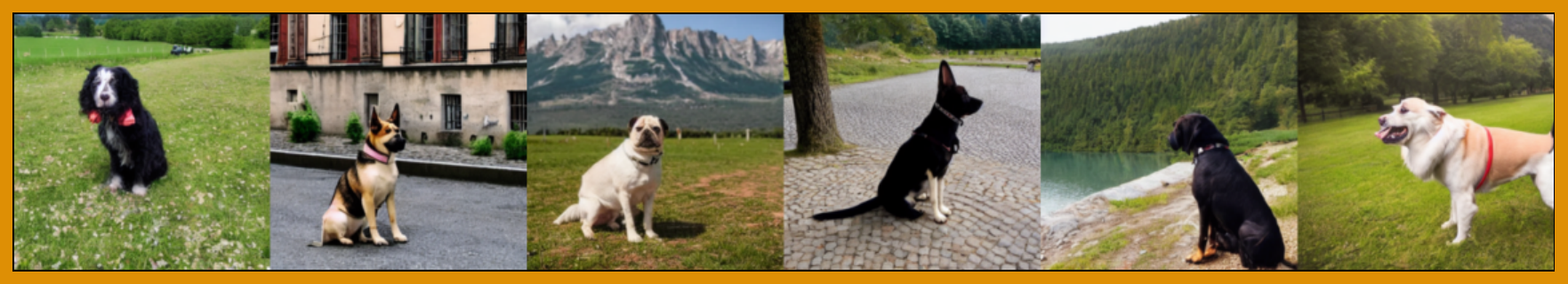}  
            \includegraphics[width=0.49\textwidth]{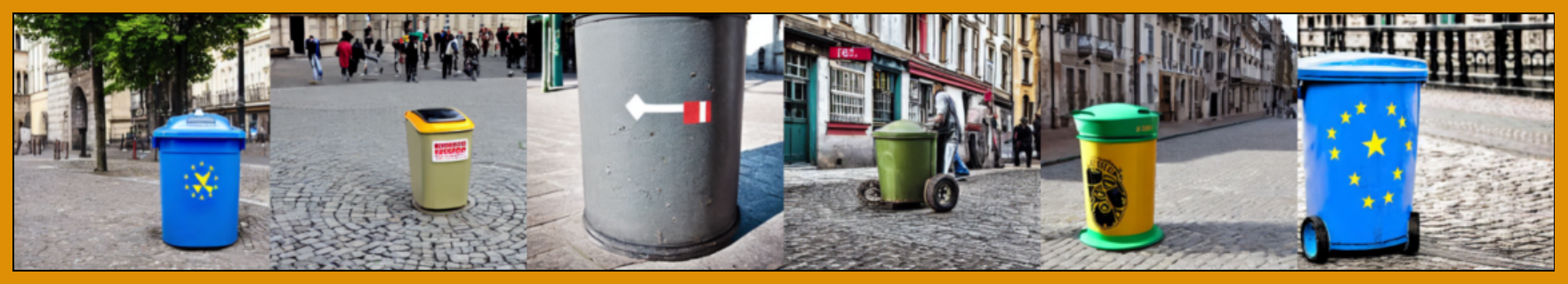}  \includegraphics[width=0.49\textwidth]{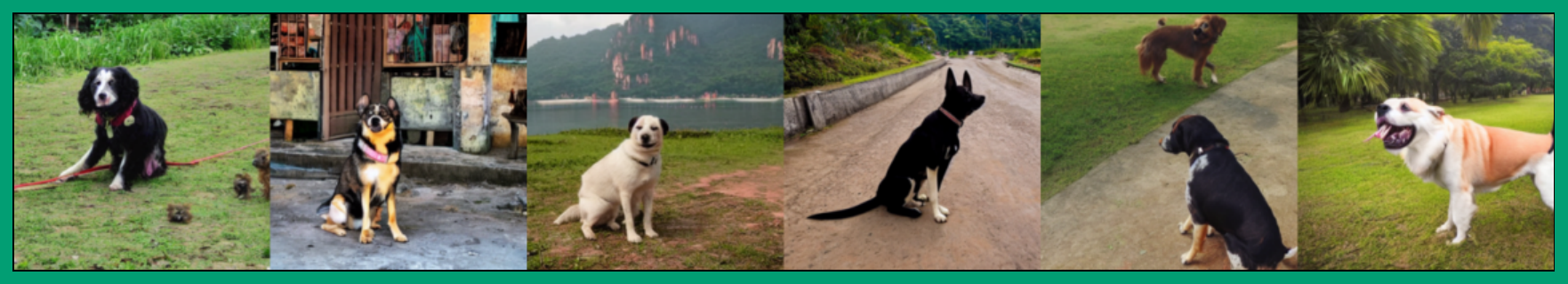}  \includegraphics[width=0.49\textwidth]{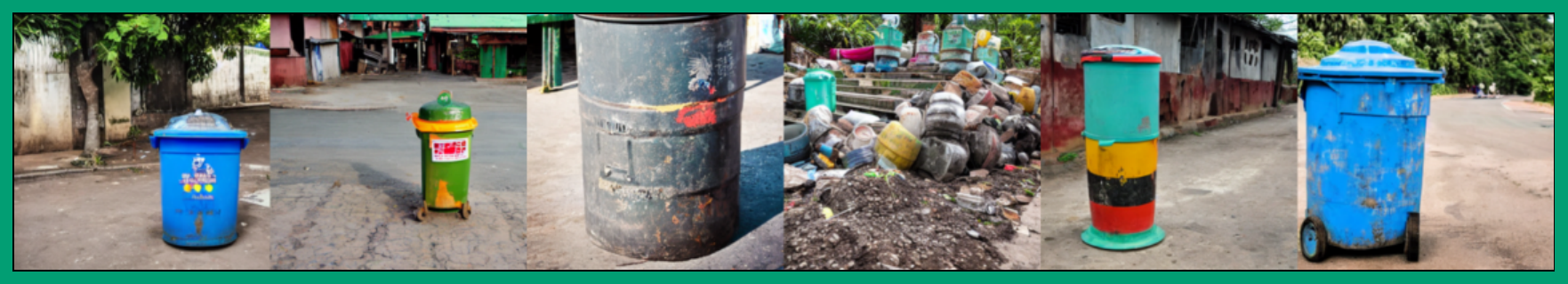}  

            \caption{LDM baseline, No \methodnameshort, \ie \objreg baseline}
        \end{subfigure} 
        \begin{subfigure}[b]{\textwidth}
            \centering
            \includegraphics[width=0.49\textwidth]{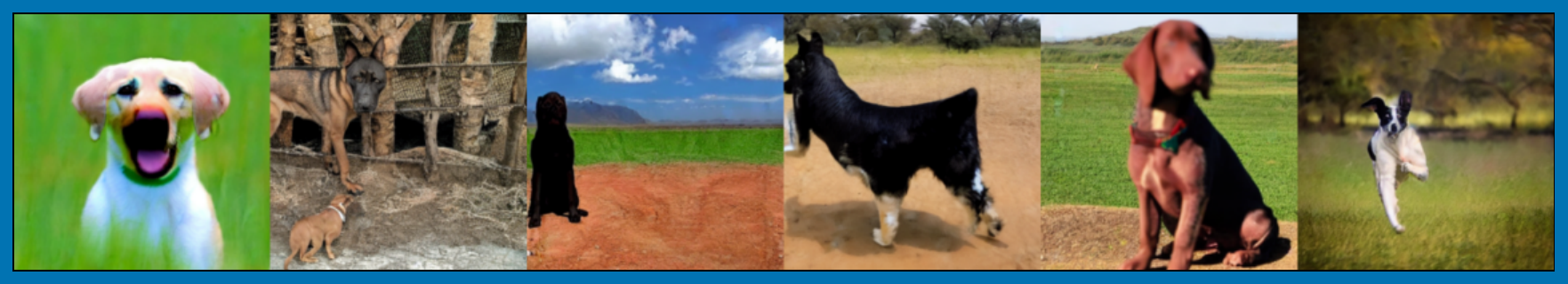}  
            \includegraphics[width=0.49\textwidth]{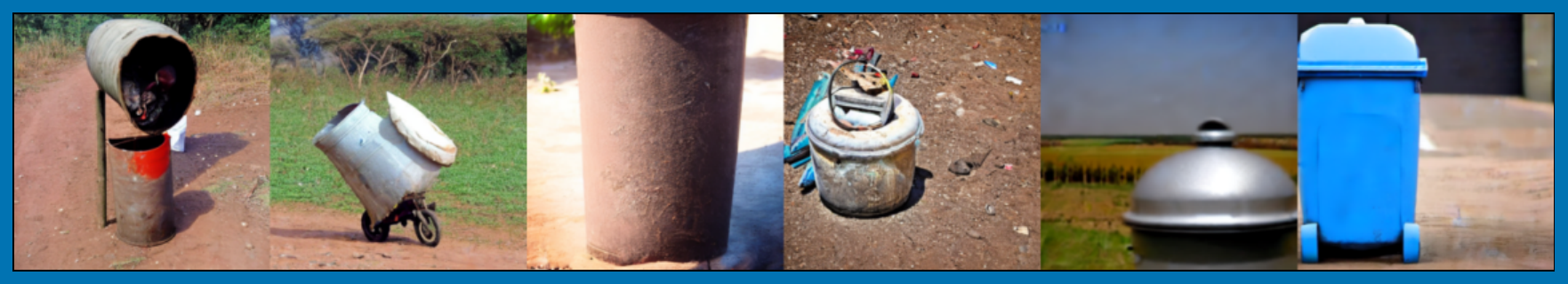}  
            \includegraphics[width=0.49\textwidth]{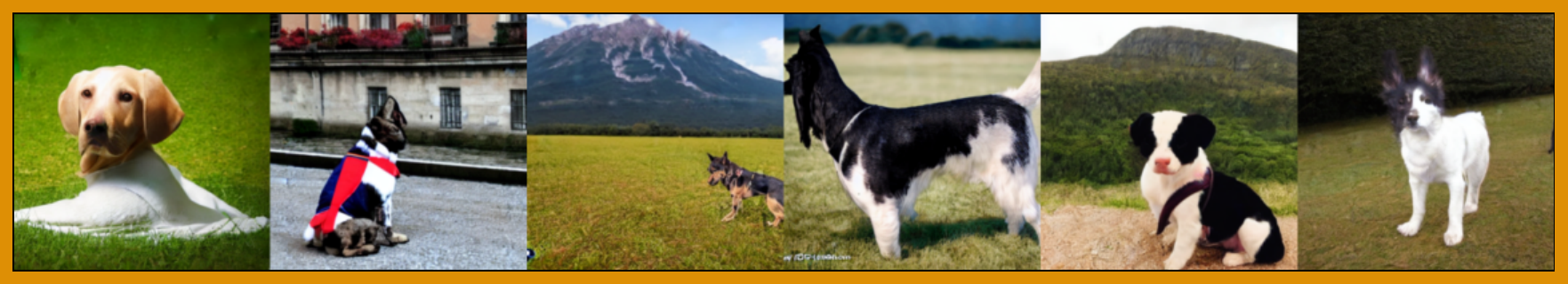}  
            \includegraphics[width=0.49\textwidth]{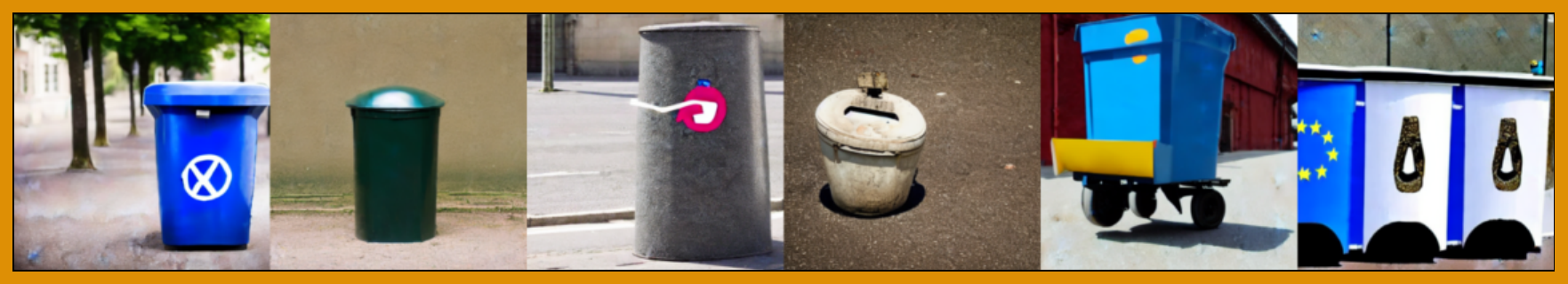}  
            \includegraphics[width=0.49\textwidth]{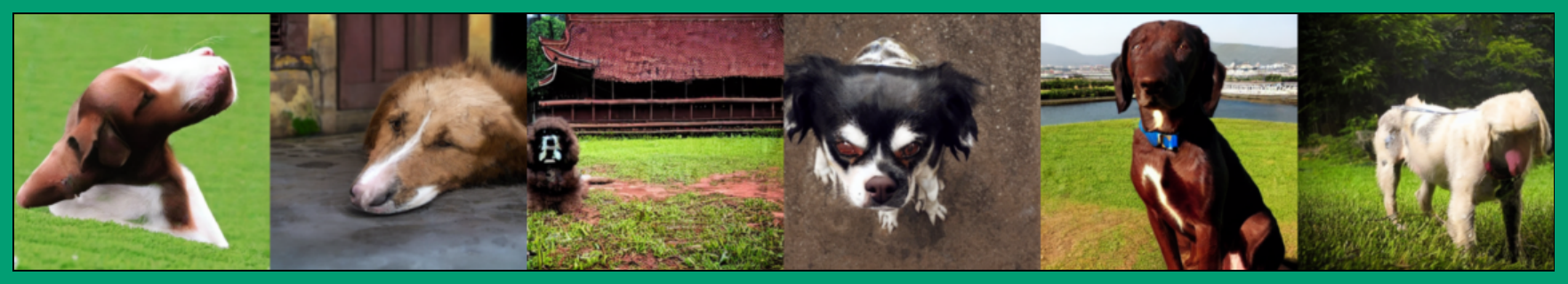}  
            \includegraphics[width=0.49\textwidth]{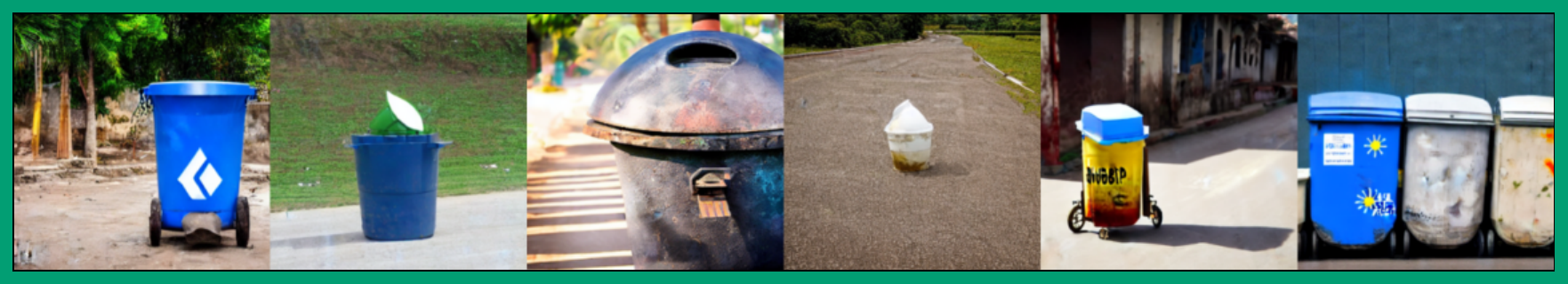}  
            \caption{ \methodname}
        \end{subfigure} 
        \begin{subfigure}[b]{\textwidth}
            \centering
            \includegraphics[width=0.49\textwidth]{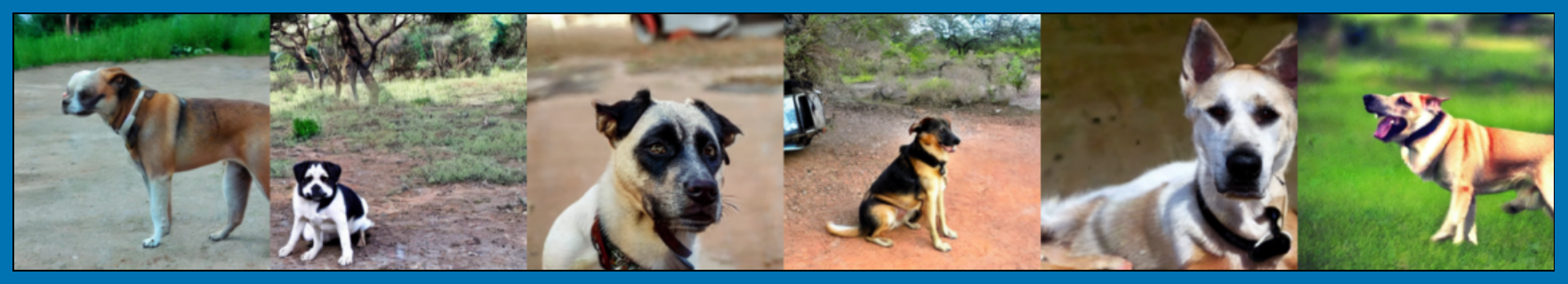}  
            \includegraphics[width=0.49\textwidth]{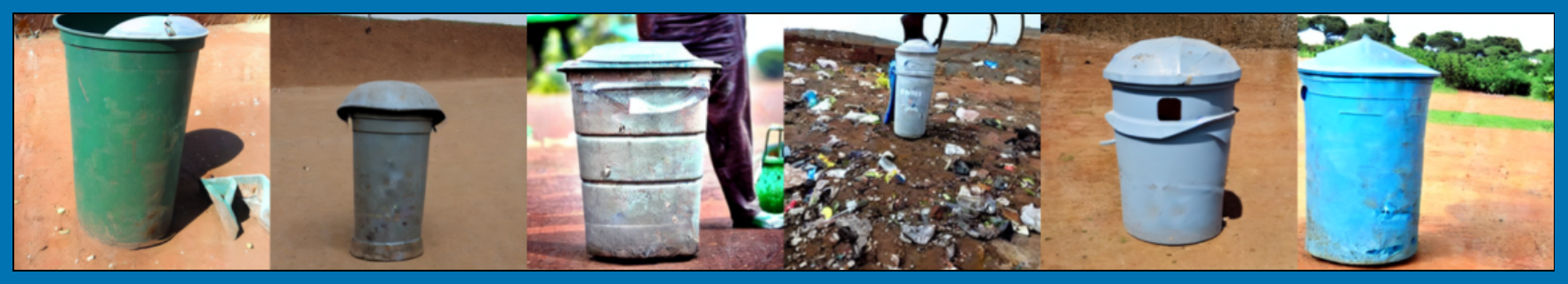}  
            \includegraphics[width=0.49\textwidth]{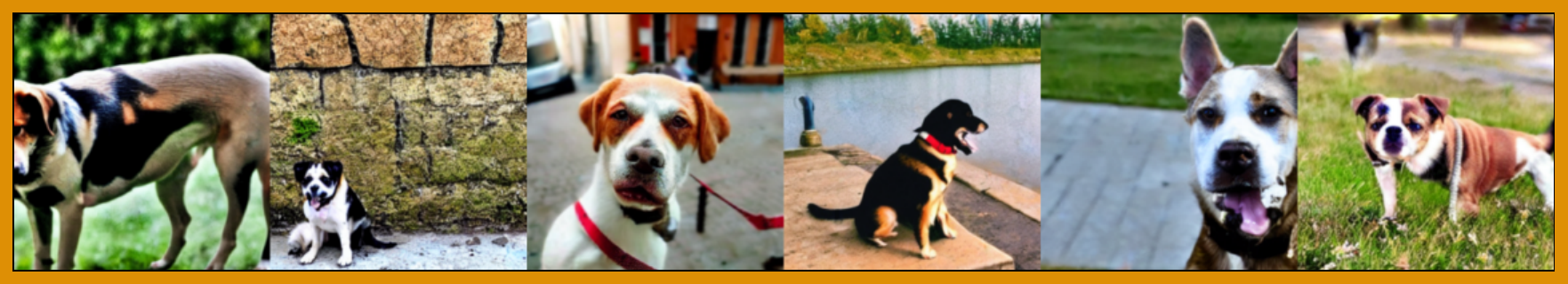}  
            \includegraphics[width=0.49\textwidth]{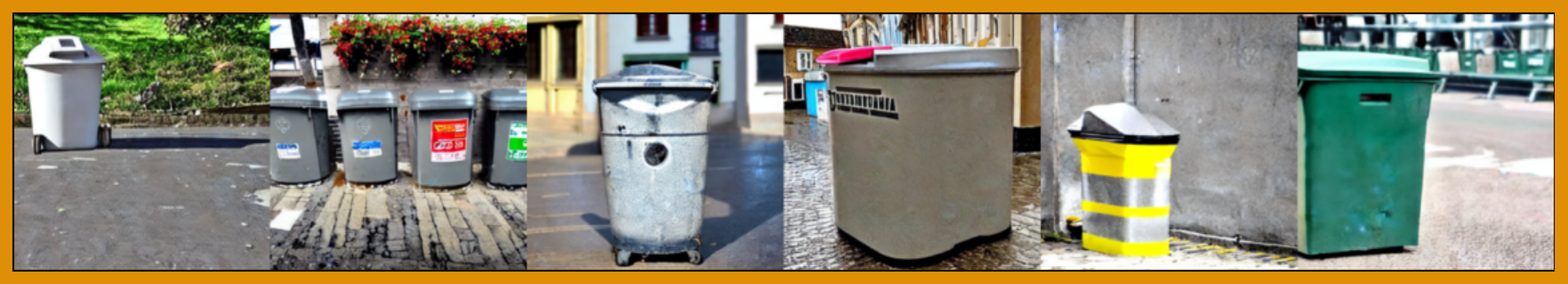}  
            \includegraphics[width=0.49\textwidth]{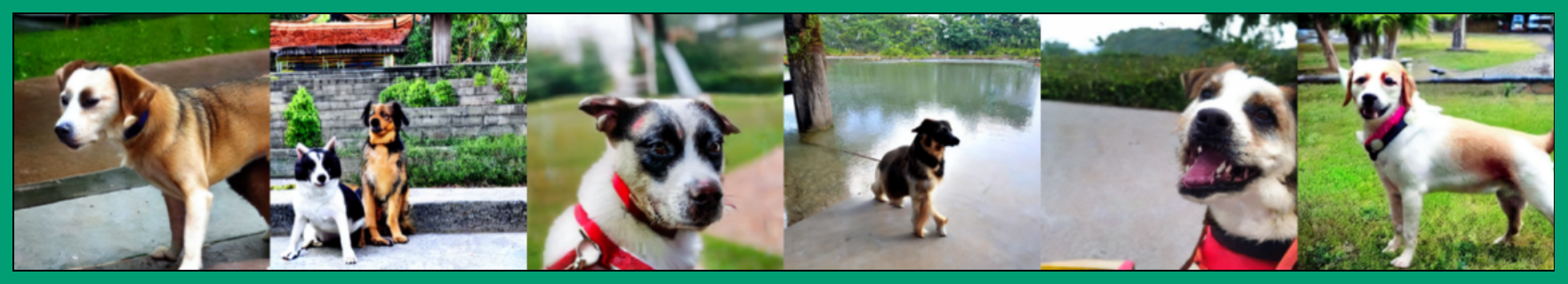}  
            \includegraphics[width=0.49\textwidth]{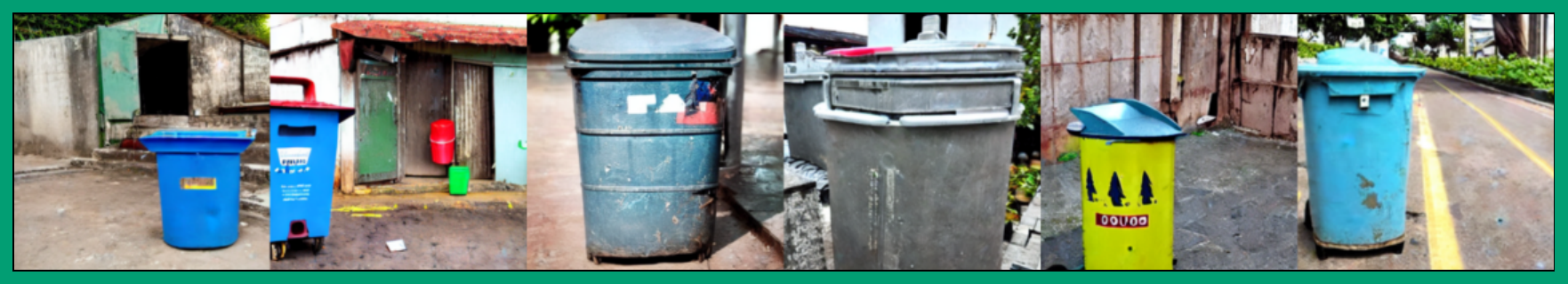}  
            \caption{ Only contextualizing images}
        \end{subfigure} 
        \begin{subfigure}[b]{\textwidth}
            \centering\includegraphics[width=0.49\textwidth]{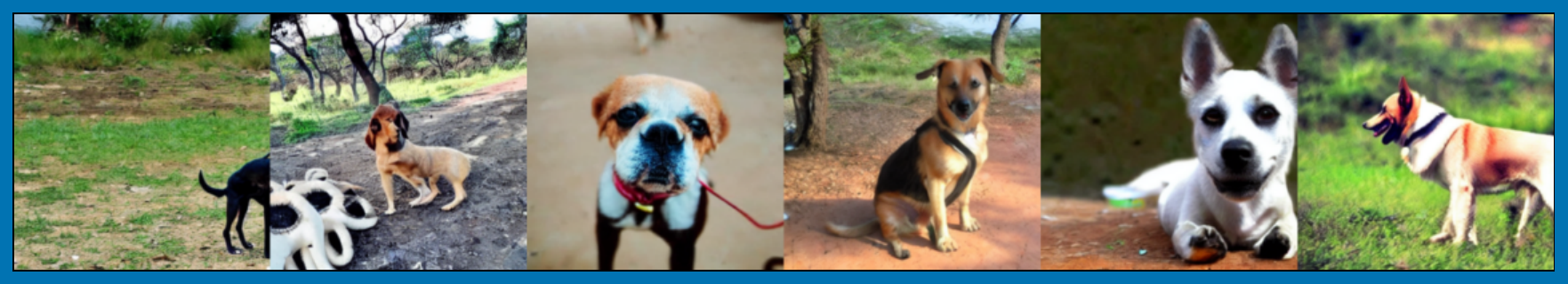}
            \includegraphics[width=0.49\textwidth]{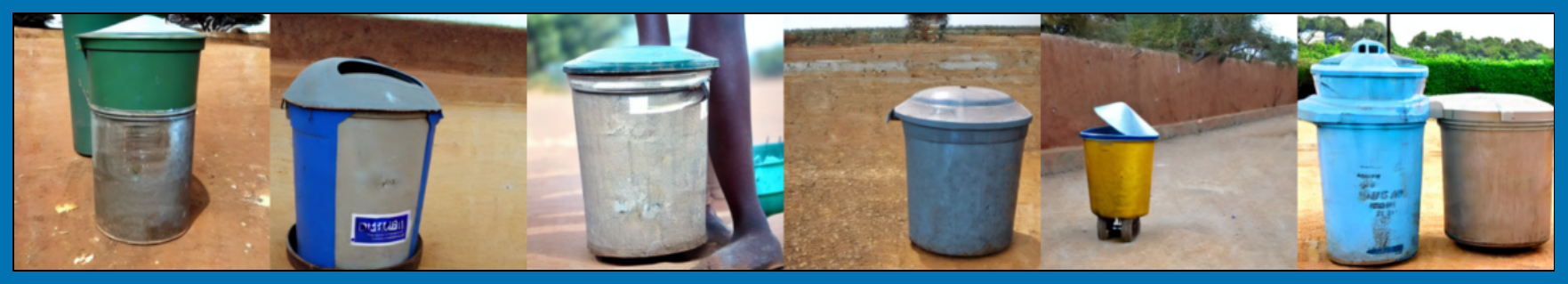}
            \includegraphics[width=0.49\textwidth]{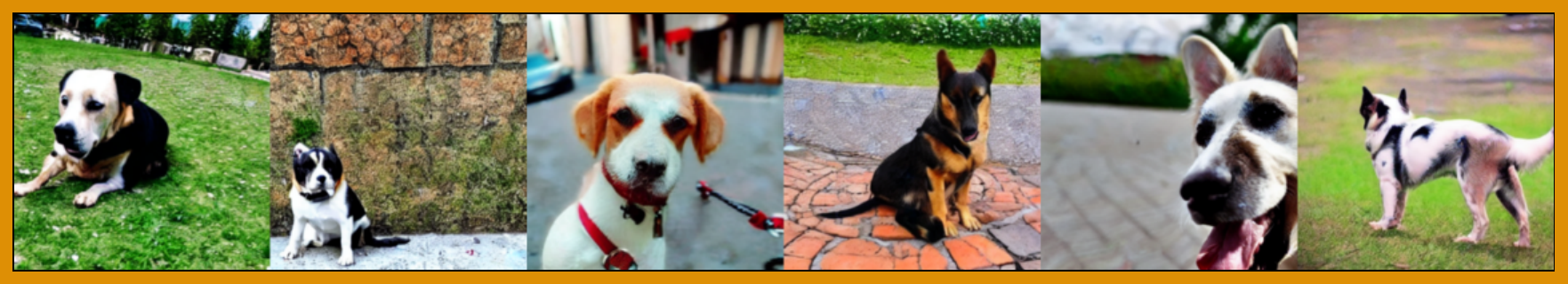}
            \includegraphics[width=0.49\textwidth]{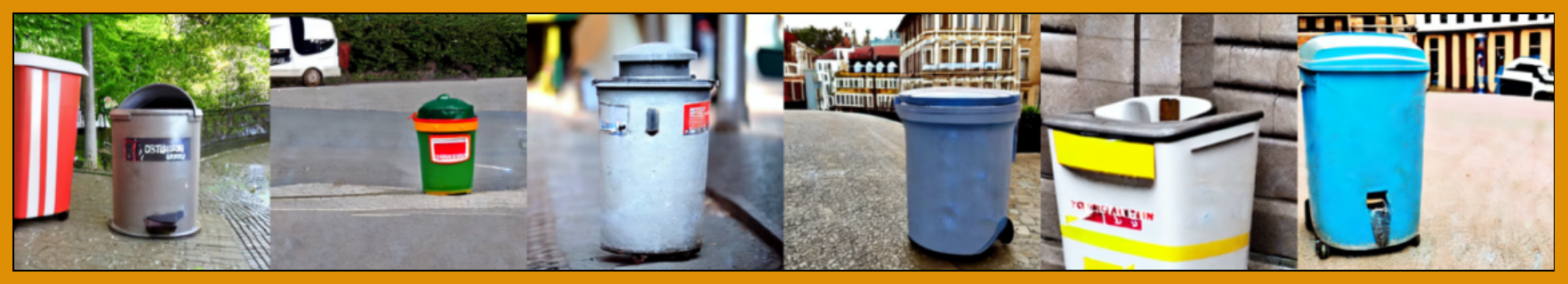}
            \includegraphics[width=0.49\textwidth]{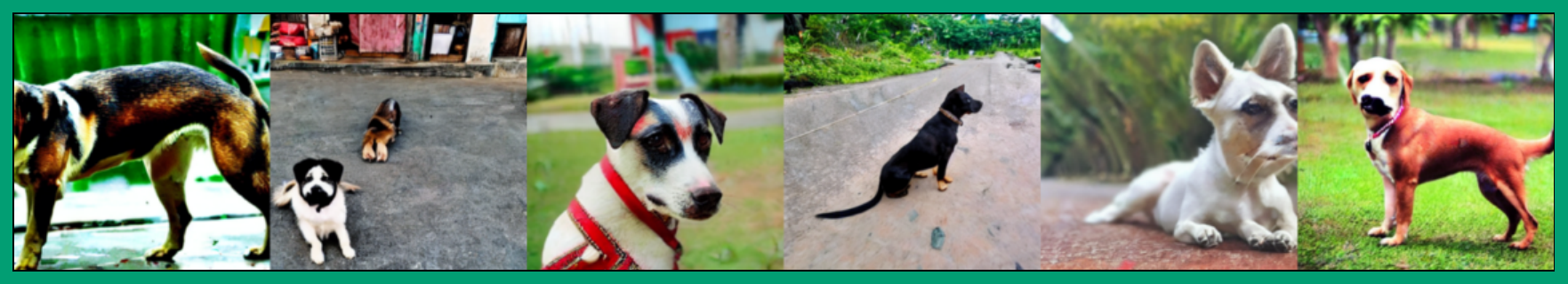}
            \includegraphics[width=0.49\textwidth]{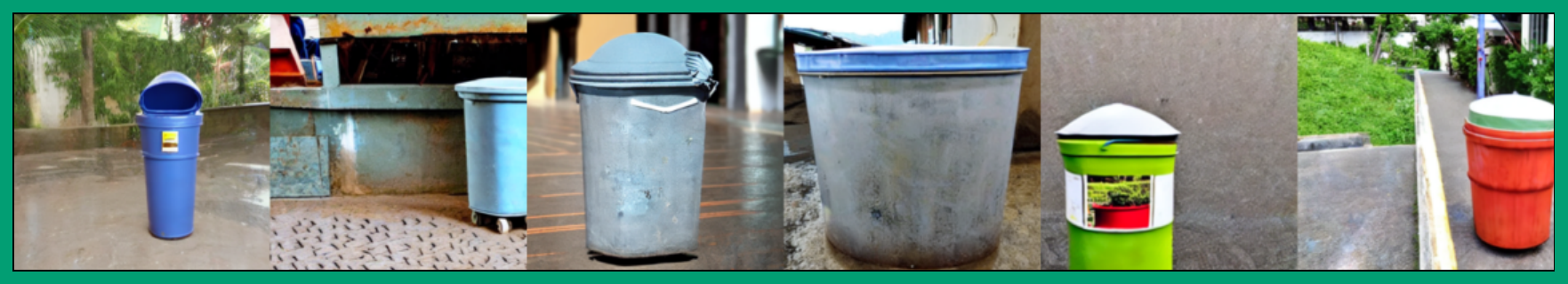}
            \caption{ Contextualized \methodname}
        \end{subfigure} 
    \end{subfigure}
    \caption{
    Generated images of dogs (\textbf{Left}) and dustbins (\textbf{Right}).
    The same seed is used for a given column, and the box colors indicate images pertaining to \textcolor{blue}{Africa}, \textcolor{orange}{Europe}, and \textcolor{darkgreen}{Southeast Asia}.
    }
    \label{fig:res_visuals_abalations_2}
\end{figure}

\begin{figure}[!ht]
    \centering
    \begin{subfigure}[b]{0.99\textwidth}
        \centering  
        \includegraphics[width=0.60\textwidth]{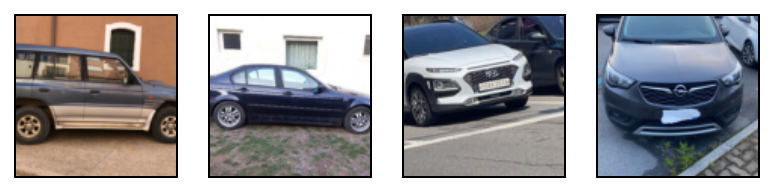}
        \caption{Reference images used for c-\methodnameshort}
    \end{subfigure}
        \begin{subfigure}[b]{0.32\textwidth}
             \includegraphics[width=0.99\textwidth]{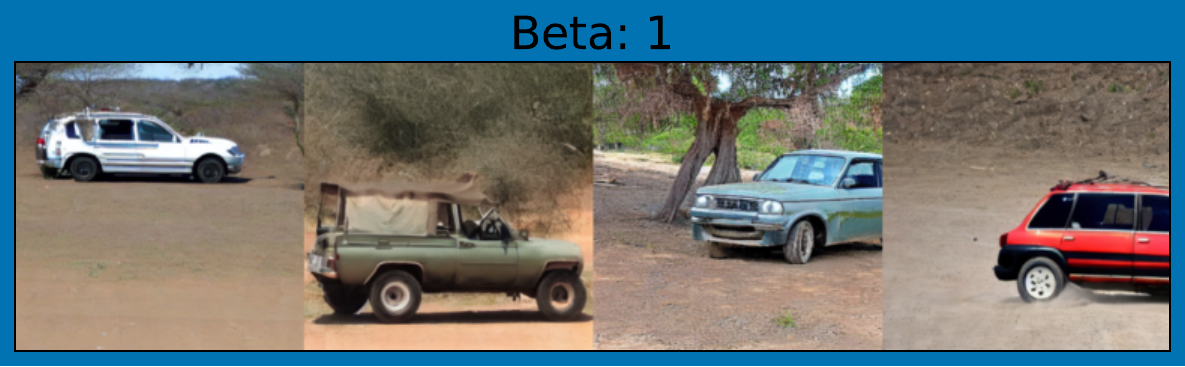}
             \includegraphics[width=0.99\textwidth]{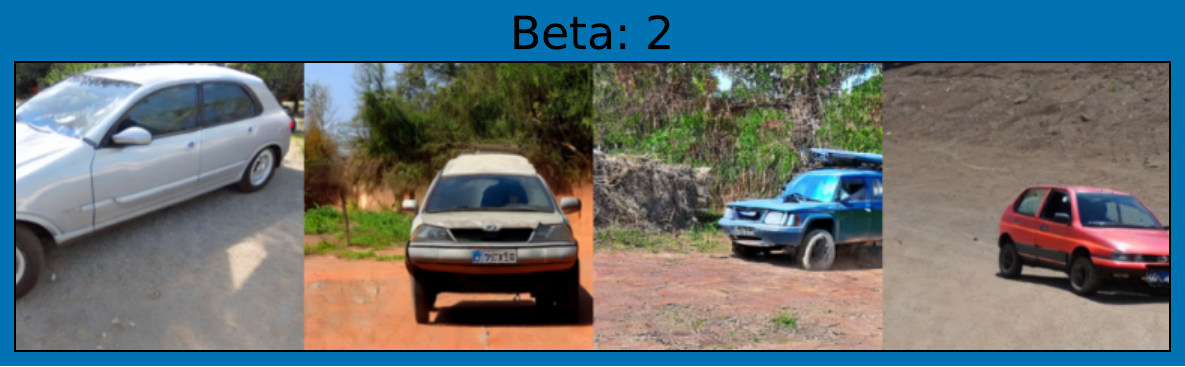}
            \includegraphics[width=0.99\textwidth]{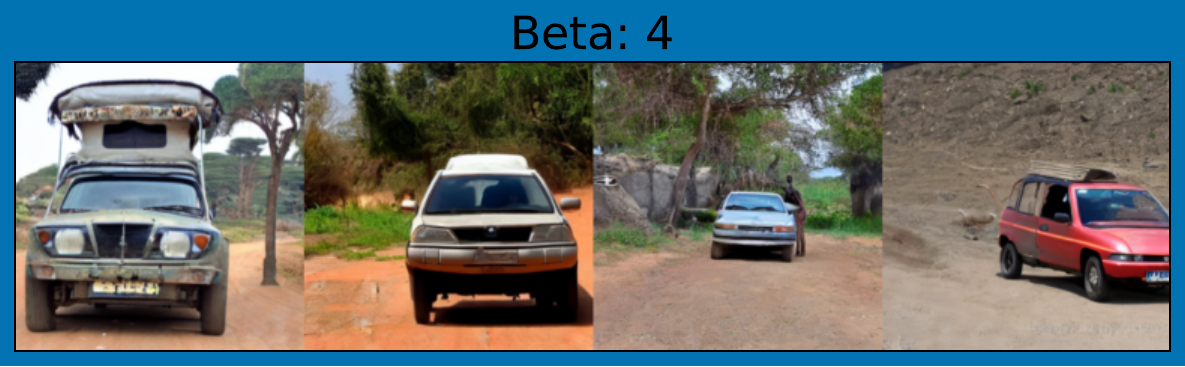}
        \end{subfigure}
        \begin{subfigure}[b]{0.32\textwidth}
             \includegraphics[width=0.99\textwidth]{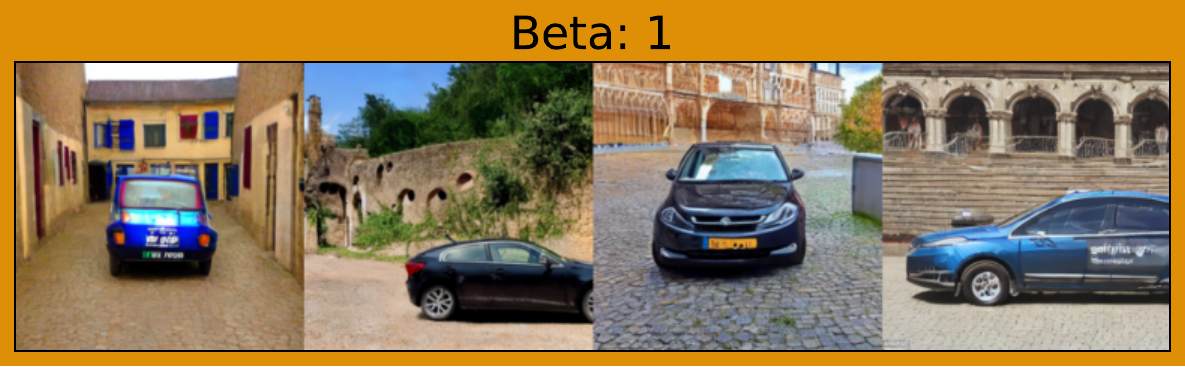}
             \includegraphics[width=0.99\textwidth]{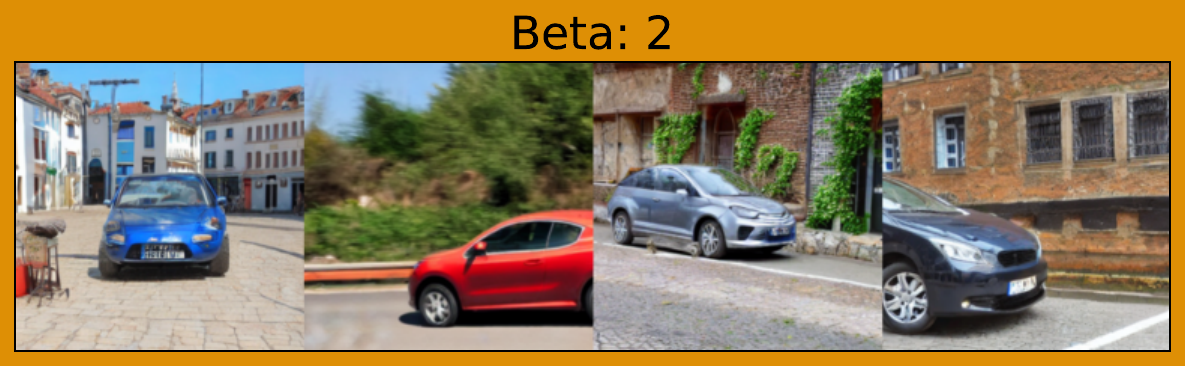}
            \includegraphics[width=0.99\textwidth]{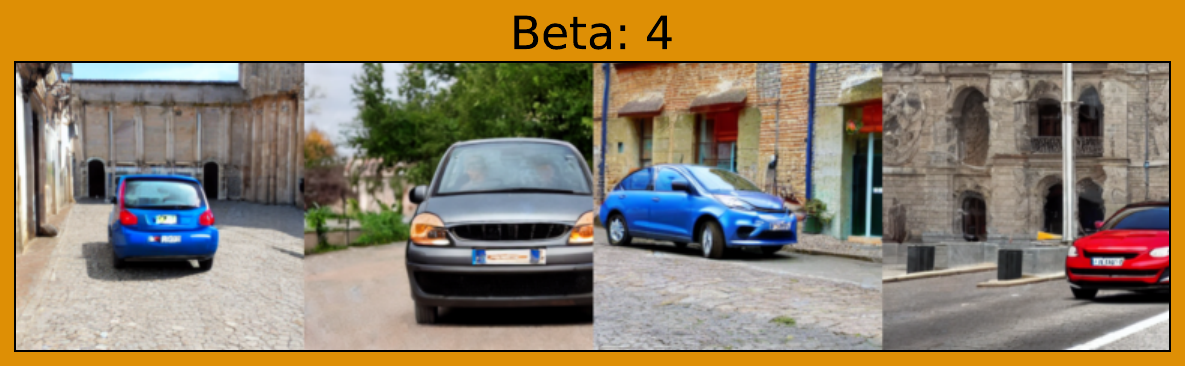}
        \end{subfigure}
        \begin{subfigure}[b]{0.32\textwidth}
             \includegraphics[width=0.99\textwidth]{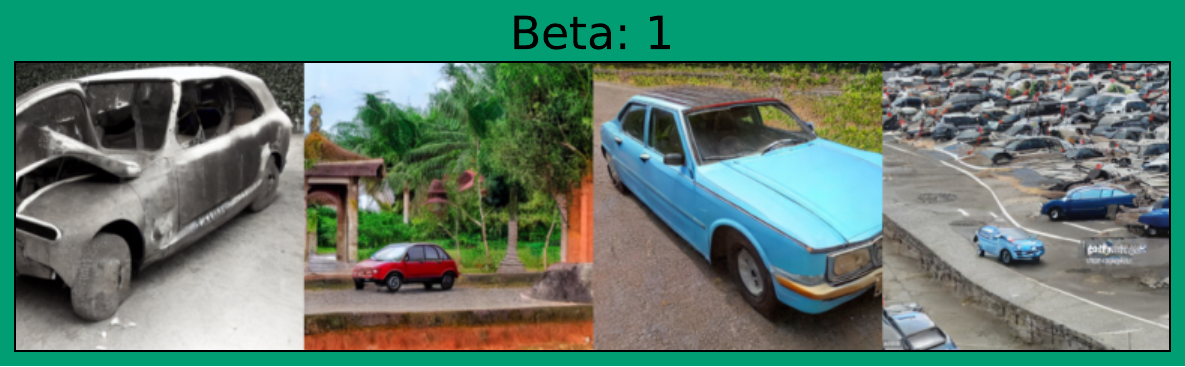}
             \includegraphics[width=0.99\textwidth]{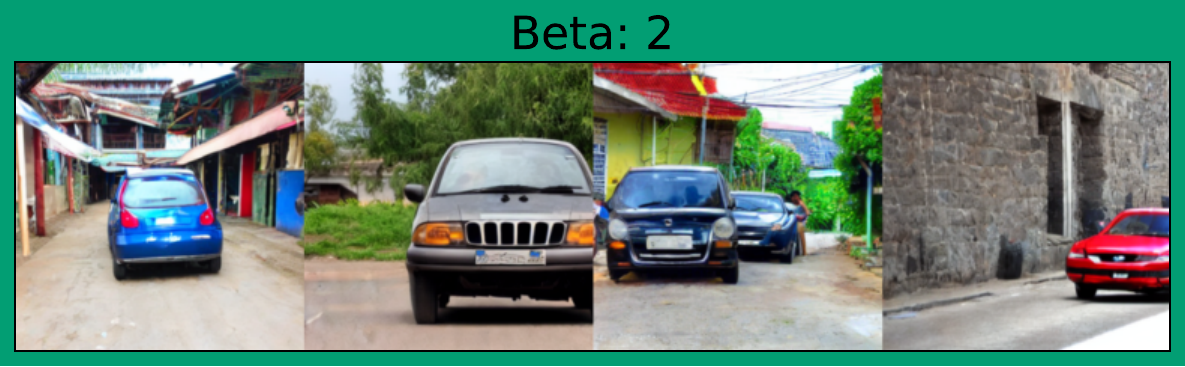}
            \includegraphics[width=0.99\textwidth]{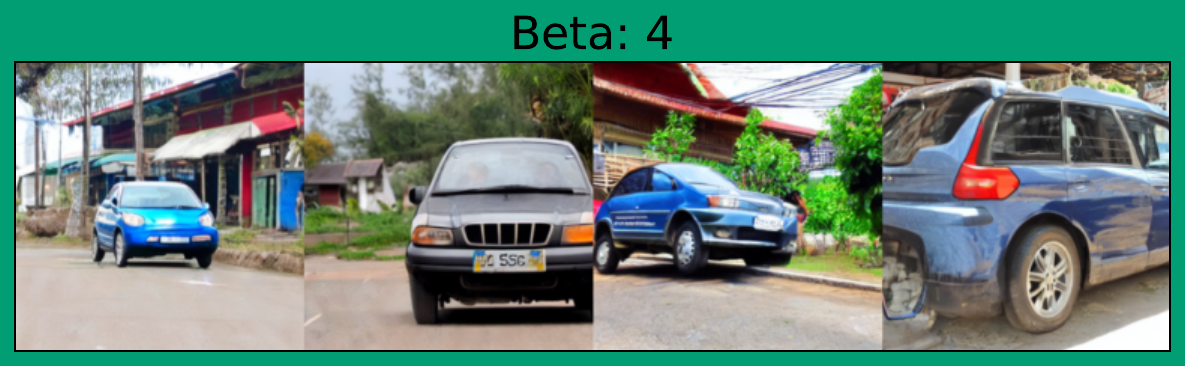}
        \end{subfigure}
    \caption{Generated images of cars with c-\methodnameshort. The same seed is used for a given column, and the box colors indicate images pertaining to \textcolor{blue}{Africa}, \textcolor{orange}{Europe}, and \textcolor{darkgreen}{Southeast Asia}.}
    \label{fig:weight}
\end{figure}

Furthermore, we examine the strength of exemplar images. 
In Figure \ref{fig:weight}, we show examples where increased weighting of exemplar images can coincide with more directly front-facing views of cars, which are uncommon in the generated images without \methodnameshort (see Figure \ref{fig:res_visuals} and Figure \ref{fig:diversity_issue_example}).

\begin{figure}[!t]
    \begin{subfigure}[b]{0.99\textwidth}
        \begin{subfigure}[b]{\textwidth}
            \centering
            \includegraphics[width=0.49\textwidth]{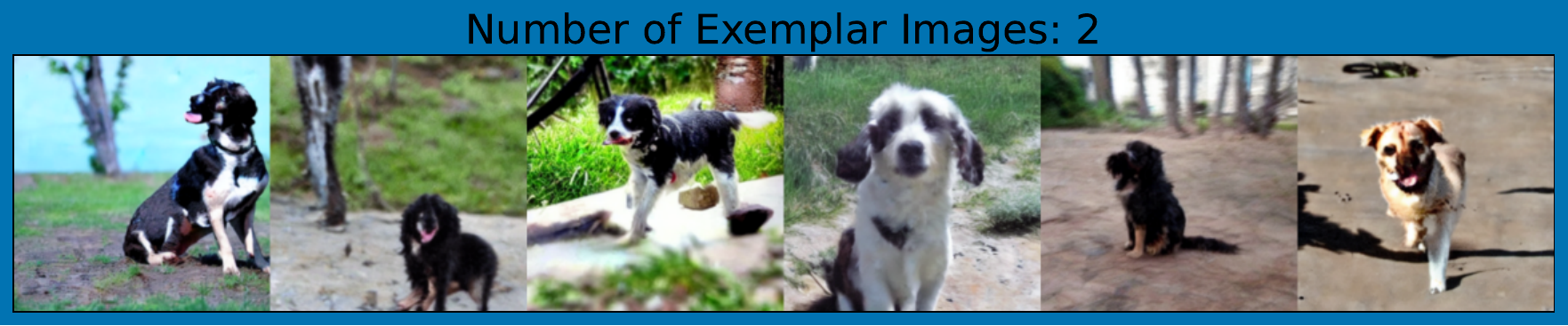}  
            \includegraphics[width=0.49\textwidth]{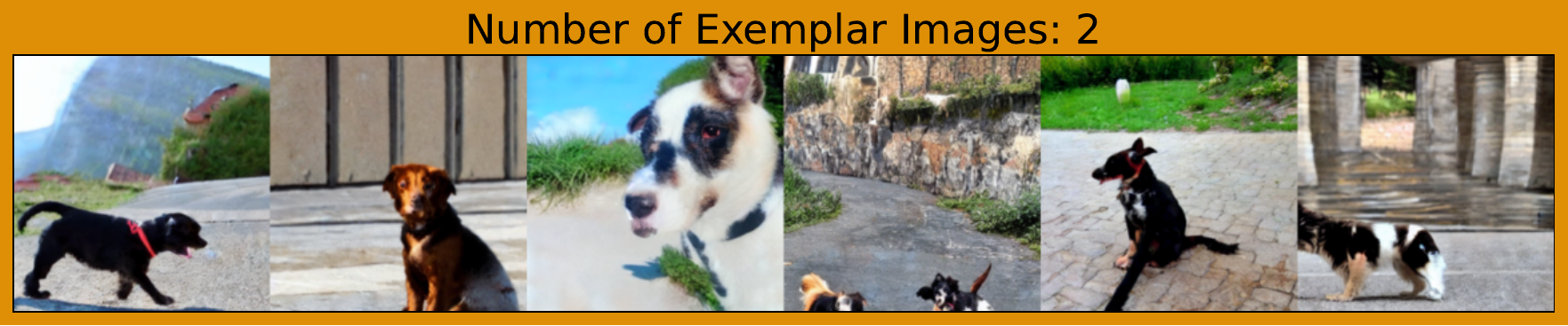}  \includegraphics[width=0.49\textwidth]{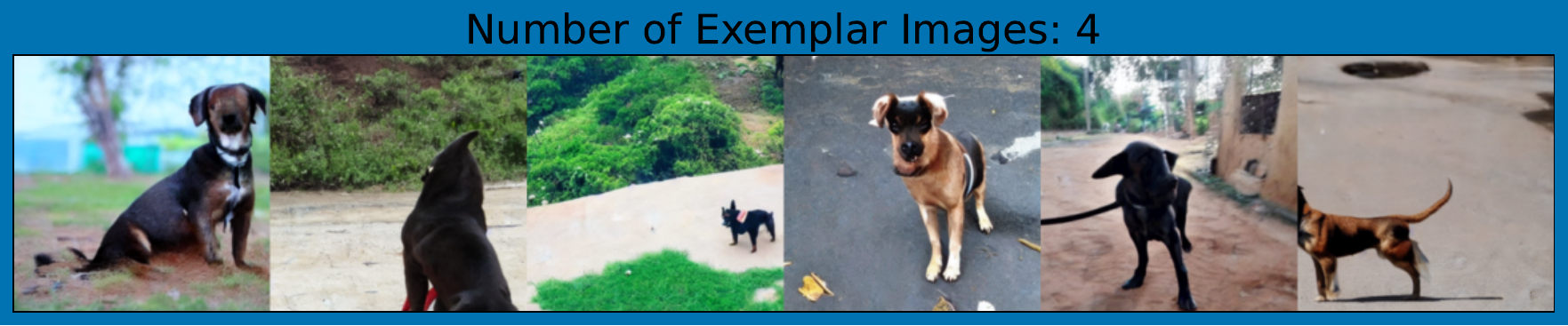}  
            \includegraphics[width=0.49\textwidth]{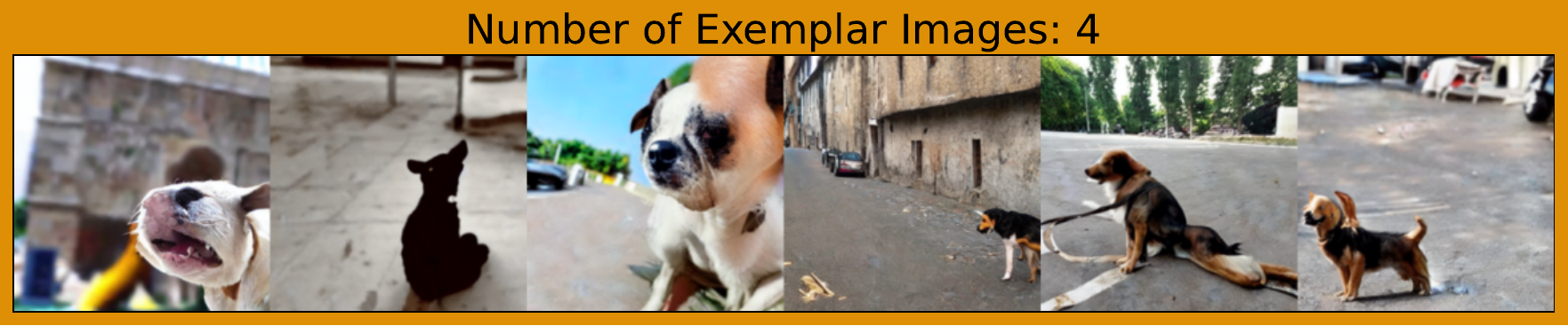}  \includegraphics[width=0.49\textwidth]{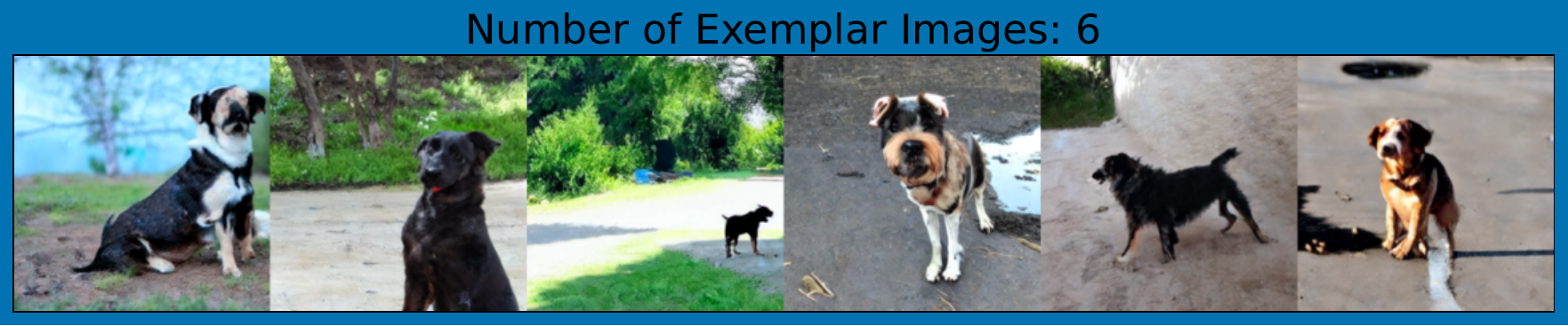}  
            \includegraphics[width=0.49\textwidth]{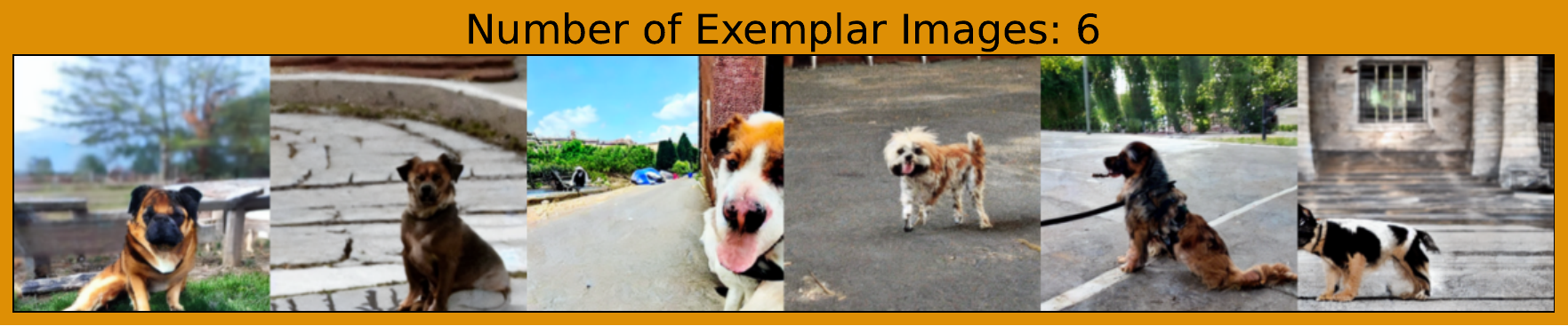}  
            \includegraphics[width=0.49\textwidth]{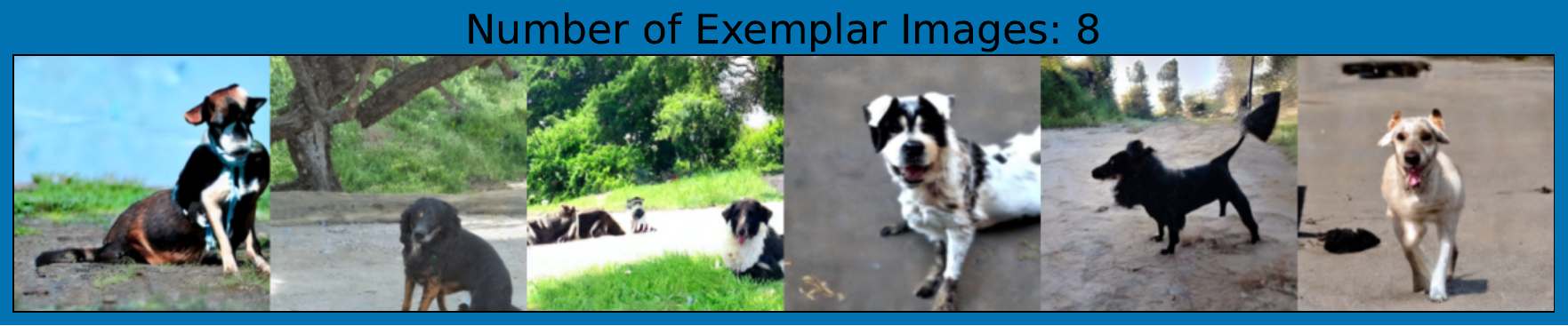}  
            \includegraphics[width=0.49\textwidth]{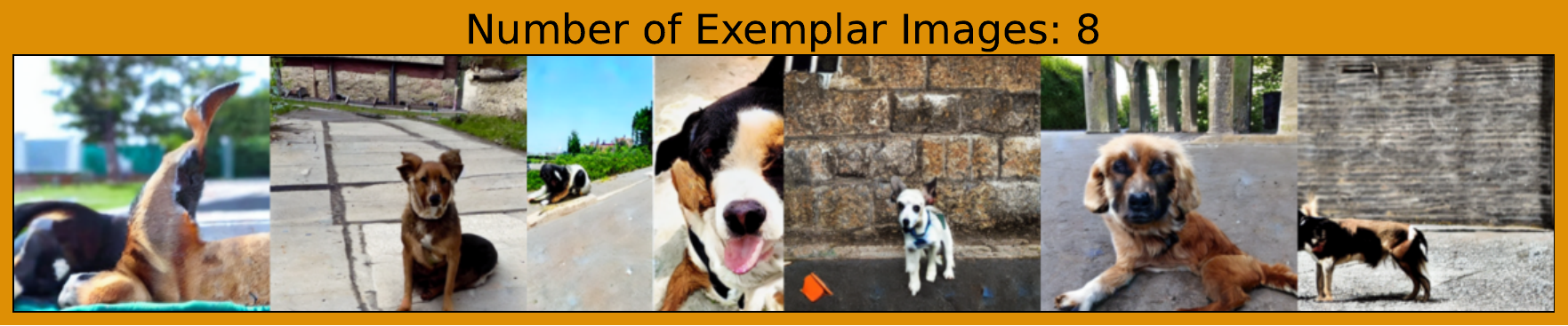}  
            \includegraphics[width=0.49\textwidth]{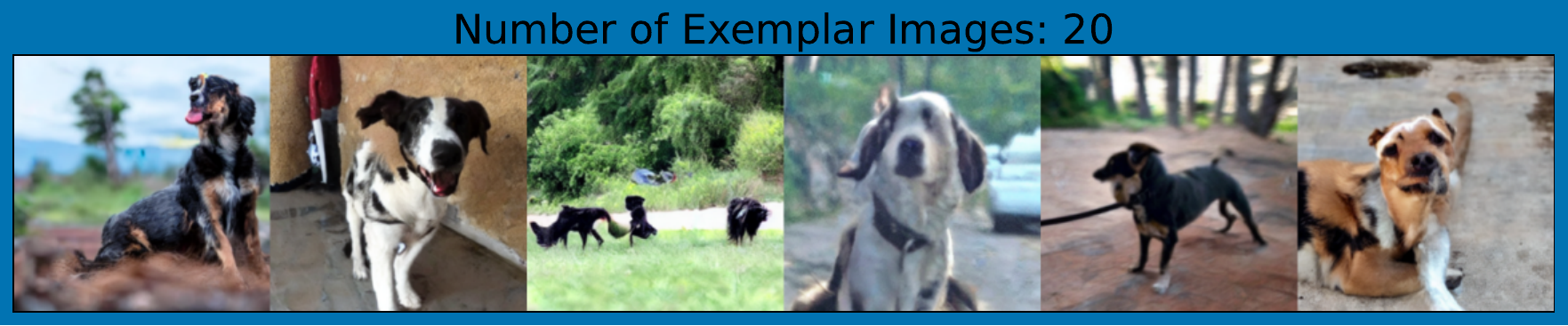}  
            \includegraphics[width=0.49\textwidth]{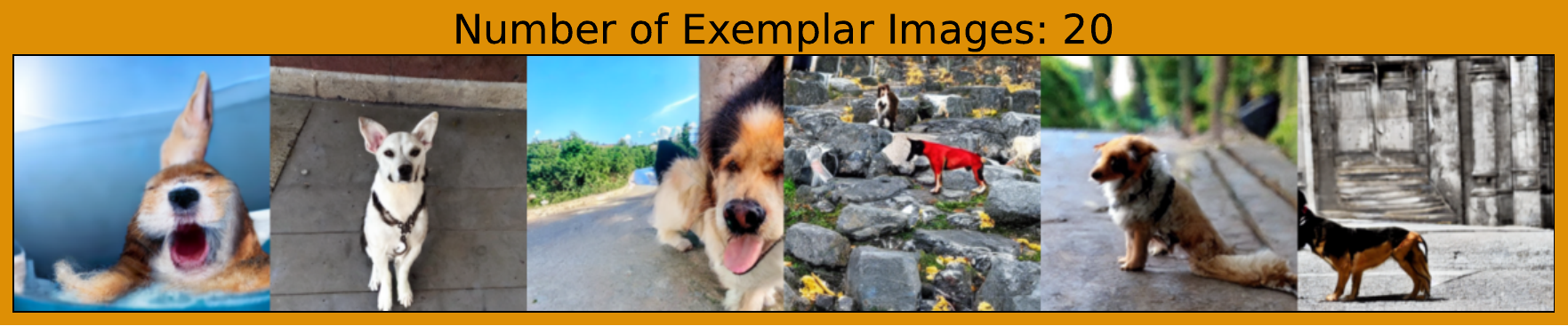}  
        \end{subfigure} 
    \end{subfigure}
    \caption{
    Generated images of dogs in Africa (\textbf{Left}) and Europe (\textbf{Right}).
    The same seed is used for a given column.
    }
    \label{fig:exemplar_quantity}
\end{figure}

Finally, we found that using fewer exemplar images tends to help with diversity while increasing this number helps with quality, although these trends appear minor in some cases.
We show visual examples in Figure \ref{fig:exemplar_quantity} and find that trends in quality and consistency for a single seed can also reverse as the number of images increases, first improving, then worsening (or vice versa).
We note that while in other experiments we perform a hyperparameter search over CLIP guidance scales, this experiment is conducted exclusively with CLIP guidance scale of 2, which may explain some of the variability.

\subsection{Human study}
\label{human_study}
\begin{figure}[h]
    \centering
    \begin{subfigure}{0.75\columnwidth}
    \includegraphics[width=\textwidth]{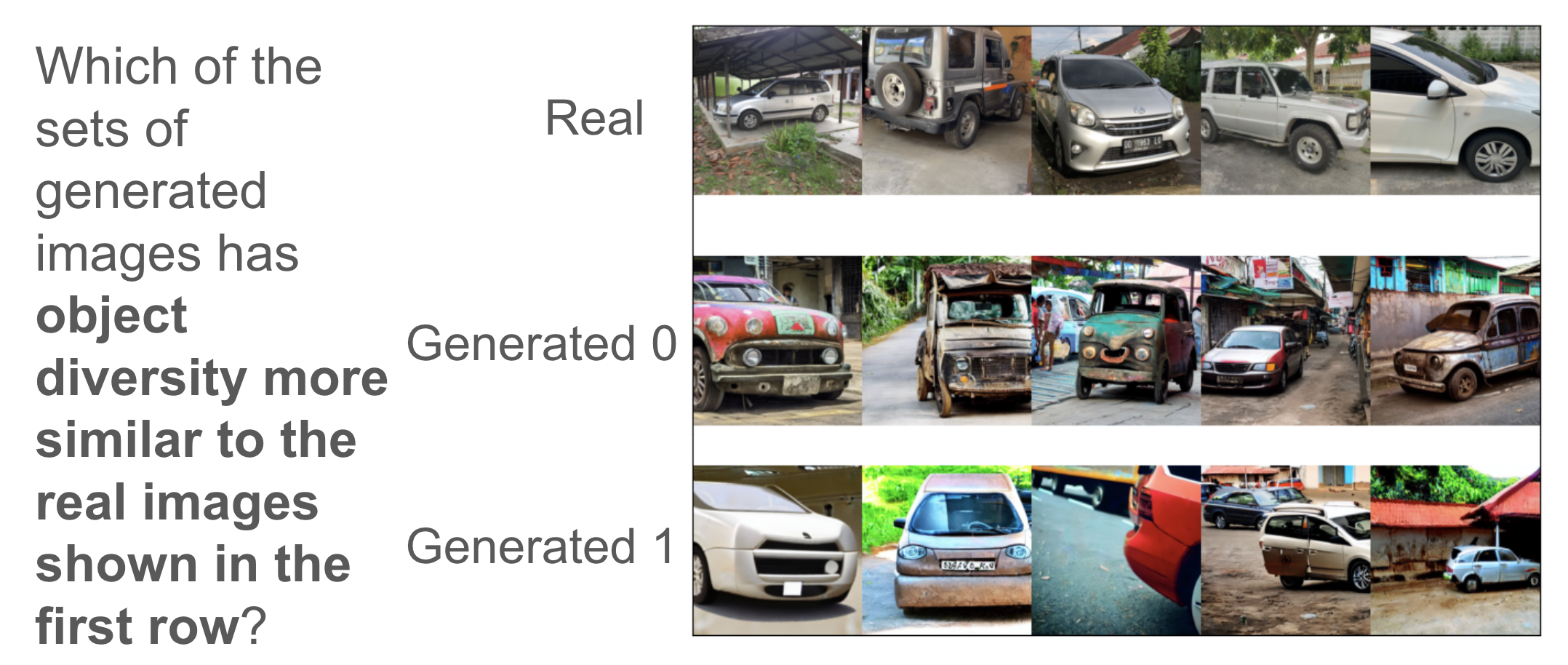}
    \end{subfigure}
    \caption{User study comparing real image diversity to baseline LDM (Gen 0) and C-VSG (Gen 1)} 
    \label{fig:userstudyex} 
\end{figure}

\begin{figure}[h]
    \centering
    \begin{subfigure}{0.34\columnwidth}
        \includegraphics[width=\textwidth]{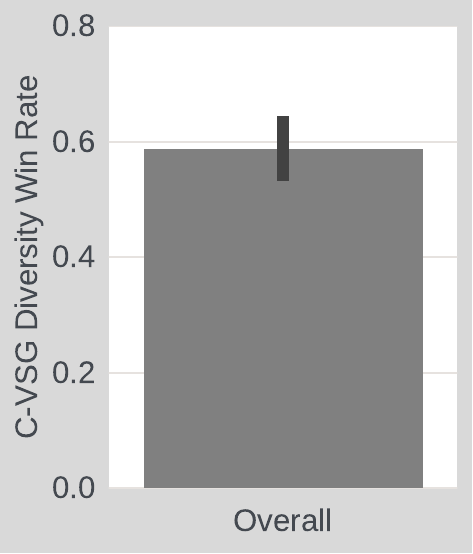}
    \end{subfigure}
    \begin{subfigure}{0.64\columnwidth}
        \includegraphics[width=\textwidth]{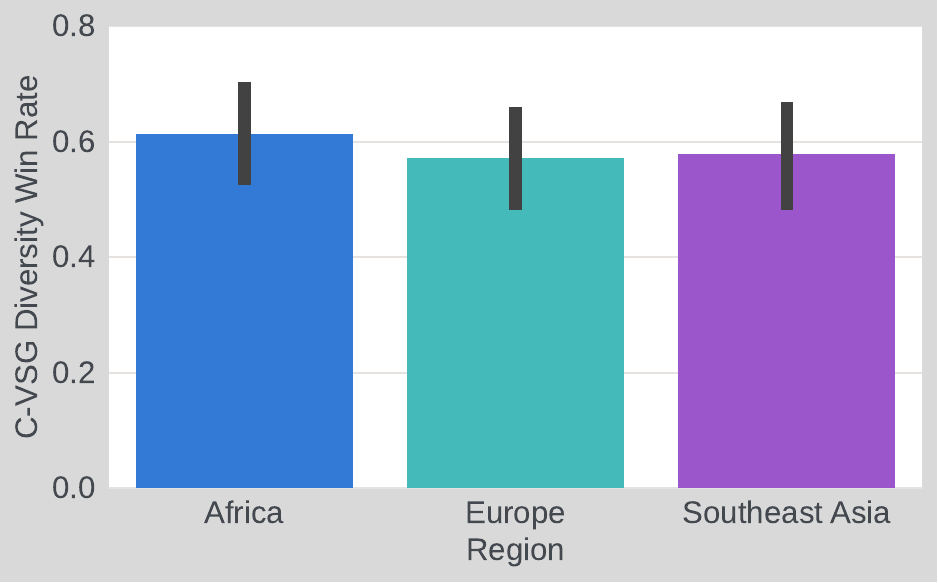}
    \end{subfigure}
    \caption{People perceive diversity improvements when c-\methodnameshort is used.}
    \label{fig:userstudyres} 
\end{figure}

To better understand the efficacy of c-\methodnameshort, we perform a preliminary human study. 
We collected 900 annotations corresponding to the diversity of 1200 generated and 600 real images from individuals located in five countries, allowing us to measure subjective perspectives about c-\methodnameshort. 
Fig \ref{fig:userstudyex} shows an example of the images provided to the participants and Figure \ref{fig:userstudyres} depicts results.
Our study finds that annotators consider images generated with c-VSG contain diversity more similar to real images 1.44$\times$ as often as the baseline LDM.
In addition, c-VSG has a win-rate of up to 72.6\% for individual objects (\textit{e.g.} \textit{car}) and 61.4\% for regions (\textit{e.g.} \textit{Africa}).

As a cautionary note, this study is among the first human studies pertaining to generated image diversity and different task layout designs may yield variable results.
We leave research into methods for more robust human studies of generated image diversity to future work.

\subsection{Additional Samples for Each Method}\label{app:samples}
\begin{figure}[!h]
\centering
    \begin{subfigure}[b]{0.95\textwidth}
    \begin{subfigure}[b]{\textwidth}
            \centering
            \includegraphics[width=0.49\textwidth]{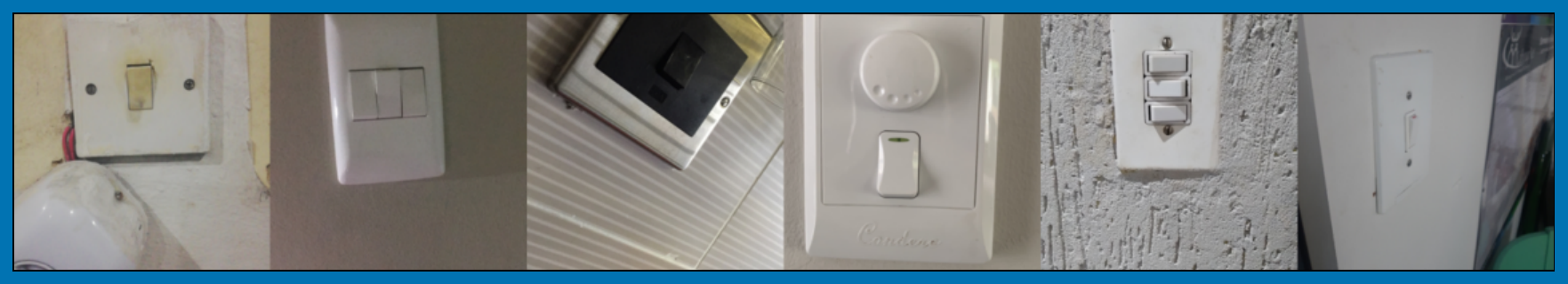}  
            \includegraphics[width=0.49\textwidth]{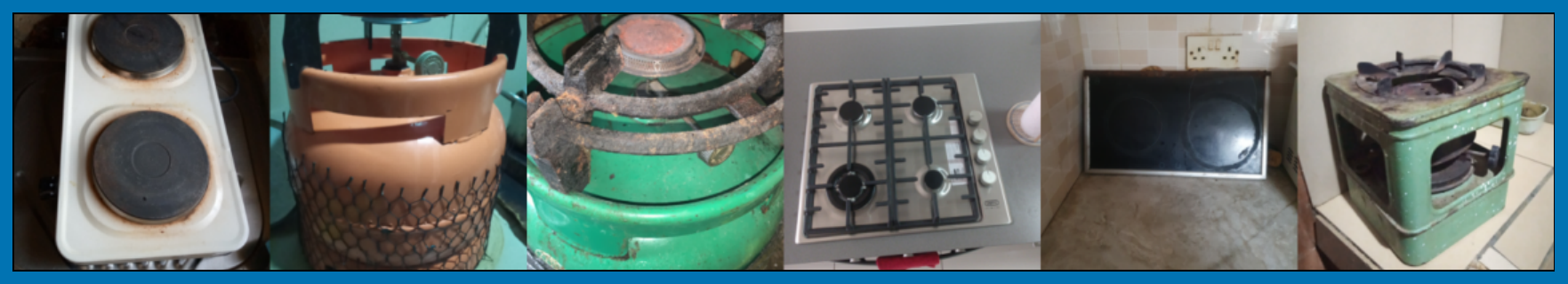}  \includegraphics[width=0.49\textwidth]{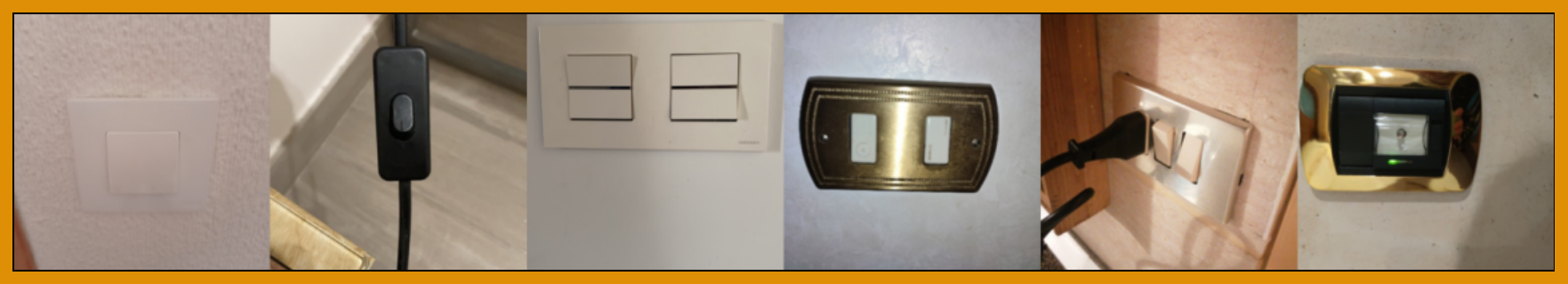}  
            \includegraphics[width=0.49\textwidth]{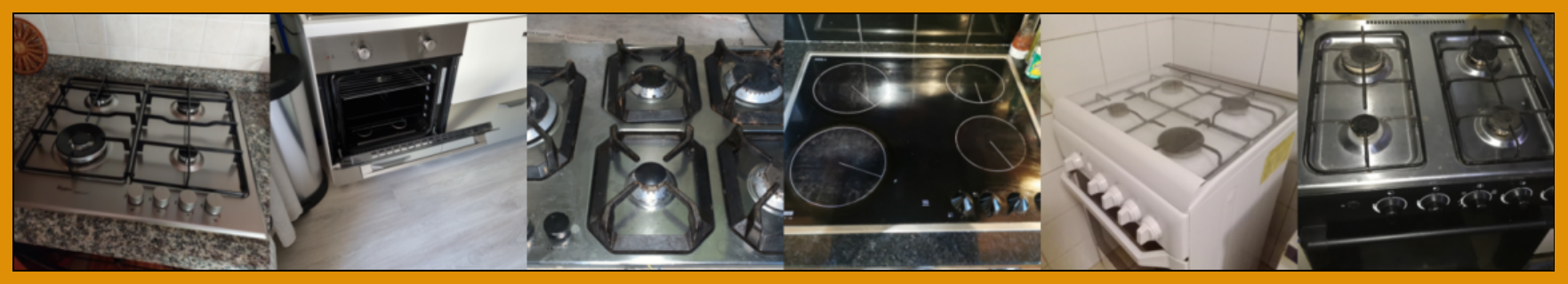}  \includegraphics[width=0.49\textwidth]{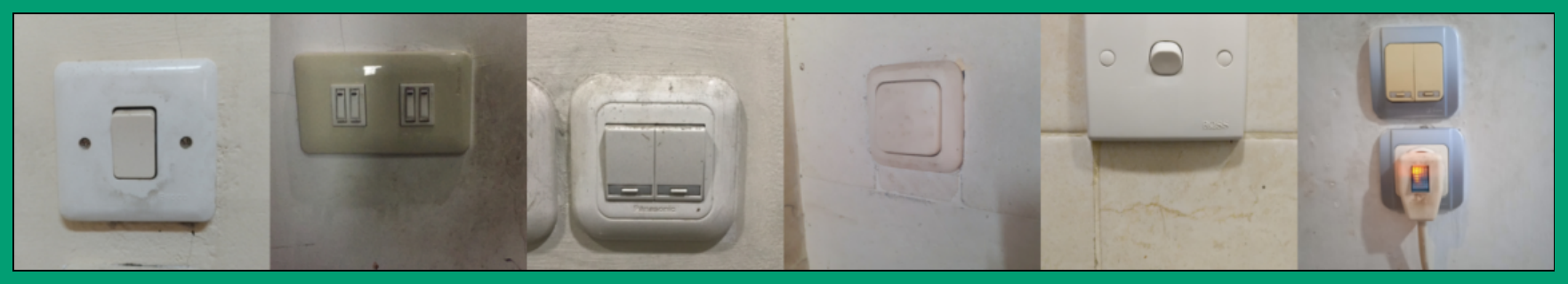}  \includegraphics[width=0.49\textwidth]{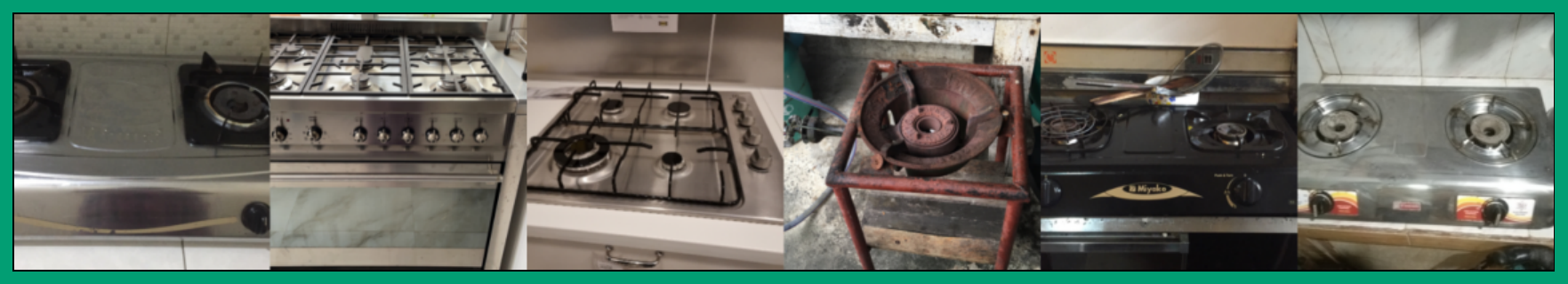}  
        \end{subfigure}
    \end{subfigure}
    \caption{
    Real images of light switches (\textbf{Left}) and stoves (\textbf{Right}).
    The box colors indicate images pertaining to \textcolor{blue}{Africa}, \textcolor{orange}{Europe}, and \textcolor{darkgreen}{Southeast Asia}.
    }
    \label{fig:res_compare_methods_real}
\end{figure}

\begin{figure}[!h]
\centering
    \begin{subfigure}[b]{0.8\textwidth}
    \begin{subfigure}[b]{\textwidth}
            \centering
            \includegraphics[width=0.49\textwidth]{gen_light_switch_Africa_2024_02_14_22_11_47_223859_0_6.pdf}  
            \includegraphics[width=0.49\textwidth]{gen_stove_Africa_2024_02_14_22_11_47_223859_0_6.pdf}  \includegraphics[width=0.49\textwidth]{gen_light_switch_Europe_2024_02_14_22_11_47_223859_0_6.pdf}  
            \includegraphics[width=0.49\textwidth]{gen_stove_Europe_2024_02_14_22_11_47_223859_0_6.pdf}  \includegraphics[width=0.49\textwidth]{gen_light_switch_Southeast_Asia_2024_02_14_22_11_47_223859_0_6.pdf}  \includegraphics[width=0.49\textwidth]{gen_stove_Southeast_Asia_2024_02_14_22_11_47_223859_0_6.pdf}  

            \caption{LDM baseline, No \methodnameshort, \ie \objreg baseline}
        \end{subfigure}
            \begin{subfigure}[b]{\textwidth}
            \centering
            \includegraphics[width=0.49\textwidth]{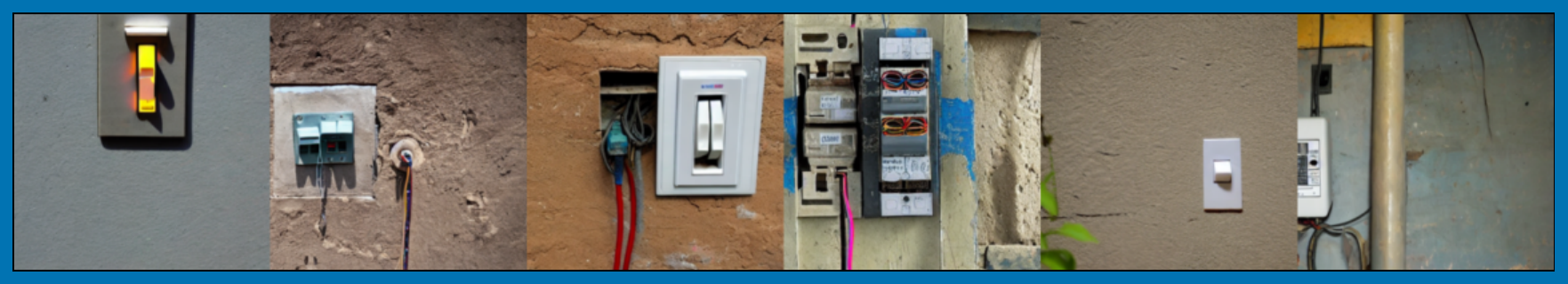} 
            \includegraphics[width=0.49\textwidth]{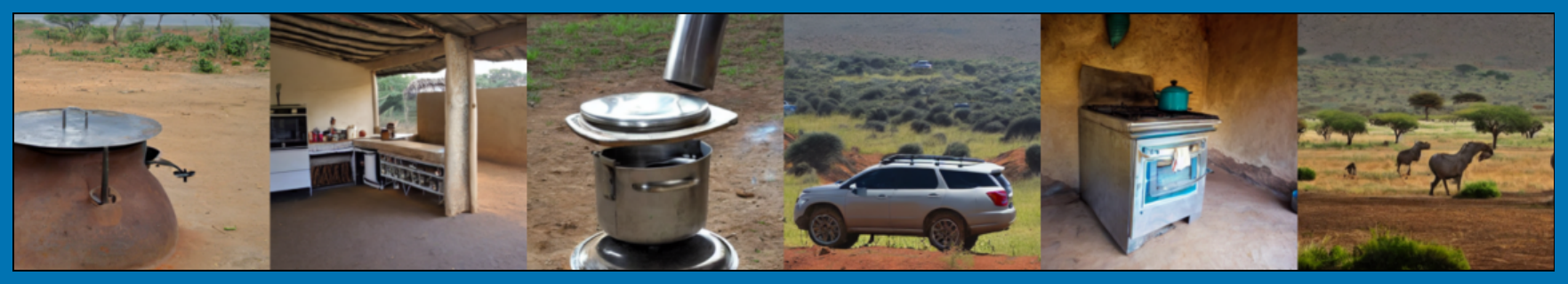}  \includegraphics[width=0.49\textwidth]{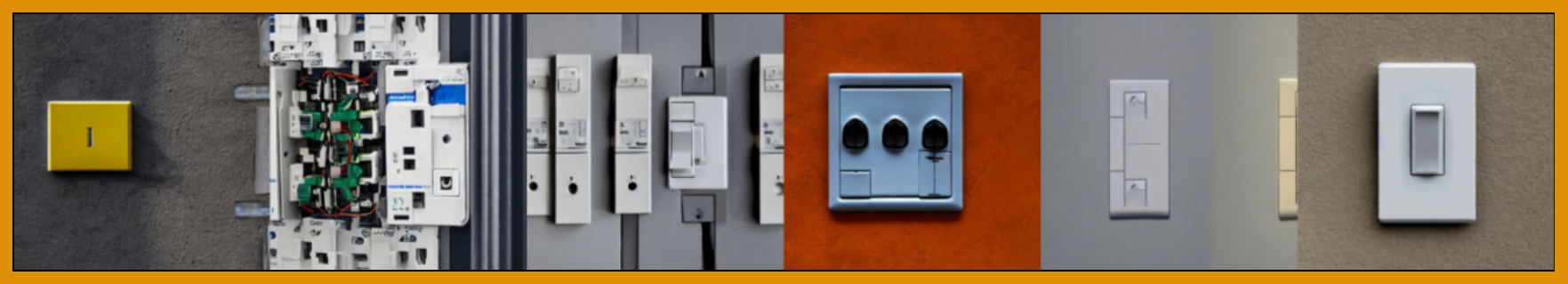}  
            \includegraphics[width=0.49\textwidth]{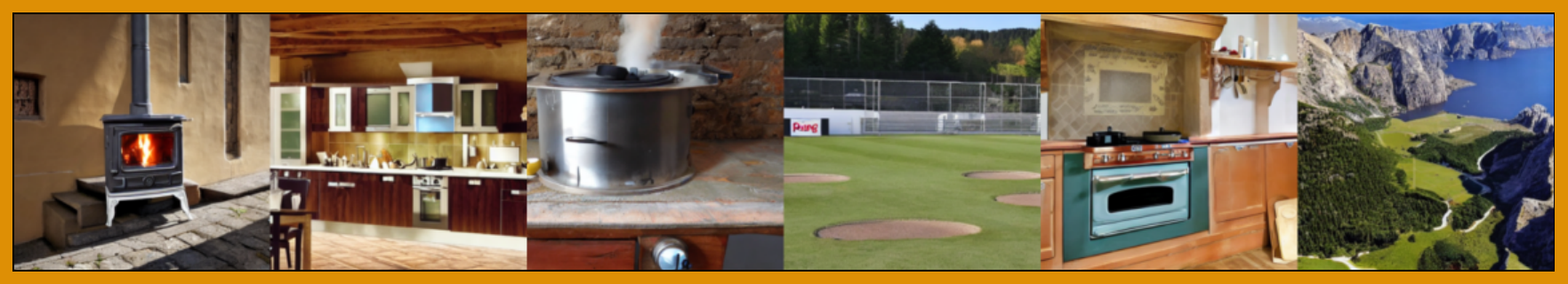}  \includegraphics[width=0.49\textwidth]{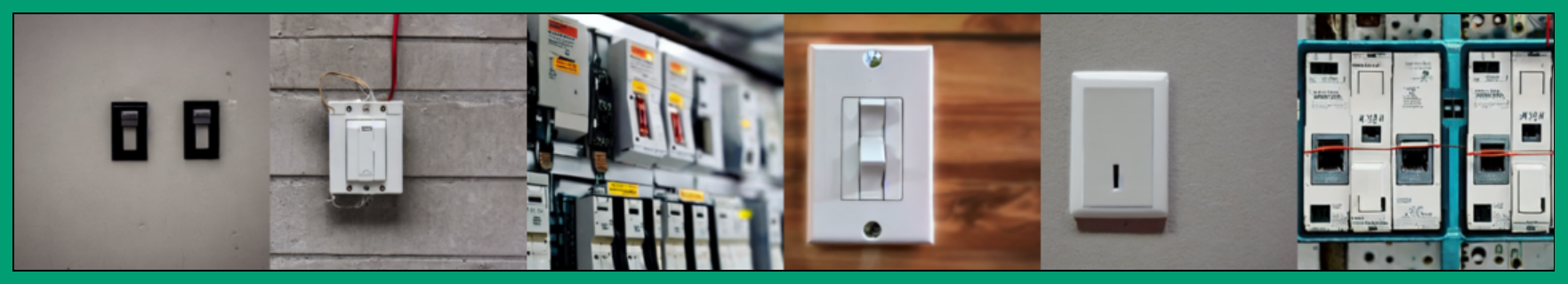}  
            \includegraphics[width=0.49\textwidth]{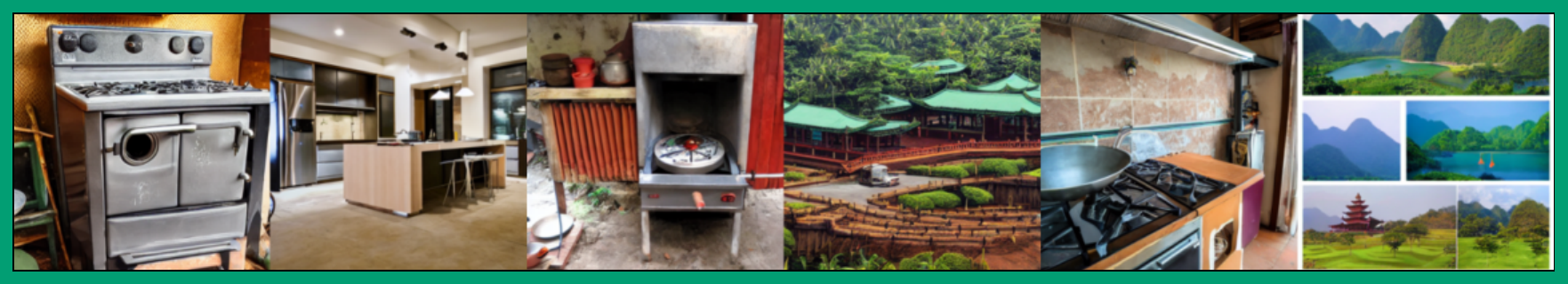}  
            \caption{Synonym Baseline}
        \end{subfigure}
            
            \begin{subfigure}[b]{\textwidth}
            \centering
            \includegraphics[width=0.49\textwidth]{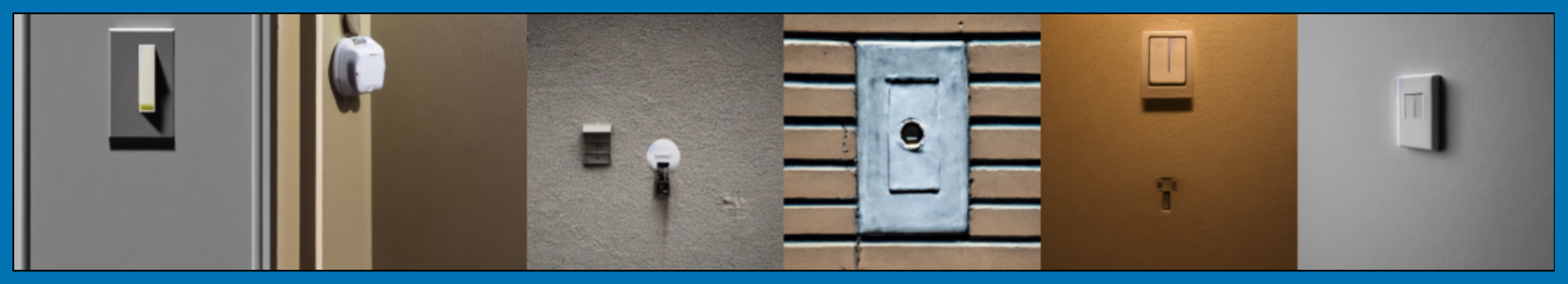} 
            \includegraphics[width=0.49\textwidth]{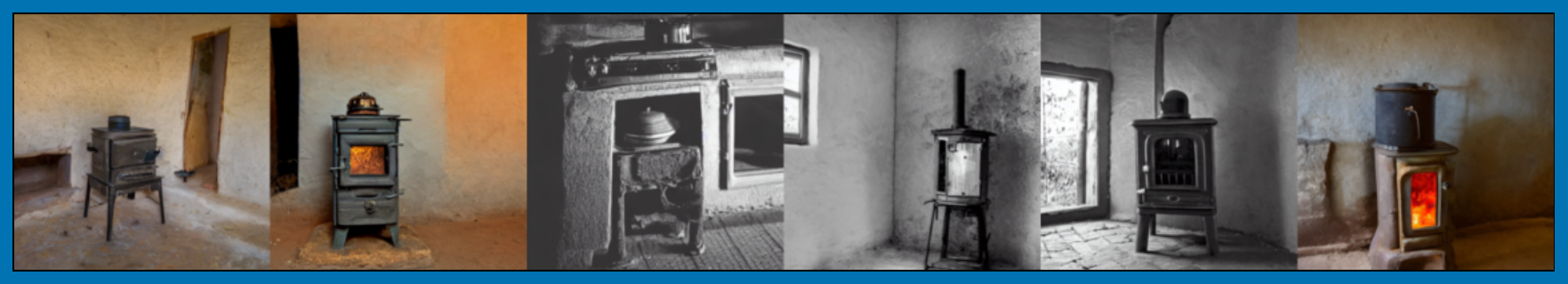}  \includegraphics[width=0.49\textwidth]{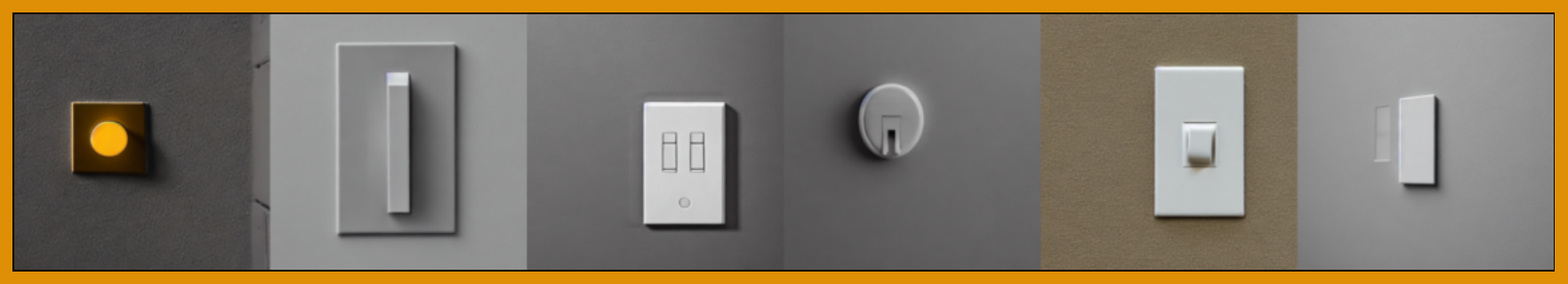}  
            \includegraphics[width=0.49\textwidth]{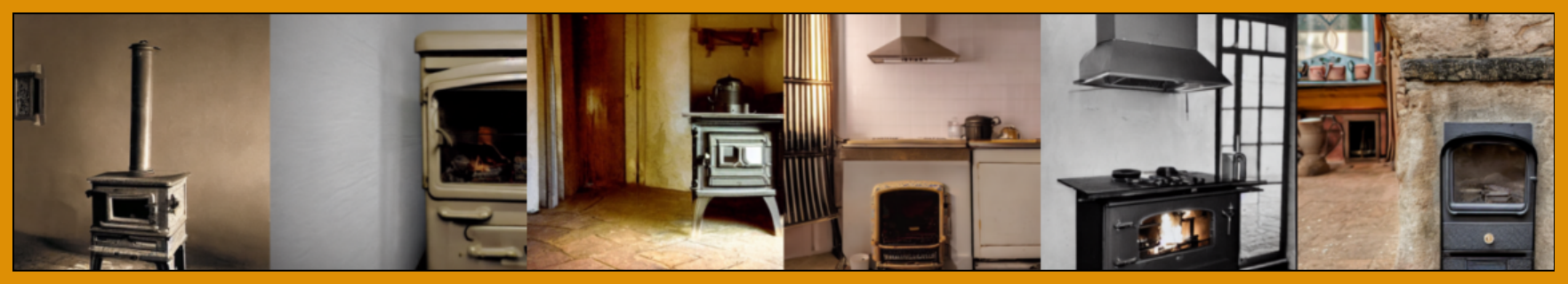}  \includegraphics[width=0.49\textwidth]{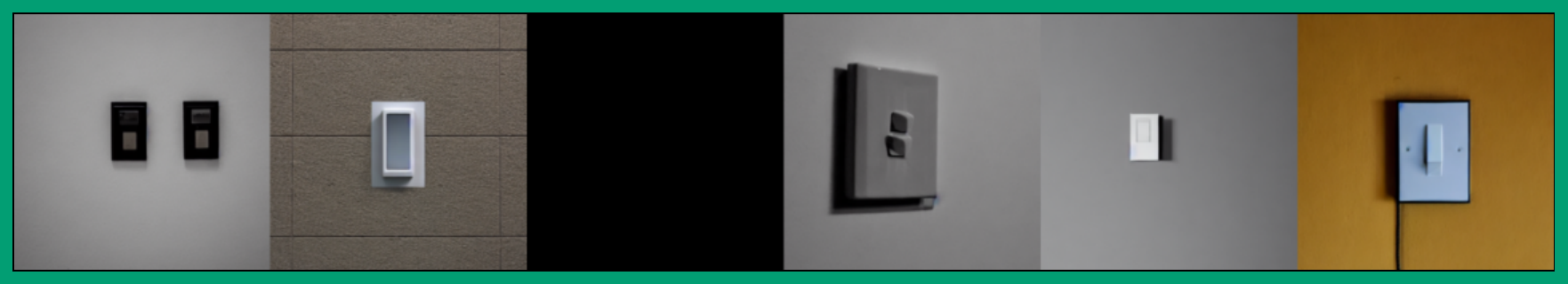}  
            \includegraphics[width=0.49\textwidth]{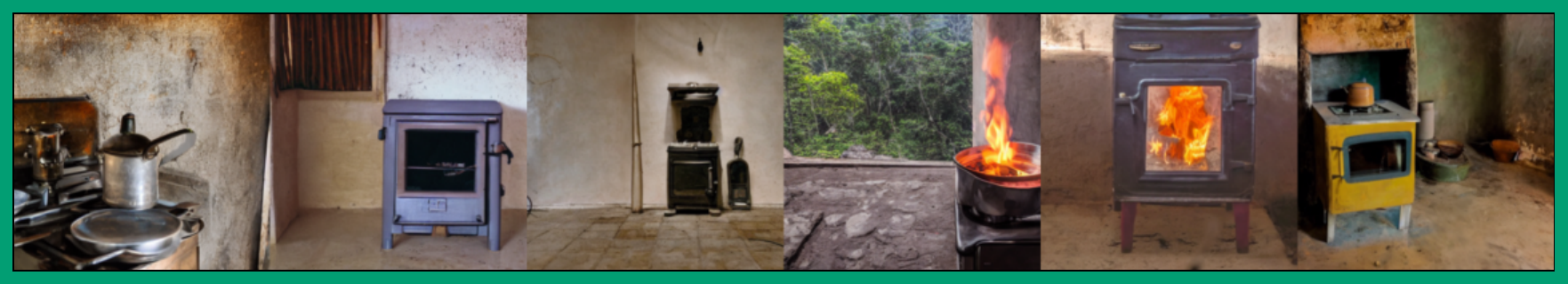}  
            \caption{Paraphrase Baseline}
        \end{subfigure}

    \end{subfigure}
    \caption{
    Generated images of light switches (\textbf{Left}) and stoves (\textbf{Right}).
    The box colors indicate images pertaining to \textcolor{blue}{Africa}, \textcolor{orange}{Europe}, and \textcolor{darkgreen}{Southeast Asia}.
    }
    \label{fig:res_compare_methods_noguidanace}
\end{figure}

\begin{figure}[]
\centering
    \begin{subfigure}[b]{0.8\textwidth}
    \begin{subfigure}[b]{\textwidth}
            \centering
            \includegraphics[width=0.49\textwidth]{gen_light_switch_Africa_2024_02_14_22_11_47_223859_0_6.pdf}  
            \includegraphics[width=0.49\textwidth]{gen_stove_Africa_2024_02_14_22_11_47_223859_0_6.pdf}  \includegraphics[width=0.49\textwidth]{gen_light_switch_Europe_2024_02_14_22_11_47_223859_0_6.pdf}  
            \includegraphics[width=0.49\textwidth]{gen_stove_Europe_2024_02_14_22_11_47_223859_0_6.pdf}  \includegraphics[width=0.49\textwidth]{gen_light_switch_Southeast_Asia_2024_02_14_22_11_47_223859_0_6.pdf}  \includegraphics[width=0.49\textwidth]{gen_stove_Southeast_Asia_2024_02_14_22_11_47_223859_0_6.pdf}  

            \caption{LDM baseline, No \methodnameshort, \ie \objreg baseline}
        \end{subfigure}
        \begin{subfigure}[b]{\textwidth}
            \centering
            \includegraphics[width=0.49\textwidth]{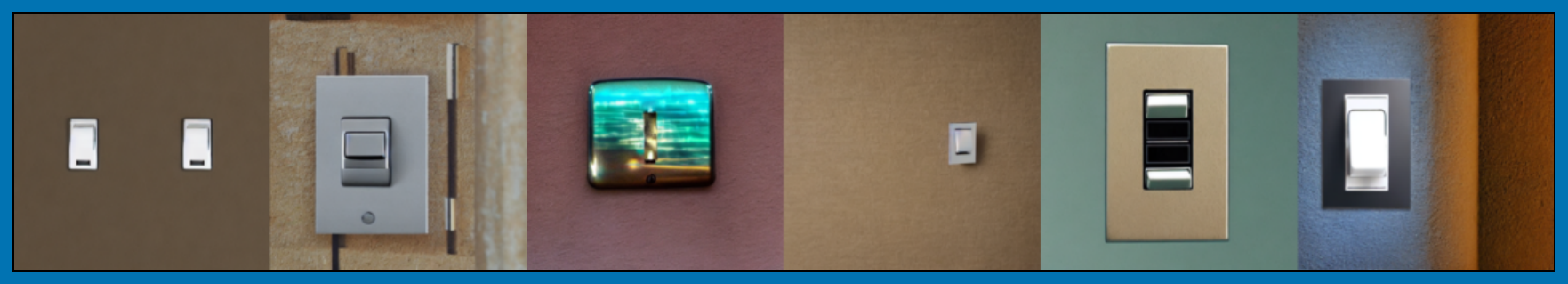}  
            \includegraphics[width=0.49\textwidth]{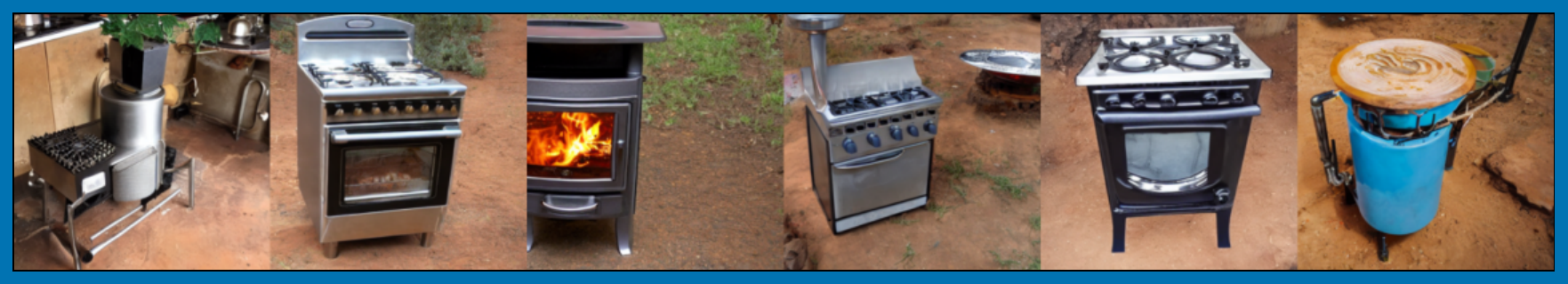}  \includegraphics[width=0.49\textwidth]{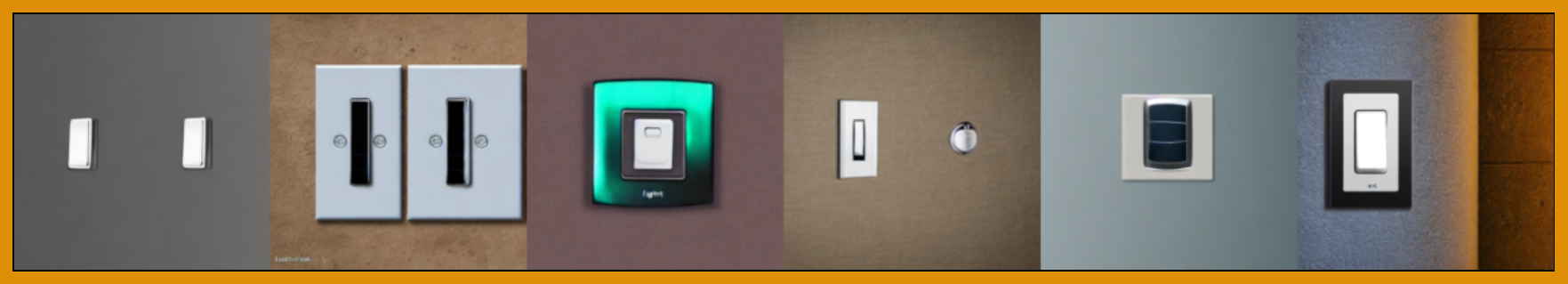}  
            \includegraphics[width=0.49\textwidth]{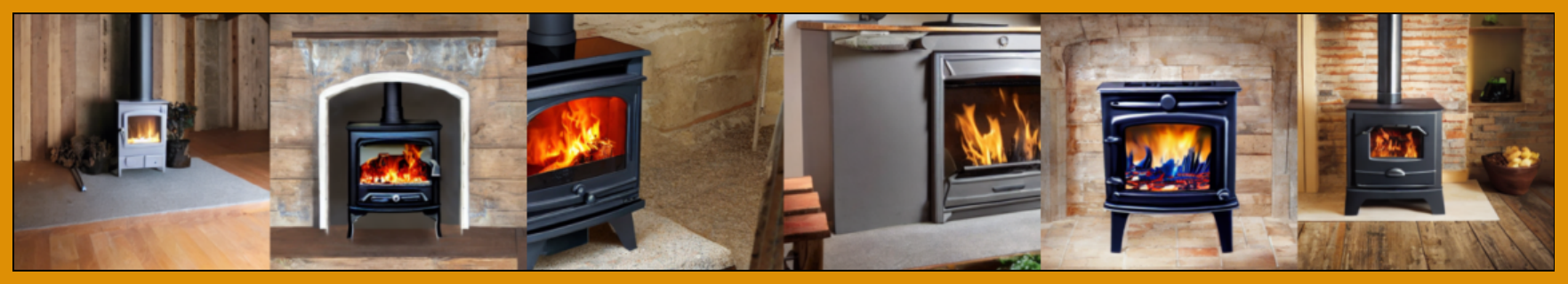}  \includegraphics[width=0.49\textwidth]{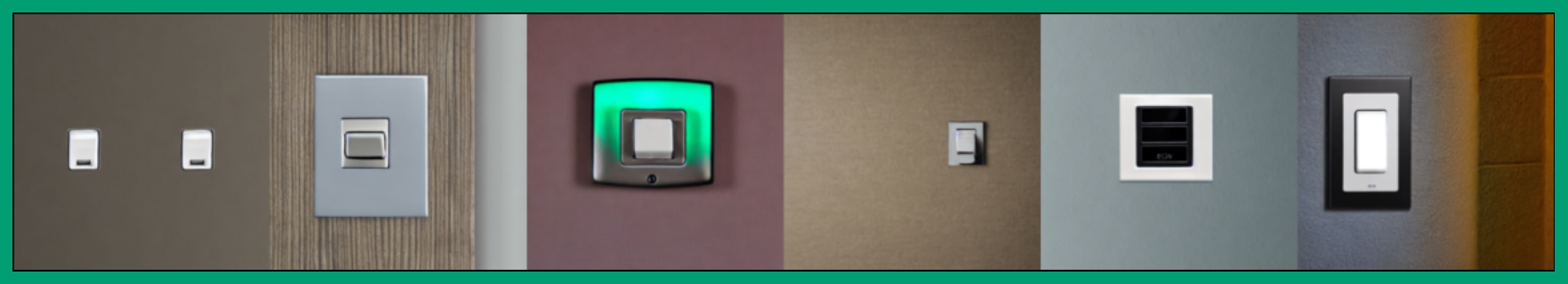}  \includegraphics[width=0.49\textwidth]{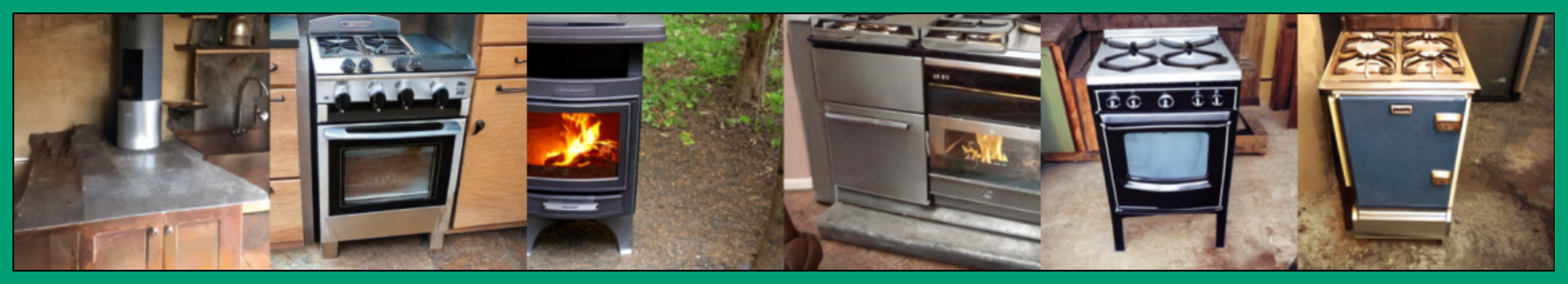}  
            \caption{Semantic guidance}
        \end{subfigure} 
        \begin{subfigure}[b]{\textwidth}
            \centering
            \includegraphics[width=0.49\textwidth]{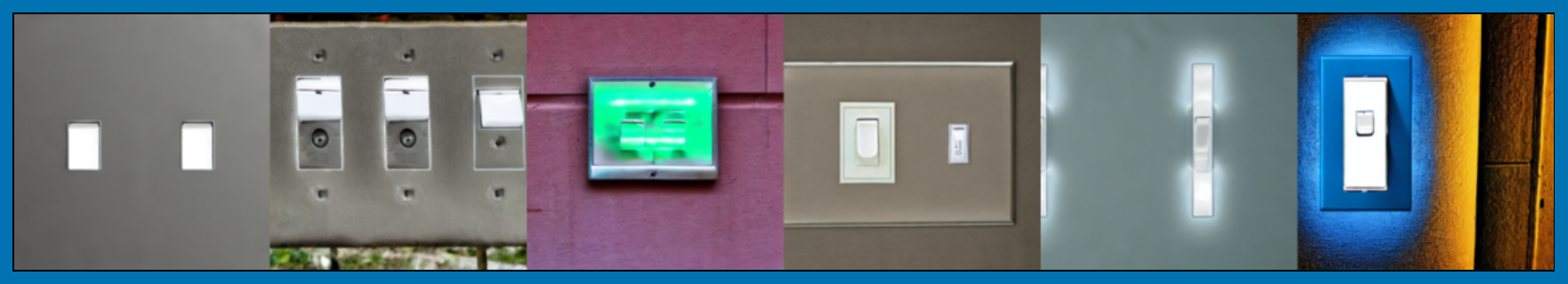}  
            \includegraphics[width=0.49\textwidth]{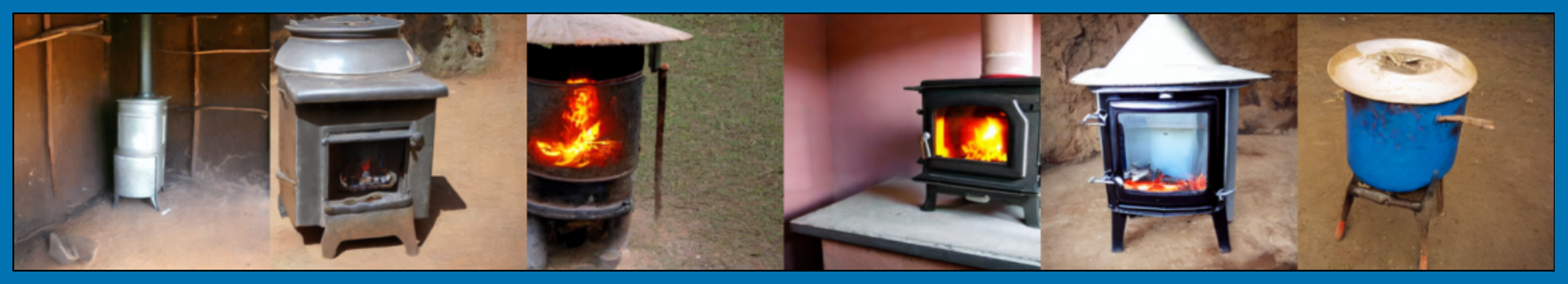}  \includegraphics[width=0.49\textwidth]{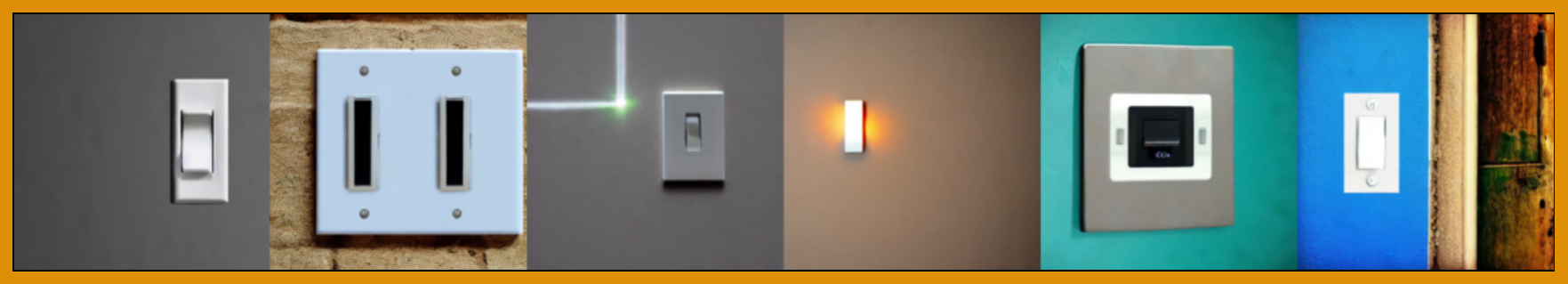}  
            \includegraphics[width=0.49\textwidth]{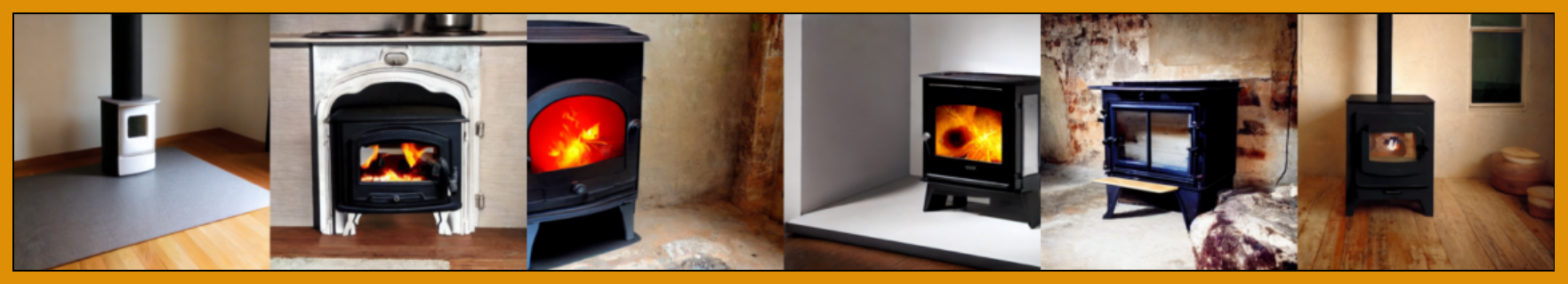}  \includegraphics[width=0.49\textwidth]{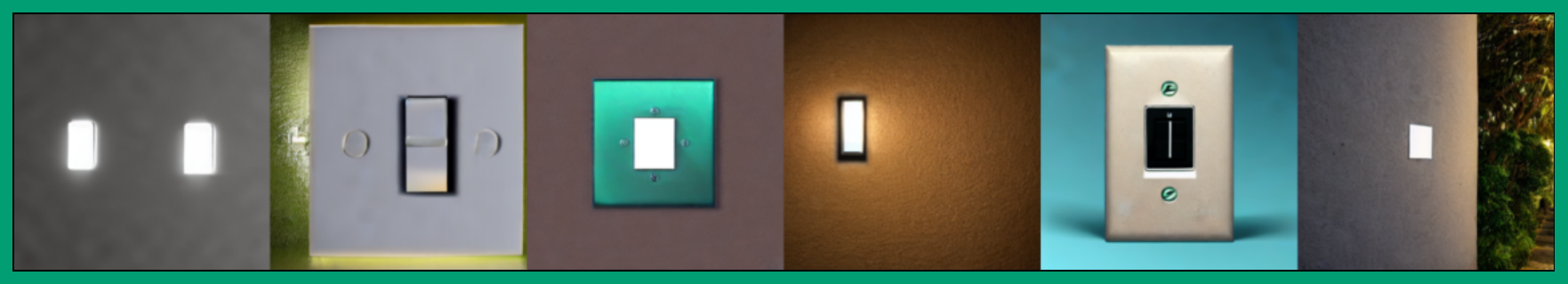}  \includegraphics[width=0.49\textwidth]{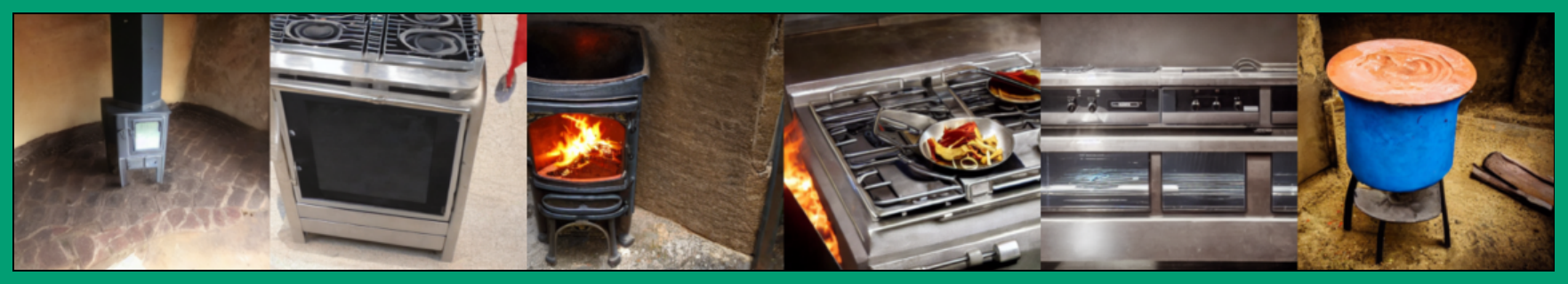}  
            \caption{CLIP Feedback Guidance - Loss}
        \end{subfigure} 
        \begin{subfigure}[b]{\textwidth}
            \centering
            \includegraphics[width=0.49\textwidth]{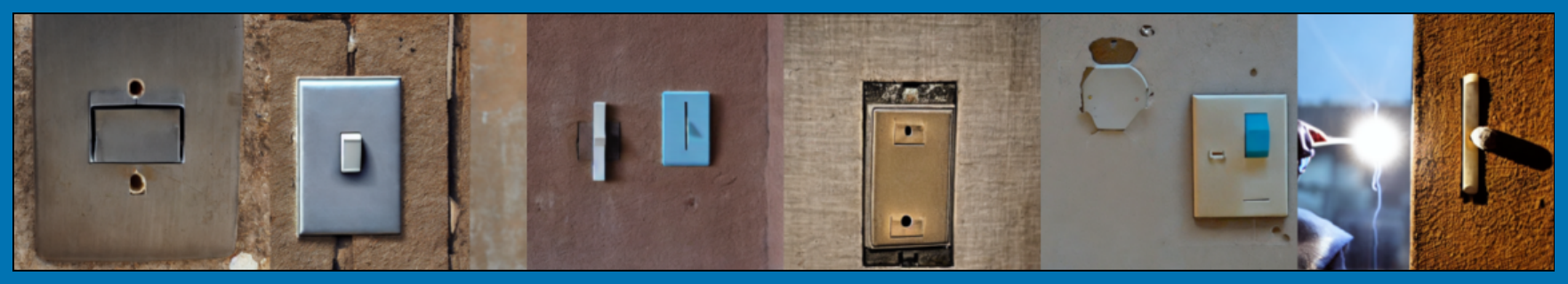}  
            \includegraphics[width=0.49\textwidth]{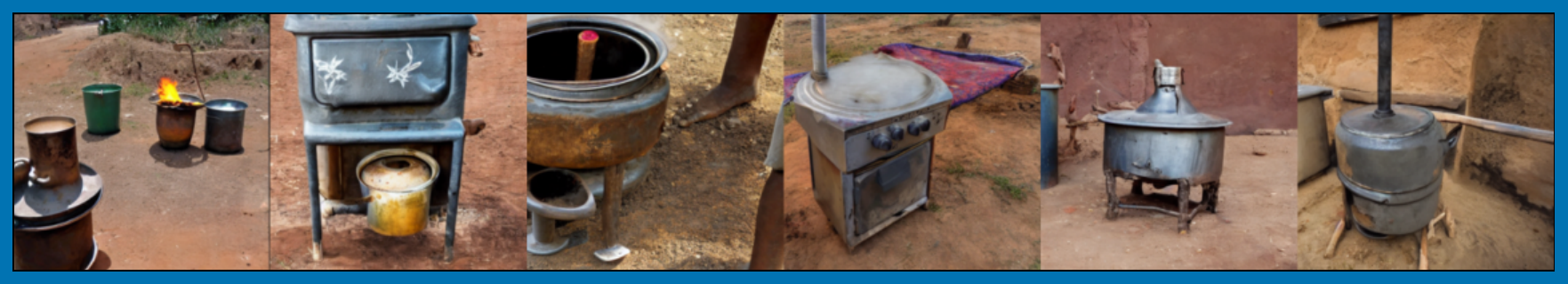}  \includegraphics[width=0.49\textwidth]{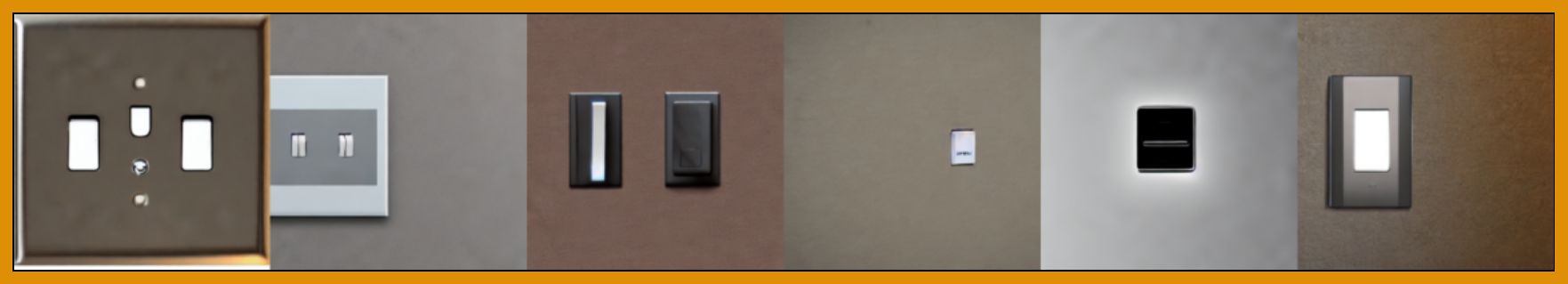}  
            \includegraphics[width=0.49\textwidth]{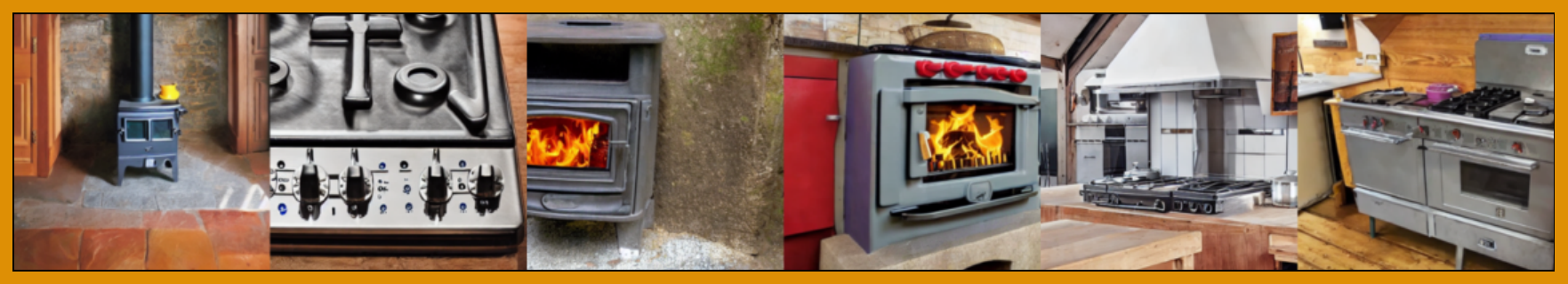}  \includegraphics[width=0.49\textwidth]{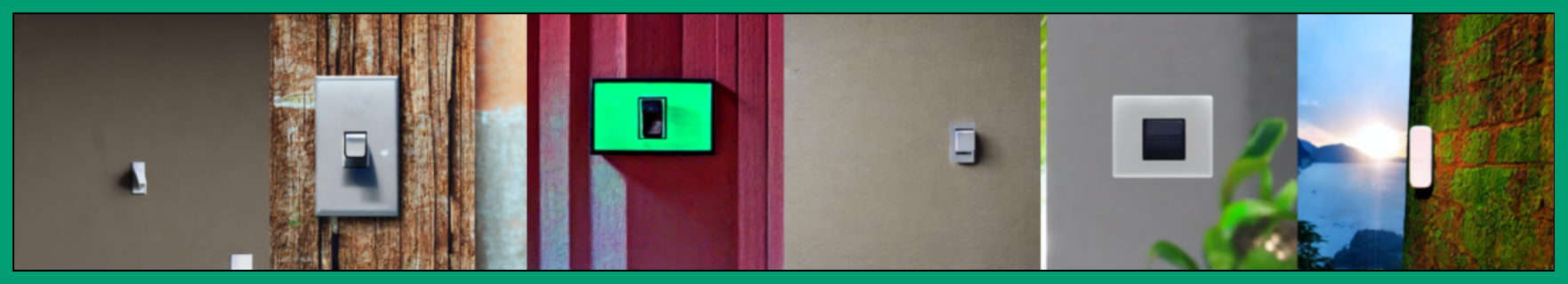}  \includegraphics[width=0.49\textwidth]{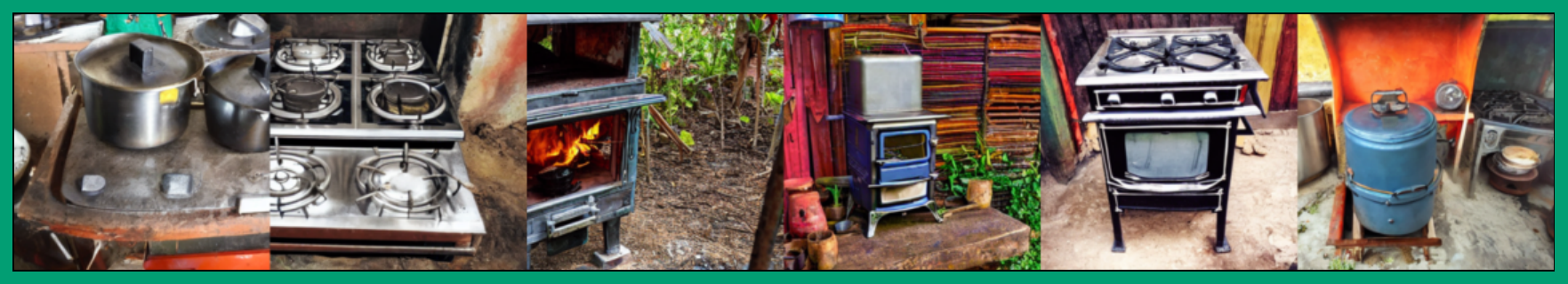}  
            \caption{CLIP Feedback Guidance - Entropy}
        \end{subfigure} 
        \begin{subfigure}[b]{\textwidth}
            \centering
            \includegraphics[width=0.49\textwidth]{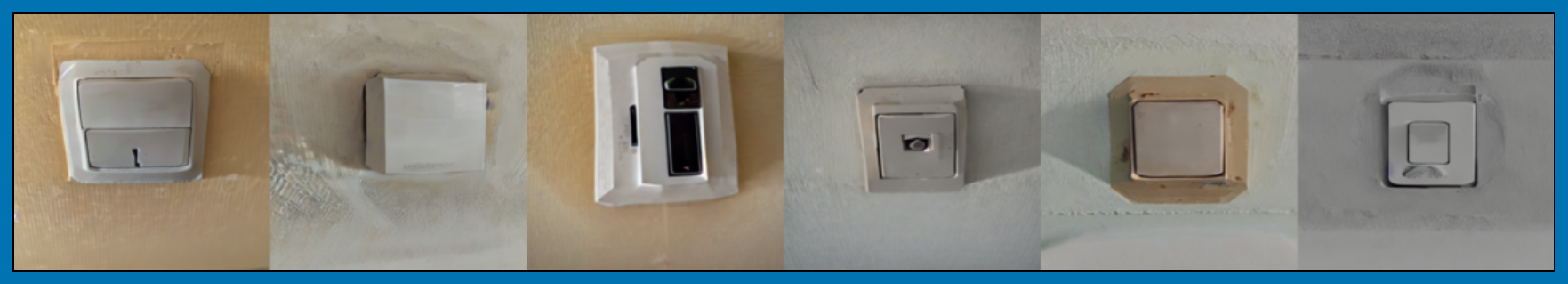}  
            \includegraphics[width=0.49\textwidth]{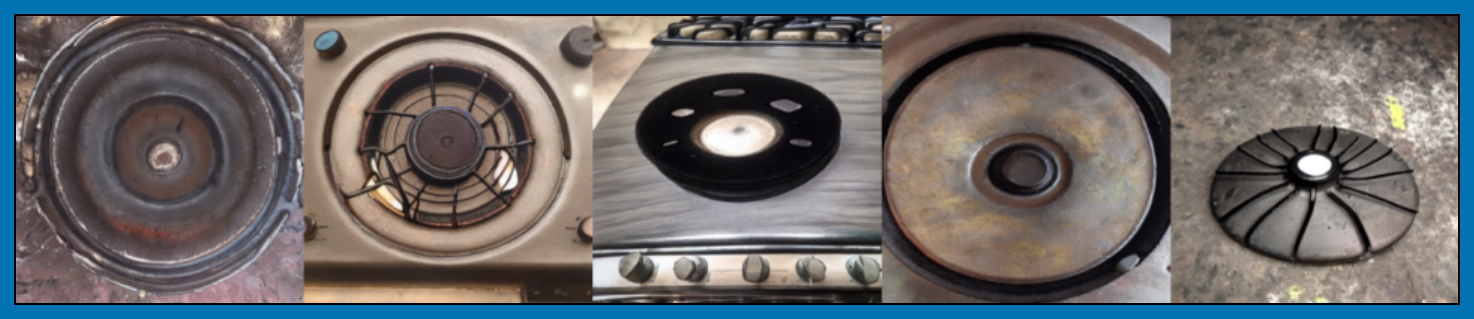}  \includegraphics[width=0.49\textwidth]{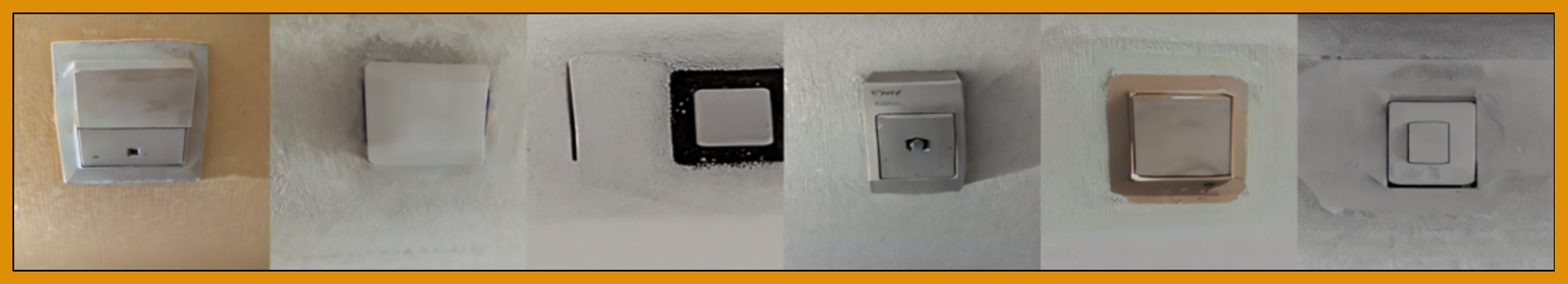}  
            \includegraphics[width=0.49\textwidth]{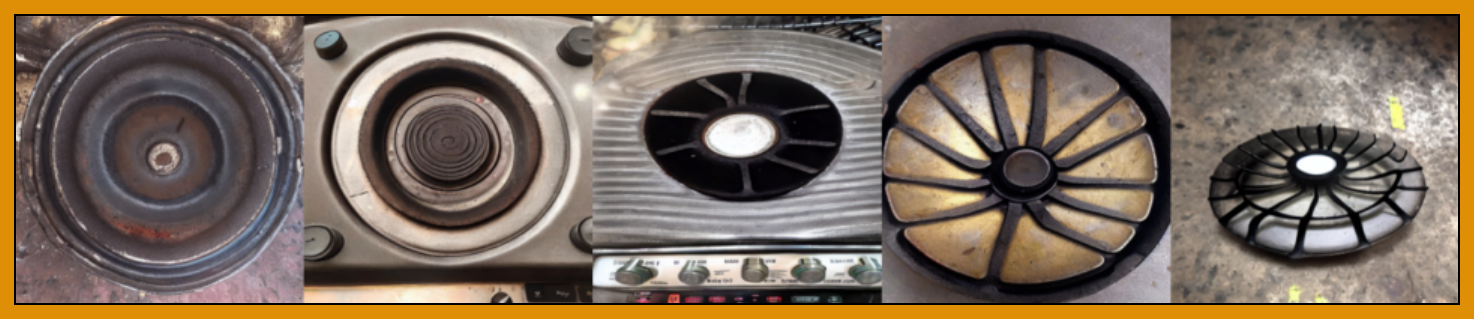}  \includegraphics[width=0.49\textwidth]{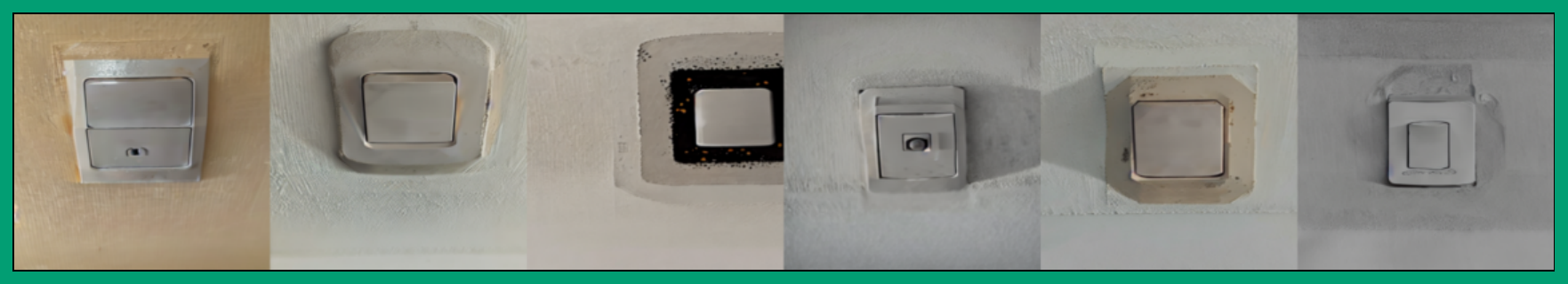}  \includegraphics[width=0.49\textwidth]{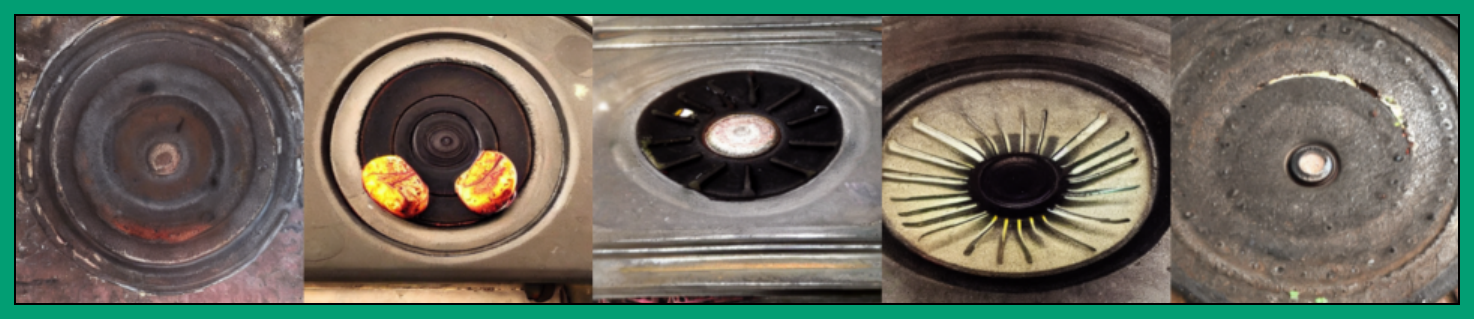}  
            \caption{Textual Inversion}
        \end{subfigure} 
        
        \begin{subfigure}[b]{\textwidth}
            \centering\includegraphics[width=0.49\textwidth]{gen_light_switch_Africa_2024_02_20_21_31_38_770133_0.06_6.pdf}
            \includegraphics[width=0.49\textwidth]{gen_stove_Africa_2024_02_20_21_31_38_770133_0.06_6.pdf}
            \includegraphics[width=0.49\textwidth]{gen_light_switch_Europe_2024_02_20_21_31_38_770133_0.06_6.pdf}
            \includegraphics[width=0.49\textwidth]{gen_stove_Europe_2024_02_20_21_31_38_770133_0.06_6.pdf}
            \includegraphics[width=0.49\textwidth]{gen_light_switch_Southeast_Asia_2024_02_20_21_31_38_770133_0.06_6.pdf}
            \includegraphics[width=0.49\textwidth]{gen_stove_Southeast_Asia_2024_02_20_21_31_38_770133_0.06_6.pdf}
            \caption{ Contextualized \methodname}
        \end{subfigure}
    \end{subfigure}
    \caption{
    Generated images of light switches (\textbf{Left}) and stoves (\textbf{Right}).
    The same seed is used for a given column, and the box colors indicate images pertaining to \textcolor{blue}{Africa}, \textcolor{orange}{Europe}, and \textcolor{darkgreen}{Southeast Asia}.
    }
    \label{fig:res_compare_methods}
\end{figure}

In this section we visually study the generations of different baseline methods presented in the paper.  
Figure \ref{fig:res_compare_methods_real} shows real images of light switches and stoves, while \cref{fig:res_compare_methods_noguidanace} and \cref{fig:res_compare_methods} show examples of different methods without and with additional information (respectively), applied to the same seed and prompt. 

For methods without additional information, we see that synonyms can lead to increased diversity, beyond the variation that is present in the real dataset.

For methods with additional information, Semantic guidance and CLIP FG with Loss and Entropy aim to remove the regional information. 
Textual inversion uses the learned prompt embedding of a given object instead of relying on the prompt. Upon visual inspection, we observe that CLIP FG with Entropy and c-\methodnameshort have the highest diversity and details in generations. 
Images with textual inversion have very low diversity, although they look most similar to the real images in coloring and camera angle.

\subsection{Computational Complexity}\label{app:comp_complexity}

We compute the average wall-clock time (in seconds) to generate one sample for a given method. The results are reported in Table \ref{tab:time_complexity}, averaged over 100 samples computed on the same model and the same GPU machine (A100) without batch-generation. 

We report both the generation time required per sample as well as the pre-computation time required for the GeoDE dataset. Note that many methods and baselines such as LDM, Synonyms, Semantic Guidance and VSG do not require any pre-computation.

For Paraphrasing baseline, 150 seconds is required to optimize the meta-prompt using the LLM. For textual inversion, per object, 30 minutes is required to learn the text embedding. This text embedding is then reused during inference time. For c-VSG, we pre-compute the similarity matrix for the exemplar reference images (2 per object). These values are then re-used during generation time.

Note that faster GPU machines with larger RAM result in faster sample generation. To further improve the sampling time complexity, one can also consider using diffusion distillation techniques such as consistency models~\cite{song2023consistency} or progressive distillation~\cite{salimans2022progressive}.
\begin{table}[t]
  \centering
  \caption{\textcolor{black}{Average generation and pre-computation time in seconds. Generation time for each method is reported on an average of 180 samples with float16 precision using the same GPU machine. Values are subject to change on different GPU machines and different computation precision. Pre-computation time is reported for the GeoDE dataset with 27 objects.}}
  \resizebox{\textwidth}{!}{%
    \begin{tabular}{lccc}
    \toprule
    {\textbf{Method}} & {\textbf{Generation Time (seconds)}} & \textbf{\ Pre-computation Time (seconds)}\\
    \midrule
    LDM / Synonyms  & $2.4915 \pm 0.0003$ & \ding{55} \\ 
    Paraphrasing & $2.4915 \pm 0.0003$ & 150 \\ 
    Semantic Guidance& $ 3.1981 \pm 0.0003$ & \ding{55}   \\ 
    FG CLIP (loss)& $7.01245 \pm 0.0007$ & \ding{55}   \\ 
    FG CLIP (entropy) & $7.13161 \pm 0.0001$  & \ding{55}      \\ 
    Textual Inversion & $2.5505 \pm 0.0003$ & 48600 \\ 
    Vendi Score Guidance (VSG)   & $ 6.5968\pm 0.0007$  & \ding{55}   \\ 
    contextualized Vendi Score Guidance (c-VSG)  & $ 6.6925\pm 0.0009$ &  $3$  \\ 
    \bottomrule
  \end{tabular}\label{tab:time_complexity}
  }
\end{table}

To additionally reduce the generation time, one can increase the guidance frequency (Gfreq in Algorithm 1). In all our experiments we used a Gfreq of 5, meaning c-VSG was applied every 5 steps of the reverse diffusion process. However, by increasing the Gfreq to 25, and applying c-VSG every 25 steps, we can substantially reduce the computation time while having significant improvements over the baseline LDM in terms of the recall and F1. See Table \ref{tab:geode_res_freq}.

\begin{table}[h!]
\centering
\caption{We observe significant improvement in terms of the F1 and recall with Gfreq=25 compared to LDM with a very little compute overhead. Generation reported on an \textbf{A100} GPU.}
\resizebox{\textwidth}{!}{%
\begin{tabular}{lcccccccccc}
\toprule
\textbf{Method} & \textbf{Gen.} & \textbf{Guidance} & \multicolumn{2}{c}{\textbf{F1}} & \multicolumn{2}{c}{\textbf{Precision}} & \multicolumn{2}{c}{\textbf{Recall}} & \multicolumn{2}{c}{\textbf{CLIPScore}}\\
\cmidrule(lr){4-5}
\cmidrule(lr){6-7}
\cmidrule(lr){8-9}
\cmidrule(lr){10-11}
& \textbf{time} & \textbf{Freq.} & \text{Avg.} & \text{Worst-Reg.} & \text{Avg.} & \text{Worst-Reg.} & \text{Avg.} & \text{Worst-Reg.} & \text{Avg.} & \text{Worst-Reg.}\\
\midrule
    LDM   & 2.49 (s) & - & 0.364 & 0.322 & {0.413} & 0.273 & 0.337 & 0.395 & 0.242 &0.218 \\
    {c-VSG}  & 2.95 (s)& 25 &  0.420 & {0.384} & 0.364 & 0.327 & 0.498 & 0.465 & 0.240 &0.236 \\
    {c-VSG}  & 6.69 (s)& 5 &  {0.455} & {0.444} & {0.424} & {0.417} & {0.493} & {0.476} & {0.254} & 0{.253}\\
    
\bottomrule
\end{tabular}
  \label{tab:geode_res_freq}}
\end{table}

\subsection{Applications of c-\methodnameshort}
While we save deeper investigations of downstream applications to future work, we believe that c-VSG may be applied to increase representation diversity in interfaces that generate \textit{multiple} images per user-provided prompt, such as in Gemini,\footnote{https://gemini.google.com/app} MetaAI,\footnote{https://www.meta.ai/} and StableDiffusion.\footnote{https://huggingface.co/spaces/stabilityai/stable-diffusion}
In addition, c-VSG may be useful in increasing the utility of recent efforts \cite{he2023syntheticdatagenerativemodels,hemmat2023feedbackguided} that explore generating hundreds of images for a single class to supplement training data for discriminative models by enabling more diverse generations.
\end{document}